\documentclass{article}

\usepackage{pratik}
\def \d      {\trm{d}}
\def \dt    {\d\t}
\def \dw    {\d w}

\def \ns    {{N_s}}
\def \nt    {{N_t}}

\def \kl    {\trm{KL}}
\def \dfr    {d_{\trm{FR}}}
\def \ws    {{w_s}}
\def \wb    {{w_t}}
\def \wt    {{w(\tau)}}
\def \wtp   {{w(\tau+\d\tau)}}
\def \wtk    {{w^k(\tau)}}
\def \wtpk    {{w^k(\tau+\d \tau)}}

\def \xs    {x_s}
\def \xt    {x_t}
\def \ys    {y_s}
\def \yt    {y_t}
\def \omt    {\Omega_\tau}

\def \Ds    {{D_s}}
\def \Dt    {{D_t}}

\def \ph    {\hat{p}}
\def \ps    {{p_s}}
\def \pt    {{p_t}}
\def \psh    {\hat{p}_s}
\def \pth    {\hat{p}_t}
\def \ptau    {{p_\tau}}
\def \ptauh    {{\hat{p}_\tau}}

\def \ells {\ell_s}

\def \elltau {\ell_\tau}

\def \fr {\text{FR}}

\def \ds   {\d s}

\usepackage[hang,flushmargin]{footmisc}
\usepackage[accepted]{icml2021}

\usepackage[title]{appendix}

\def \mscale {0.9}

\begin{document}

\twocolumn[
\icmltitle{An Information-Geometric Distance on the Space of Tasks}
\begin{icmlauthorlist}
\icmlauthor{Yansong Gao}{1}
\icmlauthor{Pratik Chaudhari}{2}
\end{icmlauthorlist}
\icmlaffiliation{1}{Department of Applied Mathematics and Computational Science,
University of Pennsylvania}
\icmlaffiliation{2}{Department of Electrical and Systems Engineering,
University of Pennsylvania}
\icmlcorrespondingauthor{Yansong Gao}{gaoyans@sas.upenn.edu}
\icmlcorrespondingauthor{Pratik Chaudhari}{pratikac@seas.upenn.edu}
\vskip 0.15in
]
\printAffiliationsAndNotice{}

\begin{abstract}
This paper prescribes a distance between learning tasks modeled as joint distributions on data and labels. Using tools in information geometry, the distance is defined to be the length of the shortest weight trajectory on a Riemannian manifold as a classifier is fitted on an interpolated task. The interpolated task evolves from the source to the target task using an optimal transport formulation. This distance, which we call the ``coupled transfer distance'' can be compared across different classifier architectures. We develop an algorithm to compute the distance which iteratively transports the marginal on the data of the source task to that of the target task while updating the weights of the classifier to track this evolving data distribution. We develop theory to show that our distance captures the intuitive idea that a good transfer trajectory is the one that keeps the generalization gap small during transfer, in particular at the end on the target task. We perform thorough empirical validation and analysis across diverse image classification datasets to show that the coupled transfer distance correlates strongly with the difficulty of fine-tuning.
\end{abstract}

\section{Introduction}
\label{s:intro}

A part of the success of Deep Learning stems from the fact that deep networks learn features that are discriminative yet flexible. Models pre-trained on a particular task can be easily adapted to perform well on other tasks. The transfer learning literature forms an umbrella for such adaptation techniques, and it works well, see for instance~\citet{mahajan2018exploring,dhillon2019a,kolesnikov2019large,joulinLearningVisualFeatures2016} for image classification or~\citet{47751} for language modeling, to name a few large-scale studies. There are also situations when transfer learning does not work well, e.g., a pre-trained model on ImageNet is a poor representation to transfer to MRI data~\citep{merkow2017deepradiologynet}.

It stands to reason that if source and target tasks are ``close'' to each other then we should expect transfer learning to work well. It may be difficult to transfer across tasks that are ``far away''. We lack theoretical tools to characterize the difficulty of adapting a model training on a source task to the target task. While there are numerous candidates in the literature (see Related Work in~\cref{s:related_work}) for characterizing the distance between tasks, a unified understanding of these domain-specific methods is missing.

\begin{figure}[!t]
\centering
\includegraphics[width=0.6\linewidth]{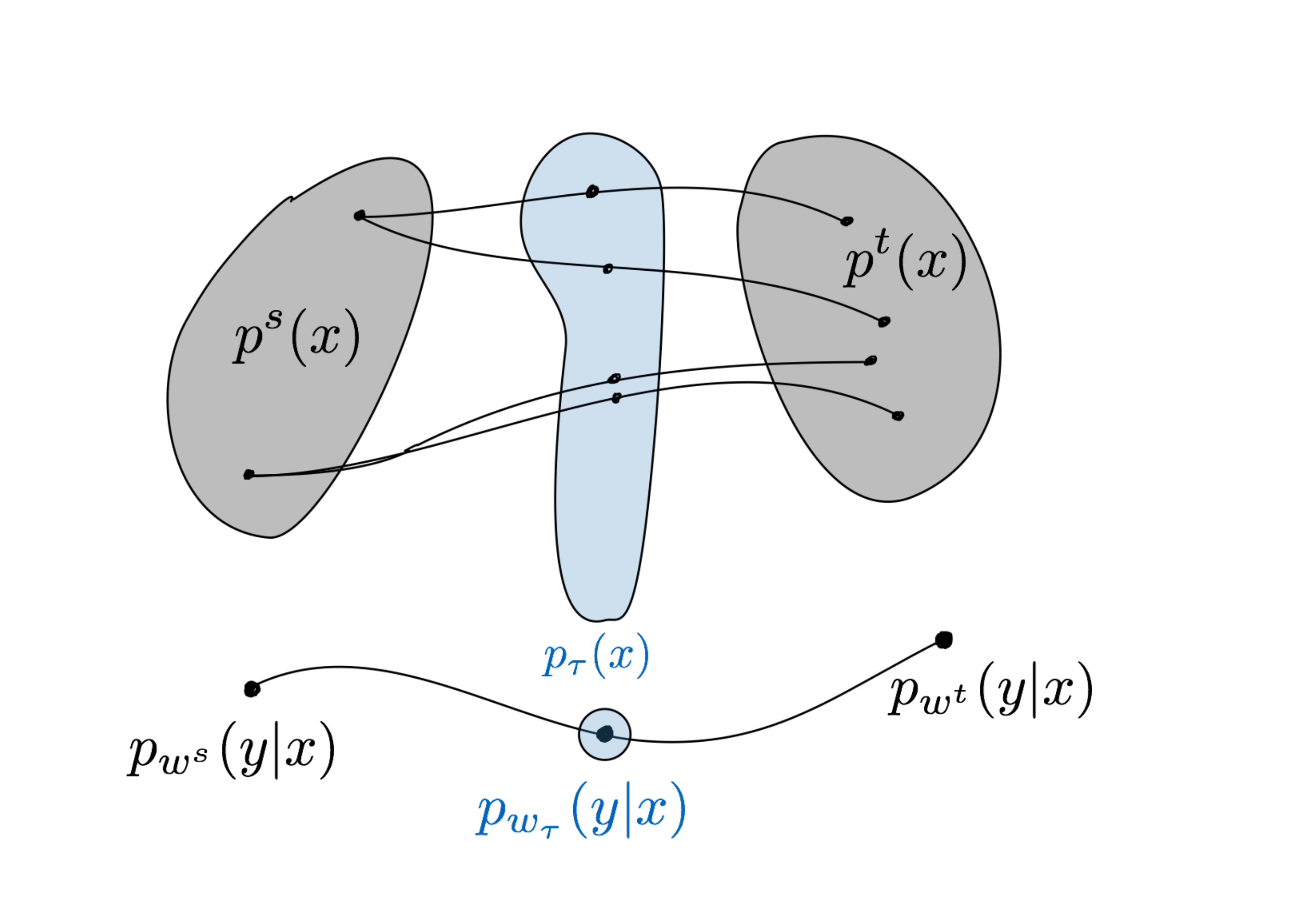}
\caption
{\tbf{Coupled transfer of the data and the conditional distribution}.
We solve an optimization problem that transports
the source data distribution $\ps(x)$ to the
target distribution $\pt(x)$ as $\t \to 1$
while simultaneously updating the model
using samples from the interpolated distribution $\ptau(x)$. This modifies the
conditional distribution $p_\ws(y|x)$ on the source task to the
corresponding distribution on the target task $p_\wb(y|x)$.
The ``coupled transfer distance'' between source and target tasks is the length
of the shortest such weight trajectory under the Fisher Information Metric.
}
\label{fig:schematic}
\end{figure}

\heading{Desiderata}
Our desiderata for a task distance are as follows. First, it should be a distance between learning tasks, i.e., it should explicitly incorporate the hypothesis space of the model that is being transferred and accurately reflect the difficulty of transfer. For example, it is often observed in practice that transferring larger models is easier, we would like our task distance to capture this fact. Such a distance is different than discrepancy measures on the input, or the joint input-output space, which do not consider the model.

Second, we would like a theoretical framework to prescribe this distance. Task distances in the literature often depend upon quantities such as the number of epochs of fine-tuning to reach a certain accuracy, where different hyper-parameters may result in different conclusions. Also, as the present paper explores at depth, there are mechanisms for transfer other than fine-tuning that may transfer easily across tasks that are considered far away for fine-tuning.

\heading{Contributions}
We formalize a ``coupled transfer distance'' between learning tasks as the length of the shortest trajectory on a Riemannian manifold (statistical manifold of parametrized conditional distributions of labels given data) that the weights of a classifier travel on when they are adapted from the source task to the target task. At each instant during this transfer, weighs are fitted on a interpolating task that evolves along the optimal transportation (OT) trajectory between source and target tasks. Evolution of weights and the interpolated task is \emph{coupled} together. In particular, we set the ground metric which defines the cost of transporting unit mass in OT to be the Fisher-Rao distance.

We give an algorithm to compute the coupled transfer distance. It alternately update the OT map and the weight trajectory; the former uses the latest ground metric computed as the length of the weight trajectory under the Fisher Information Metric (FIM) whereas the weight trajectory is updated to fit to a new sequence of interpolated tasks given by the updated OT. We develop several techniques to scale up this algorithm and show that we can compute the coupled transfer distance between standard benchmark datasets.

We study this distance using Rademacher complexity. We show that given an OT between tasks, the Fisher-Rao distance between the initial and final weights, which our coupled transfer distance computes, corresponds to finding a weight trajectory that keeps the generalization gap small on the interpolated tasks. The coupled transfer distance thus captures the intuitive idea that a good transfer trajectory is the one that keeps the generalization gap small during transfer, in particular at the end on the target task.

We perform thorough empirical validation and analysis of the coupled transfer distance across diverse image classification datasets (MNIST~\citep{lecun1998gradient}, CIFAR-10, CIFAR-100~\citep{krizhevskylearningmultiplelayers2009} and Deep Fashion~\citep{liuLQWTcvpr16DeepFashion}).

\section{Theoretical setup}

We are interested in the supervised learning problem in this paper.
Consider a source dataset $\Ds = \cbr{(\xs^i, \ys^i)}_{i=1}^{\ns}$ and a target dataset $\Dt = \cbr{(\xt^i, \yt^i)}_{i=1}^{\nt}$ where $\xs^i, \xt^i \in X$ denote input data and $\ys^i, \yt^i \in Y$ denote ground-truth annotations. Training a parameterized classifier, say a deep network with weights $w \in \reals^p$, on the source task involves minimizing the cross-entropy loss $\ells(w) = -\f{1}{\ns} \sum_{i=1}^\ns \log p_w(\ys^i | \xs^i)$ using stochastic gradient descent (SGD):
\beq{
    w(\t + \dt)  = w(\t) - \hat{\grad} \ells(w(\t))\ \dt;\ w(0) = \ws;
    \label{eq:sgd}
}
The notation $\widehat{\grad} \ells(w)$ indicates a stochastic estimate of the gradient using a mini-batch of data. The parameter $\d \t$ is the learning rate. Let us define the distribution $\psh(x,y) = \f{1}{\ns} \sum_{i=1}^\ns \delta_{\xs^i}(x) \delta_{\ys^i}(y)$ and its input-marginal $\psh(x) = \f{1}{\ns} \sum_{i=1}^\ns \delta_{\xs^i}(x)$;
distributions $\pth(x,y), \pth(x)$ are defined analogously.

\subsection{Fisher-Rao metric on the manifold of probability distributions}
\label{ss:fisher_rao}

Consider a manifold $\MM = \cbr{p_w(z): w \in \reals^p}$ of probability distributions. Information Geometry~\citep{amariInformationGeometryIts2016} studies invariant geometrical structures on such manifolds.
For two points $w,w' \in \MM$, we can use the Kullback-Leibler (KL) divergence
$
    \kl \sbr{p_w, p_{w'}} = \int \d p_w(z) \log \rbr{p_w(z)/p_{w'}(z)},
$
to obtain a Riemannian structure on $M$. This allows the infinitesimal distance $\ds$ on the manifold to be written as
\beq{
    \ds^2 = 2 \kl \sbr{p_w, p_{w + \dw}} = \sum_{i,j=1}^p g_{ij}\ \dw_i \dw_j
    \label{eq:riemannian_distance}
}
\beq{
    g_{ij}(w) = \int \d p_w(z) \rbr{\partial_{w_i} \log p_w(z)} \rbr{\partial_{w_j} \log p_w(z)}
    \label{eq:fim_def}
}
are elements of the Fisher Information Matrix (FIM) $g$. Weights $w$ play the role of a coordinate system for computing the distance. The FIM is the Hessian of the $\kl$-divergence; we may think of the FIM as quantifying the amount of information present in the model about the data it was trained on. The FIM is the unique metric on $\MM$ (up to scaling) that is preserved under diffeomorphisms~\citep{bauerUniquenessFisherRao2016}, in particular under representation of the model.

Given a continuously differentiable curve $\cbr{w(\t)}_{\t \in [0,1]}$ on the manifold $M$ we can compute its length by integrating the infinitesimal distance $\abr{\ds}$ along it. The shortest length curve between two points
$w, w' \in \MM$ induces a metric on $\MM$ known as the Fisher-Rao distance~\citep{rao1945information}
\beq{
    \scalemath{\mscale}{
    \dfr(w, w') = \min_{\substack{w:\ w(0) = w\\ w(1) = w'}}\ \int_0^1
        \sqrt{ \inner{\dot{w}(\t)}{g(w(\t)) \dot{w}(\t)}}\ \dt}.
    \label{eq:fisher_rao}
}
Shortest paths on a Riemannian manifold are geodesics, i.e., they are locally ``straight lines''.

\heading{Computing the Fisher-Rao distance by integrating the KL-divergence}
Let us focus on the conditional distribution $p_w(y | x)$. For the factorization $p(x,y) = p(x) p(y | x)$ where only the latter is parametrized, the FIM in~\cref{eq:fim_def} is given by
\[
    \scalemath{\mscale}{
    g_{ij}(w) = \E_{\substack{x \sim p(x),\ y \sim p_w(y | x)}}
            \sbr{\partial_{w_i} \log p_w(y | x)\ \partial_{w_j} \log p_w(y | x)
            }
    };
\]
here the input distribution $p(x)$ and the weights $w$ will be chosen in the following sections. The FIM is difficult to compute for large models and approximations often work poorly~\citep{kunstner2019limitations}. For our purposes, we only need to compute the infinitesimal distance $\abr{\ds}$ in~\cref{eq:riemannian_distance} and can thus rewrite~\cref{eq:fisher_rao} as
\beq{
    \scalemath{0.85}{
    \dfr(w, w') = \min_{\substack{w:\ w(0) = w\\ w(1) = w'}}\ \int_0^1
        \sqrt{ 2 \kl[p_w(y | x), p_{w+\dw}(y | x)]}
        }.
    \label{eq:fisher_rao_kl}
}

\subsection{Transporting the data distribution}
\label{ss:transport}

We next focus on the marginals on the input data $\psh(x)$ and $\pth(x)$ for the source and target tasks respectively. We are interested in computing a distance between the source marginal and the target marginal and will use tools from optimal transportation (OT) for this purpose; see~\citet{santambrogio2015optimal,peyreComputationalOptimalTransport2019} for an elaborate treatment.

\heading{OT for continuous measures}
Let $\Pi(\ps, \pt)$ be the set of joint distributions (also known as couplings or transport plans) with first marginal equal to $\ps(x)$ and second marginal $\pt(x)$. The Kantorovich relaxation of OT solves for
\[
    \inf_{\g  \in \Pi(\ps, \pt)} \int c(x, x')\ \d \g(x,x')
\]
to compute the best coupling $\g^* \in \Pi$. The cost $c(x, x') \in \reals_+$ is called the ground metric. It gives the cost of transporting unit mass from $x$ to $x'$. The popular squared-Wasserstein metric $W^2_2(\ps, \pt)$ uses $c(x, x') = \norm{x-x'}_2^2$. Given the optimal coupling $\g^*$, we can compute the trajectory that transports probability mass using displacement interpolation~\citep{mccannConvexityPrincipleInteracting1997}. For example, for the Wasserstein metric, $\g^*$ is a constant-speed geodesic, i.e., if $\ptau$ is the distribution at an intermediate time instant $\t \in [0, 1]$ then its distance from $\ps$ is proportional to $\t$
\[
    W_2(\ps, \ptau) = \t  W_2(\ps, \pt).
\]

\heading{OT for discrete measures}
We are interested in computing the constant-speed geodesic for discrete measures $\psh(x)$ and $\pth(x)$. The set of transport plans in this case is $\Pi(\psh, \pth) = \cbr{\G \in \reals^{\ns \times
    \nt}_+:\ \G \ones_\ns = \psh, \G^\top \ones_\nt = \pth}$
and the optimal coupling is given by
\beq{
    \G^* = \argmin_{\G \in \Pi(\psh, \pth)} \cbr{\inner{\G}{C} - \e H(\G)};
    \label{eq:ot_finite}
}
here $C_{ij}$ is a matrix that defines the ground metric in OT. For instance, $C_{ij} = \norm{x_i - x'_j}_2^2$ for the Wasserstein metric. The first term above measures the total cost $\sum_{ij} \G_{ij} C_{ij}$ incurred for the transport. The second term is an entropic penalty $H(\G) = -\sum_{ij} \G_{ij} \log \G_{ij}$ popularized by~\citet{cuturi2013sinkhorn} that accelerates the solution of the OT problem. McCann's interpolation for the discrete case with $C_{ij} = \norm{\xs^i - \xt^j}_2^2$ can be written explicitly as a sum of Dirac-delta distributions supported at interpolated inputs $x = (1 - \t) \xs^i + \t \xt^j$
\beq{
    \ptauh(x) = \sum_{i=1}^\ns \sum_{j=1}^\nt \G^*_{ij}\ \delta_{(1 - \t) \xs^i + \t \xt^j}(x).
    \label{eq:interp_finite_only_x}
}
We can also create pseudo labels for samples from $\ptau$ by a linear interpolation of the one-hot encoding of their respective labels to get
\beq{
    \scalemath{0.85}{\ptauh(x, y) = \sum_{i=1}^\ns \sum_{j=1}^\nt \G^*_{ij}\ \delta_{(1 - \t) \xs^i + \t \xt^j}(x)\
    \delta_{(1 - \t) \ys^i + \t \yt^j}(y)
    }.
    \label{eq:interp_finite}
}

\section{Coupled Transfer Distance}
\label{s:methods}

We next combine the development of~\crefrange{ss:fisher_rao}{ss:transport} to transport the marginal on the data and modify the weights on the statistical manifold simultaneously. We call this method the ``coupled transfer process'' and the corresponding task distance as the ``coupled transfer distance''. We also discusses techniques to efficiently implement the process and make it scalable to large deep networks.

\subsection{Uncoupled Transfer Distance}
\label{ss:uncoupled_transfer_distance}

We first discuss a simple transport mechanism instead of OT and discuss how to compute a transfer distance. For $\t \in [0,1]$, consider the mixture distribution
\beq{
    \ptauh(x, y) = (1 -\t) \psh(x, y) + \t \pth(x, y).
    \label{eq:interp_mixture}
}
Samples from $\ptauh$ can be drawn by sampling an input-output pair from $\psh$ with probability $1-\t$ and sampling it from $\pth$ otherwise. At each time instant $\t$, the uncoupled transfer process updates the weights the classifier using SGD to fit samples from $\ptauh$
\beq{
    w(\t + \dt) = w(\t) - \hat{\grad} \elltau(w(\t))\ \dt;\ w(0) = \ws.
    \label{eq:sgd_interp}
}
Weights $w(\t)$ are thus fitted to each task $\ptau$ as $\t$ goes from 0 to 1.
In particular for $\t = 1$, weights $w(1)$ are fitted to $\pth$. As $\dt \to 0$, we obtain a continuous curve $\cbr{w(\t): t \in [0,1]}$. Computing the length of this weight trajectory using~\cref{eq:fisher_rao_kl} gives a transfer distance.

\begin{remark}[Uncoupled transfer distance entails longer weight trajectories]
\label{rem:uncoupled_longer}
For uncoupled transfer, although the task and weights are modified simultaneously, their changes are not synchronized. We therefore call this the ``uncoupled transfer distance''. To elucidate, changes in the data using the mixture~\cref{eq:interp_mixture} may be unfavorable to the current weights $w(\t)$ and may cause the model to struggle to track the distribution $\ptauh$. This forces the weights to take a longer trajectory in information space, i.e., as measured by the Fisher-Rao distance in~\cref{eq:fisher_rao_kl}. If changes in data were synchronized with the evolving weights, the weight trajectory would be necessarily shorter in information space because the KL-divergence in~\cref{eq:riemannian_distance} is large when the conditional distribution changes quickly to track the evolving data. We therefore expect the task distance computed using the mixture distribution to be larger than the coupled transfer distance which we will discuss next; our experiments in~\cref{s:expt} corroborate this.
\end{remark}

\subsection{Modifying the task and classifier synchronously}
\label{ss:synchronous_transfer}

Our coupled transfer distance that uses OT to modify the task and updates the weights synchronously to track the interpolated distribution is defined as follows.

\begin{definition}[Coupled transfer distance]
\label{def:coupled}
Given two learning tasks $\Ds$ and $\Dt$ and a $w$-parametrized classifier trained on $\Ds$ with weights $\ws$,
the coupled transfer distance between the tasks is
\beq{
    \scalemath{1}{
    \min_{\G, w(\cdot)}
            \E_{x \sim \hat{p}_\t(x)}
            \int_0^1 \sqrt{2 \kl \sbr{p_w(\cdot\ |\ x), p_{w+\dw}(\cdot\ |\ x) }}
    }
    \label{eq:coupled_transfer_def}
}
where and couplings $\G \in  \Pi(\psh(x), \pth(x))$ and $w(\cdot)$ is a continuous curve which is the limit of
\[
    w(\t + \dt) = w(\t) - \hat{\grad} \elltau(w(\t))\ \dt;\ w(0) = \ws.
\]
as $\dt \to 0$. The interpolated distribution $\hat{p}_\t(x,y)$ at time instant $\t \in [0,1]$ for a coupling $\G$ is given by~\cref{eq:interp_finite} and the loss $\elltau$ is the cross-entropy loss of fitting data from this interpolated distribution.
\end{definition}

The following remarks discuss the rationale and the properties of this definition.

\begin{remark}[Coupled transfer distance is asymmetric]
\label{rem:coupled_asymmetric}
The length of the weight trajectory for transferring from $\psh$ to $\pth$ is different from the one that transfers from $\pth$ to $\psh$. This is a desirable property, e.g., it is easier to transfer from ImageNet to CIFAR-10 than in the opposite direction.
\end{remark}

\begin{remark}[Coupled transfer distance can be compared across different architectures]
\label{rem:coupled_comparable}
An important property of the task distance in~\cref{eq:coupled_transfer_def} is that it is the Fisher-Rao distance, i.e., the shortest geodesic on the statistical manifold, of conditional distributions $p_{w(0)}(\cdot | \xs^i)$ and $p_{w(1)}(\cdot | \xt^i)$ with the coupling $\G$ determining the probability mass that is transported from $\xs^i$ to $\xt^j$. Since the Fisher-Rao distance, does not depend on the embedding dimension of the manifold $M$, the coupled transfer
distance does not depend on the architecture of the classifier; it only depends upon the capacity to fit the conditional distribution $p_w(y | x)$. This is a very desirable property: given the tasks, our distance is comparable across different architectures. Let us note that the uncoupled transfer distance in~\cref{ss:uncoupled_transfer_distance} also shares this property but coupled transfer has the benefit of computing the shortest trajectory in information space; weight trajectories of uncoupled transfer may be larger; see~\cref{rem:uncoupled_longer}.
\end{remark}

\subsection{Computing the coupled transfer distance}
\label{ss:algorithm}

We first provide an an informal description of how we compute the task distance. Each entry $\G_{ij}$ of the coupling matrix determines how much probability mass from $\xs^i$ is transported to $\xt^j$. The interpolated distribution~\cref{eq:interp_finite} allows us to draw samples from the task at an intermediate instant. For each coupling $\G$, there exists a trajectory of weights $w(\cdot) := \cbr{w(\t): \t \in [0,1]}$ that tracks the interpolated task. The algorithm treats $\G$ and the weight trajectory as the two variables and updates them alternately as follows. At the $k^{\text{th}}$ iteration, given a weight trajectory $w^k(\cdot)$ and a coupling $\G^k$, we set the entries of the ground metric $C^{k+1}_{ij}$ to be the Fisher-Rao distance between distributions $p_{w(0)}(\cdot | \xs^i)$ and $p_{w(1)}(\cdot | \xt^i)$. An updated $\G^{k+1}$ is calculated using this ground metric to result in a new trajectory $w^{k+1}(\cdot)$ that tracks the new interpolated task distribution~\cref{eq:interp_finite} for $\G^{k+1}$.

More formally, given an initialization for the coupling matrix $\G^0$ we perform the updates in~\cref{eq:algorithm}. Computing the coupled transfer distance is a non-convex optimization problem and we therefore include a proximal term in~\cref{eq:gkp} to keep the coupling matrix close to the one computed in the previous step $\G^k$. This also indirectly keeps the weight trajectory $w^{k+1}(\cdot)$ close to the trajectory from the previous iteration. Proximal point iteration~\citep{bauschkeConvexAnalysisMonotone2017} is insensitive to the step-size $\l$ and it is therefore beneficial to employ it in these updates.

\begin{figure}[htpb]
\begin{subequations}
\vspace*{-1em}
\aeq{
    &\scalemath{\mscale}{
    \G^k = \argmin_{\G \in \Pi} \cbr{\inner{\G}{C^k} - \e H(\G)
            + \l \norm{\G -\G^{k-1}}_{\text{F}}^2 }},
    \label{eq:gkp}\\
    &\scalemath{\mscale}{C^k_{ij} = \int_0^1 \sqrt{2 \kl\sbr{p_{\wtk}(\cdot |
    x_\tau^{ij}), p_{\wtpk}(\cdot | x_\tau^{ij})}}},
    \label{eq:ckp}\\
    &\scalemath{0.9}{
    w^k(\t + \dt) = w^k(\t) - \hat{\grad} \elltau(w^k(\t))\ \dt;\ w(0) = \ws.
    }
    \label{eq:sgd_prox}\\
    &\scalemath{0.8}{
    \ptauh(x, y) = \sum_{i=1}^\ns \sum_{j=1}^\nt \G^{k-1}_{ij}\ \delta_{(1 - \t) \xs^i + \t \xt^j}(x)\
    \delta_{(1 - \t) \ys^i + \t \yt^j}(y)},
    \label{eq:interp_finite_algorithm}\\
    &x_\tau^{ij}, y_\tau^{ij} \sim \ptauh(x, y).
}
\label{eq:algorithm}
\vspace*{-2em}
\end{subequations}
\end{figure}

\subsection{Practical tricks for efficient computation}
\label{ss:pratical_tricks}

The optimization problem formulated in~\cref{eq:algorithm} is conceptually simple but computationally daunting. The main hurdle is to compute the ground metric $C^k_{ij}$ for all $ i\leq \ns, j\leq \nt$ pairs in a dense transport coupling $\G$. The coupling matrix can be quite large, e.g., it has $10^8$ entries for a relatively small dataset of $\ns = \nt = 10,000$. We therefore introduce the following techniques that allow us to scale to large problems.

\heading{Block-diagonal transport couplings}
Instead of optimizing $\G$ in~\cref{eq:coupled_transfer_def} over the entire polytope $\Pi(\psh, \pth)$, we only consider block-diagonal couplings. Depending upon the source and target datasets, we use blocks of size up to 30$\times$30. At each time instant $\t \in [0, 1]$, we sample a block from the transport coupling. SGD in~\cref{eq:sgd_prox} updates weights using multiple samples from the interpolated task restricted to this block. The integrand for $C^k_{ij}$ in~\cref{eq:ckp} is also computed only on this mini-batch. Experiments in~\cref{s:expt} show that the weight trajectory converges using this technique. We can compute the coupling transfer distance for source and target datasets of size up to $\ns = \nt = 19,200$. Other approaches for handling large-scale OT problems such as hierarchical methods~\citep{lee2019hierarchical} or greedy computation~\citep{carlier2010knothe} could also be used for our purpose but we chose this one for sake of simplicity.

\heading{Initializing the transport coupling}
The ground metric $C_{ij} = \norm{\xs^i - \xt^j}_2^2$ is widely used in the OT literature. We are however interested in computing distances for image-classification datasets in this paper and such a pixel-wise distance is not a reasonable ground metric for visual data that have strong local/multi-scale correlations. We therefore set $\G^0$ to be the block-diagonal approximation of the transport coupling for the ground metric $C_{ij} = \norm{\varphi(\xs^i) - \varphi(\xt^j)}_2^2$ where $\varphi$ is some feature extractor. The feature space is much more Euclidean-like than the input space and this gives us a good initialization in practice; similar ideas are employed in the metric learning literature~\citep{snell2017prototypical,hu2015deep,qi2018low}. We use a ResNet-50~\cite{heIdentityMappingsDeep2016} pre-trained on ImageNet to initialize $\G^0$ for all our experiments. To emphasize, \emph{we use the feature extractor only for initializing the transport coupling} further updates are performed using~\cref{eq:gkp}. We have computed the coupling transfer distance for MNIST without this step and our results are similar.

\heading{Using mixup to interpolate source and target images}
The interpolating distribution~\cref{eq:interp_finite} has a peculiar nature: sampled data $x_\tau^{ij}= (1-\t) \xs^i + \t \xt^j$ from this distribution are a convex combination of source and target data. This causes artifacts for natural images for $\t$ away from 0 or 1; we diagnosed this as a large value of the training loss while executing~\cref{eq:sgd_interp}. We therefore treat the coefficient of the convex combination in~\cref{eq:interp_finite} as if it were a sample from a Beta-distribution $\trm{Beta}(\t, 1-\t)$. This keeps the samples $x_\tau^{ij}$ similar to the source or the target task and avoids visual artifacts. This trick is inspired by Mixup regularization~
\cite{zhang2017mixup}; we also use Mixup for labels $y_\tau^{ij}$.

\section{An alternative perspective using Rademacher complexity}
\label{ss:rademacher}

We have hitherto motivated the coupled transfer distance using ideas in information geometry. In this section, we study the weight trajectory under the lens of learning theory. We show that we can interpret it as the trajectory that minimizes the integral of the generalization gap as the the weights are adapted from the source to the target task.
We consider binary classification tasks in this section. Rademacher complexity~\citep{bartlettRademacherGaussianComplexities2001}
\beq{
    \scalemath{0.9}{
    \RR_N(r) = \E_{\ph \sim p} \sbr{
    \E_\s \sbr{\sup_{w \in A(r)} \f{1}{N} \sum_{i=1}^N\ \s^i \ell(w; x^i, y^i)}}
    },
    \label{eq:rad}
}
is the average over draws of the dataset $\ph \sim p$ and iid random variables $\s^i$ uniformly distributed over $\cbr{-1,1}$ of the worst case average weighted loss $\s^i \ell(w; x^i, y^i)$ for $w$ in the set $A(r)$.
We assume here that $\abs{\ell(w; x^i, y^i)} < M$ and $\ell(w; x, y)$ is Lipschitz continuous.
Classical bounds bound the generalization gap of all hypotheses $h$ in a hypothesis class $\HH$ by
$\RR_{2N}(\HH) + 2 \sqrt{\f{\log(1/\delta)}{N}}$ with probability at least $1-\delta$.
We build upon this result to get the following theorem under the assumption that weights $w(\t)$ predict well on the interpolated task $\ptauh(x,y)$ at all times $\t$.



\begin{theorem}
\label{thm:integral_of_fisher_dist_main}
Given a weight trajectory $\cbr{\wt}_{\t \in [0,1]}$ and a sequence $0 = \t_0 \leq \t_1 <\t_2 <...<\t_{K}\leq 1 $, for all
$\e > 2 \sum_{k= 1}^{K}(\t_k - \t_{k-1})\E_{x \sim p_{\t}} |\Delta \ell(w(\t_{k-1}))|$, the probability that
\[
    \scalemath{0.9}{
    \f{1}{K}\sum_{k= 1}^{K} \rbr{\E_{(x,y) \sim p_{\t_k}} \sbr{\ell(\w(\t_k), x,y)}
    - \f{1}{N} \sum_{(x,y) \sim \hat{p}_{\t_k}} \ell(\w(\t_k), x,y)}
    }
\]
is greater than $\e$ is upper bounded by
\beq{
    \scalemath{0.8}{
    \exp \cbr{-\f{2 K}{M^2} \rbr{\e - 2 \sum_{k= 1}^{K} \D \t_k \E_{x \sim p_{\t_k}}
        \sbr{\sqrt{\inner{\dot{w}(\t_k)}{g(w(\t_k)) \dot{w}(\t_k)} }}
    }}}.
    \label{eq:upper_bound_fisher}
}
We have defined $\D \t_k = \t_k - \t_{k-1}$ and $\Delta \ell(\wt) = \ell(\wtp; x, y_\t(x)) -  \ell(\wt; x, y_\t(x))$.
\end{theorem}
\cref{ss:a:proof_integral_of_fisher_dist} gives the proof.
As $\D \t_k \to 0$
\[
    \scalemath{0.75}{
    \sum_{k= 1}^K \D \t_k \E_{x \sim p_{\t_k}}
        \sbr{\sqrt{\inner{\dot{w}(\t_k)}{g(w(\t_k)) \dot{w}(\t_k)}}} \to \int_0^1 \E_{x \sim \ptauh}
        \sbr{\sqrt{\inner{\dot{w}}{g(w) \dot{w}}}} \dt
        }
\]
which is the length of the trajectory on the statistical manifold with inputs drawn from the interpolated distribution at each instant.

We can thus think of the coupled transfer distance as the length of the trajectory on the statistical manifold that starts at the given model $\ws$ on the source task and ends with the model $w(1)$ fitted to the target task, as the task is simultaneously interpolated using an optimal transport whose ground metric between samples $\xs^i$ and $\xt^j$ is
$C_{ij} = \int_0^1 \sqrt{2 \kl \sbr{p_{w(\t)}(\cdot| x_\t^{ij}), p_{w(\t+\dt)}(\cdot| x_\t^{ij})}}$ which is the length of the trajectory under the FIM. This result is a crisp theoretical characterization of the intuitive idea that if one finds a weight trajectory that transfers from the source to the target task while keeping the generalization gap small at all time instants, then the length of the trajectory is a good indicator of the distance between tasks.

\section{Experiments}
\label{s:expt}

\subsection{Setup}
\label{ss:expt:setup}

We use the MNIST, CIFAR-10, CIFAR-100 and Deep Fashion datasets for our experiments. Source and target tasks consist of subsets of these datasets, each task with one or more of the original classes inside it. We show results using an 8-layer convolutional neural network with ReLU nonlinearities, dropout, batch-normalization with a final fully-connected layer along with a larger wide-residual-network WRN-16-4~\citep{zagoruyko2016wide}. ~\cref{s:a:expt} gives details about pre-processing, architecture and training.

\subsection{Baseline methods to estimate task distances}
\label{ss:baselines}

The difficulty of \tbf{fine-tuning is the gold standard of distance between tasks}. It is therefore very popular, e.g.,~\citet{kornblith2019better} use the number of epochs during transfer as the distance. We compute the length of the weight trajectory, i.e., $\int_0^1 \abr{\d w}$ and call this the \tbf{fine-tuning distance}. The trajectory is truncated when validation accuracy on the target task is 95\% of its final validation accuracy. No transport of the task is performed and the model directly takes SGD updates on the target task after being pre-trained on the source task.

The next baseline is \tbf{Task2Vec}~\citep{achille2019task2vec} which
embeds tasks using the diagonal of the FIM of a model
trained on them individually. Cosine distance between these vectors
is defined as the task distance.

We also compare with the \tbf{uncoupled transfer distance} developed in~\cref{ss:uncoupled_transfer_distance}. This distance computes length of the weight trajectory on the Riemannian distance and also interpolates the data but does not do them synchronously.

\tbf{Discrepancy measures on the input space} are a popular way to measure task distance. We show task distance computed as the \tbf{Wasserstein $W_2^2$ metric on the the pixel-space}, the \tbf{Wasserstein $W_2^2$ metric on the embedding space} and also method that we devised ourselves where we \tbf{transfer a variational autoencoder} (VAE~\cite{kingmaautoencodingvariationalbayes2014}) from the source to the target task and compute the \tbf{length of weight trajectory} on the manifold. We transfer the VAE in two ways, (i) by directly fitting the model on the target task, and (ii) by interpolating the task using a mixture distribution as described in~\cref{ss:uncoupled_transfer_distance}.

\subsection{Quantitative comparison of distance matrices}
\label{ss:quantitative}
Metrics are not unique. We would however still like to compare two task distances across various pairs of tasks. In addition to showing these matrices and drawing qualitative interpretations, we use the Mantel test~\cite{mantel1967detection} to accept/reject the null hypothesis that variations in two distance matrices are correlated. We will always compute \tbf{correlations with the fine-tuning distance matrix} because it is a practically relevant quantity and task distances are often designed to predict this quantity. We report $p$-values and the normalized test statistic $r = 1/(n^2-n-1) \sum_{i,j=1}^n (a_{ij} - \bar{a}) (b_{ij} - \bar{b})/(\s_a \s_b)$ where $a, b \in \reals^{n \times n}$ are distance matrices for $n$ tasks, $\bar{a}, \s_a$ denote mean and standard deviation of entries respectively. Numerical values of $r$ are usually small for all data~\cite{ApeHomePage,goslee2007ecodist} but the pair $(r, p)$ are a statistically sound way of comparing distance matrices; large $r$ with small $p$ indicates better correlation.

\subsection{Transferring between subsets of benchmark datasets}

\begin{figure}[htpb]
\centering
\begin{subfigure}[t]{0.25\linewidth}
\includegraphics[width=\linewidth]{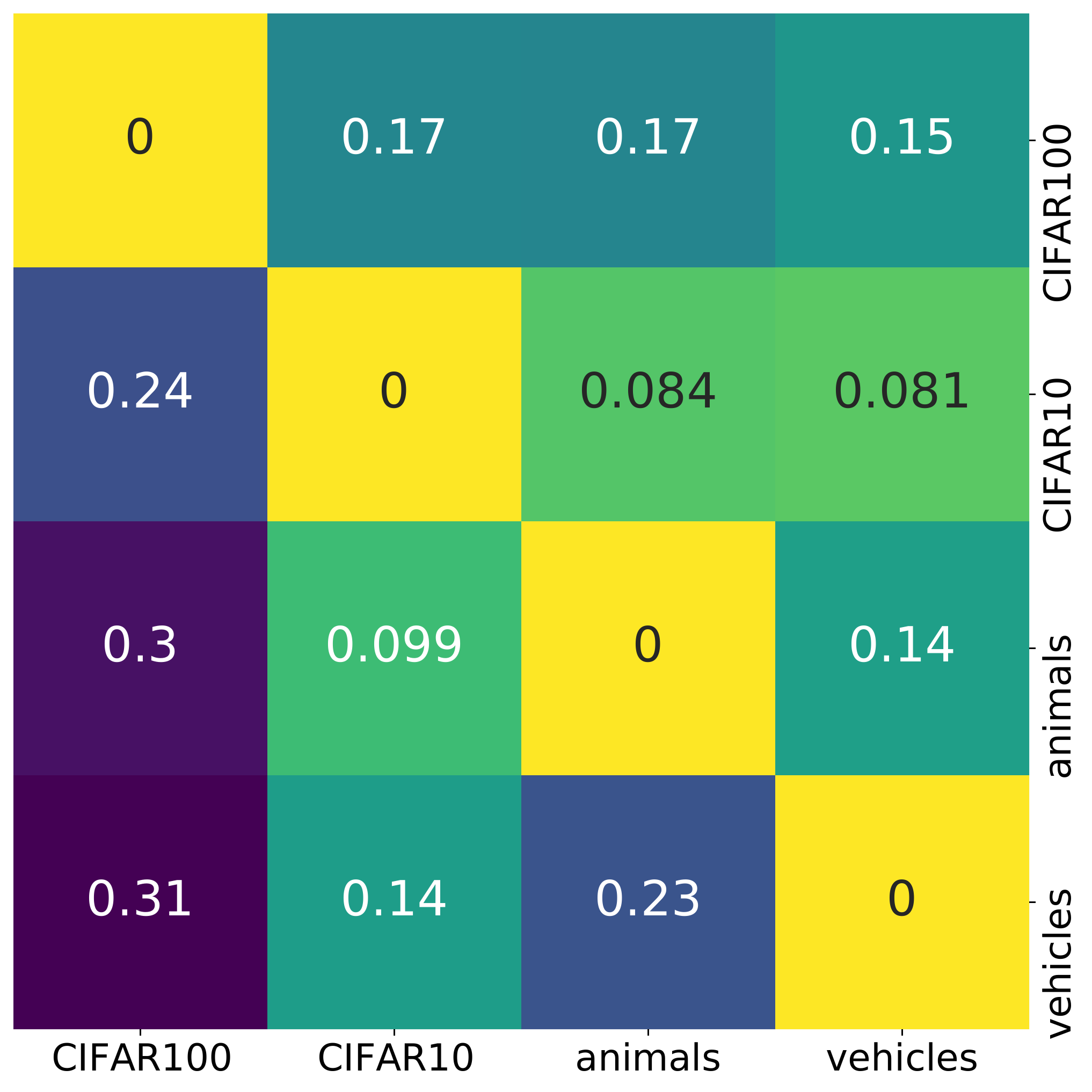}
\caption{}
\label{fig:CIFAR10}
\end{subfigure}
\begin{subfigure}[t]{0.25\linewidth}
\includegraphics[width=\linewidth]{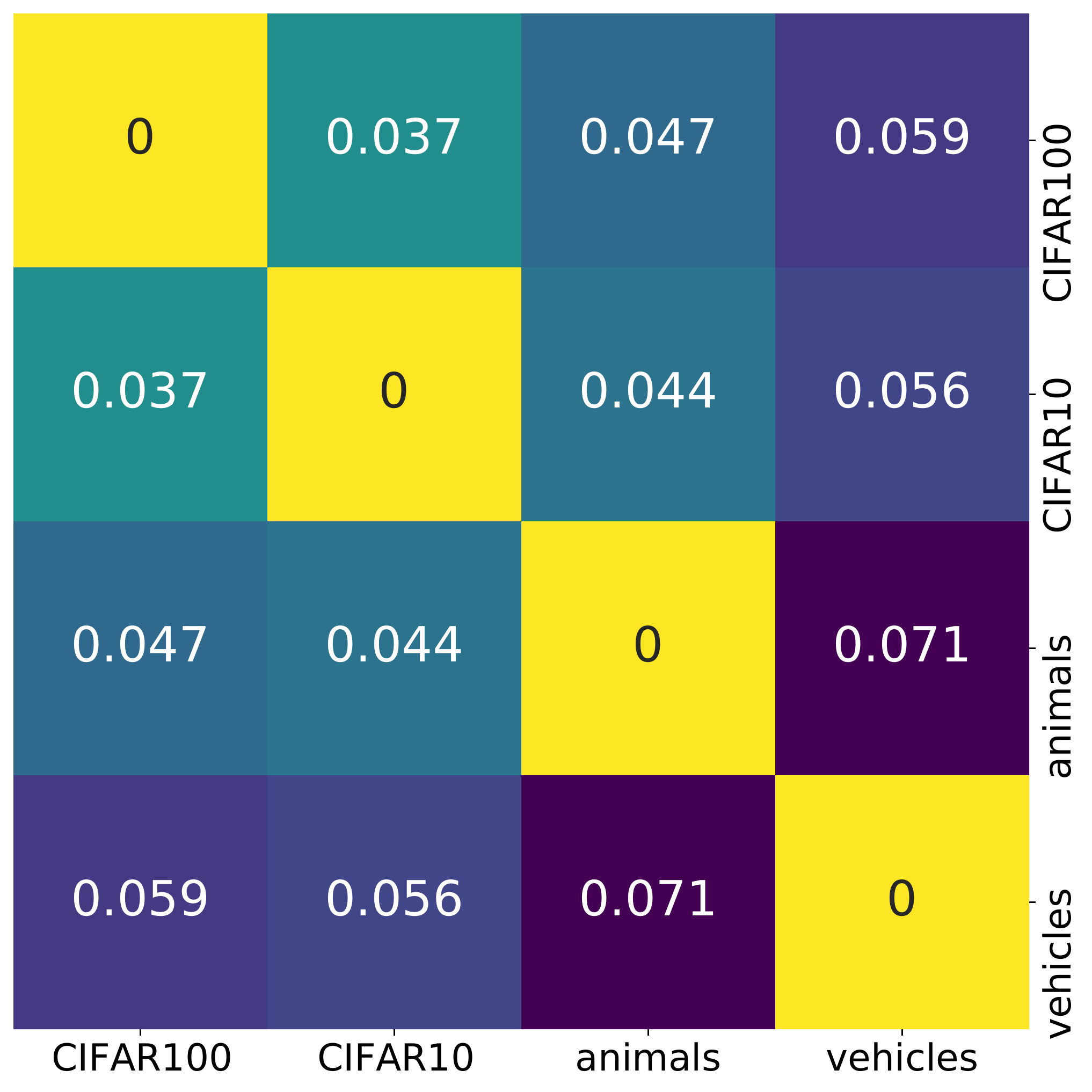}
\caption{}
\label{fig:CIFAR10-Task2Vec}
\end{subfigure}
\begin{subfigure}[t]{0.25\linewidth}
\includegraphics[width=\linewidth]{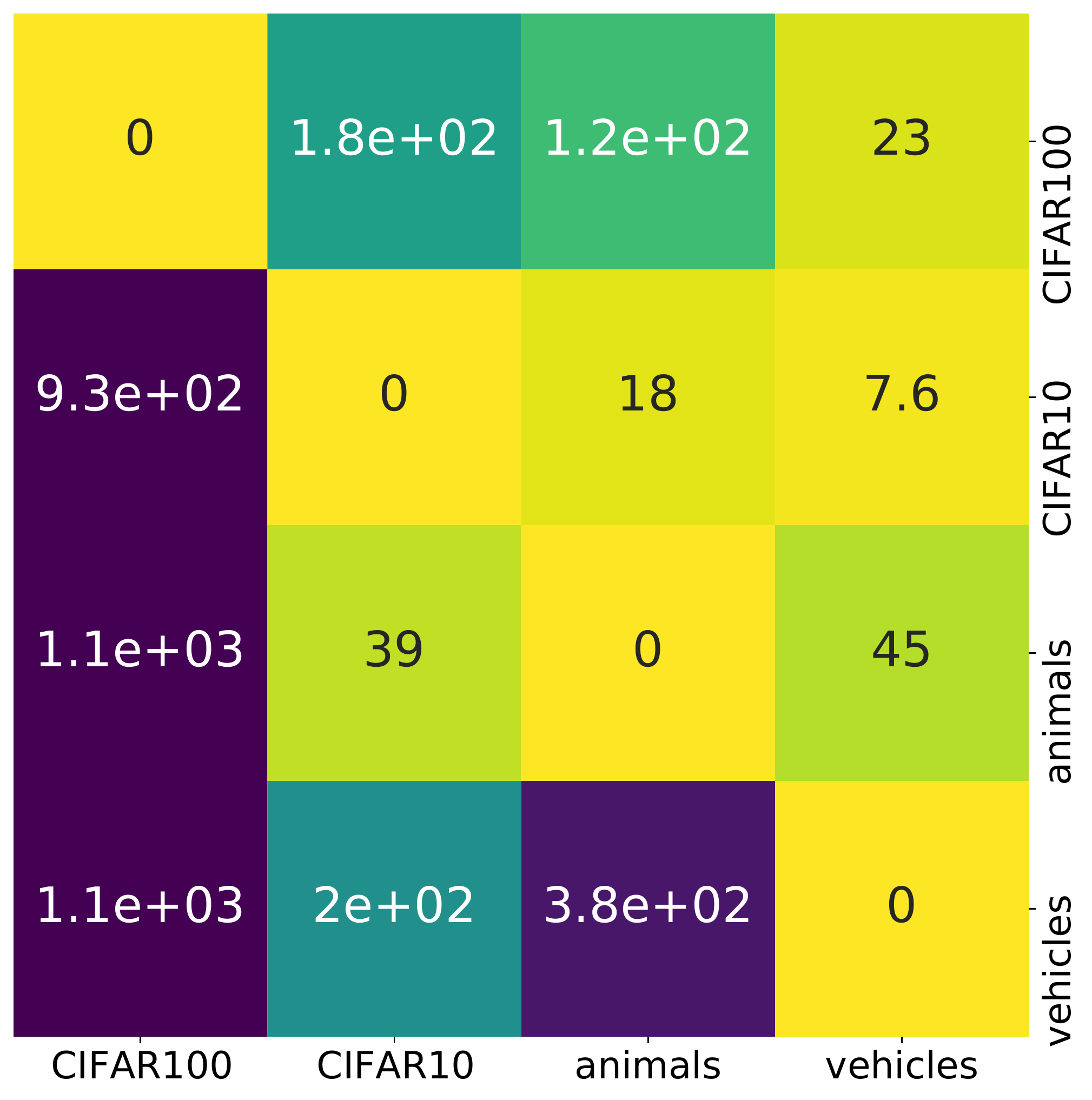}
\caption{}
\label{fig:CIFAR10_FineTune}
\end{subfigure}
\caption{
\cref{fig:CIFAR10} shows coupled transfer distance ($r$ = 0.428 $p$ = 0.13),
\cref{fig:CIFAR10-Task2Vec} shows distances estimated using Task2Vec ($r$ = 0.03, $p$ = 0.98),
\cref{fig:CIFAR10_FineTune} shows fine-tuning distance ($r$ = 0.61, $p$ = 0.09 with itself).
The numerical values of the distances in this figure are not
comparable with each other. Coupled transfer distances satisfy certain sanity
checks, e.g., transferring to a subset task is easier
than transferring from a subset task (CIFAR-10-vehicles/animals), which Task2Vec does not.
}
\label{fig:c10_c100}
\end{figure}

\heading{CIFAR-10 and CIFAR-100}
We consider four tasks (i) all vehicles (airplane, automobile, ship, truck) in CIFAR-10,
(ii) the remainder, namely six animals in CIFAR-10, (iii) the entire CIFAR-10 dataset and
(iv) the entire CIFAR-100 dataset. We show results in~\cref{fig:c10_c100}
using 4$\times$4 distance matrices where numbers in each cell indicate the distance between
the source task (row) and the target task (column).

Coupled transfer shows similar trends as fine-tuning, e.g., the tasks animals-CIFAR-10 or
vehicles-CIFAR-10
are close to each other while CIFAR-100 is far away from all tasks (it is closer to CIFAR-10
than others). Task distance is asymmetric in~\cref{fig:CIFAR10},~\cref{fig:CIFAR10_FineTune}.
Distance from CIFAR-10-animals is smaller than animals-CIFAR-10; this is
expected because animals is a subset of CIFAR-10.
Task2Vec distance estimates in~\cref{fig:CIFAR10-Task2Vec}
are qualitatively quite different from these two; the distance matrix is symmetric.
Also, while fine-tuning from animals-vehicles is relatively easy, Task2Vec
estimates the distance between them to be the largest.

This experiment also shows that our approach can scale to medium-scale datasets and can handle situations when the source and target task have different number of classes.

\begin{figure}[htpb]
\centering
\begin{subfigure}[t]{0.24\linewidth}
\includegraphics[width=\linewidth]{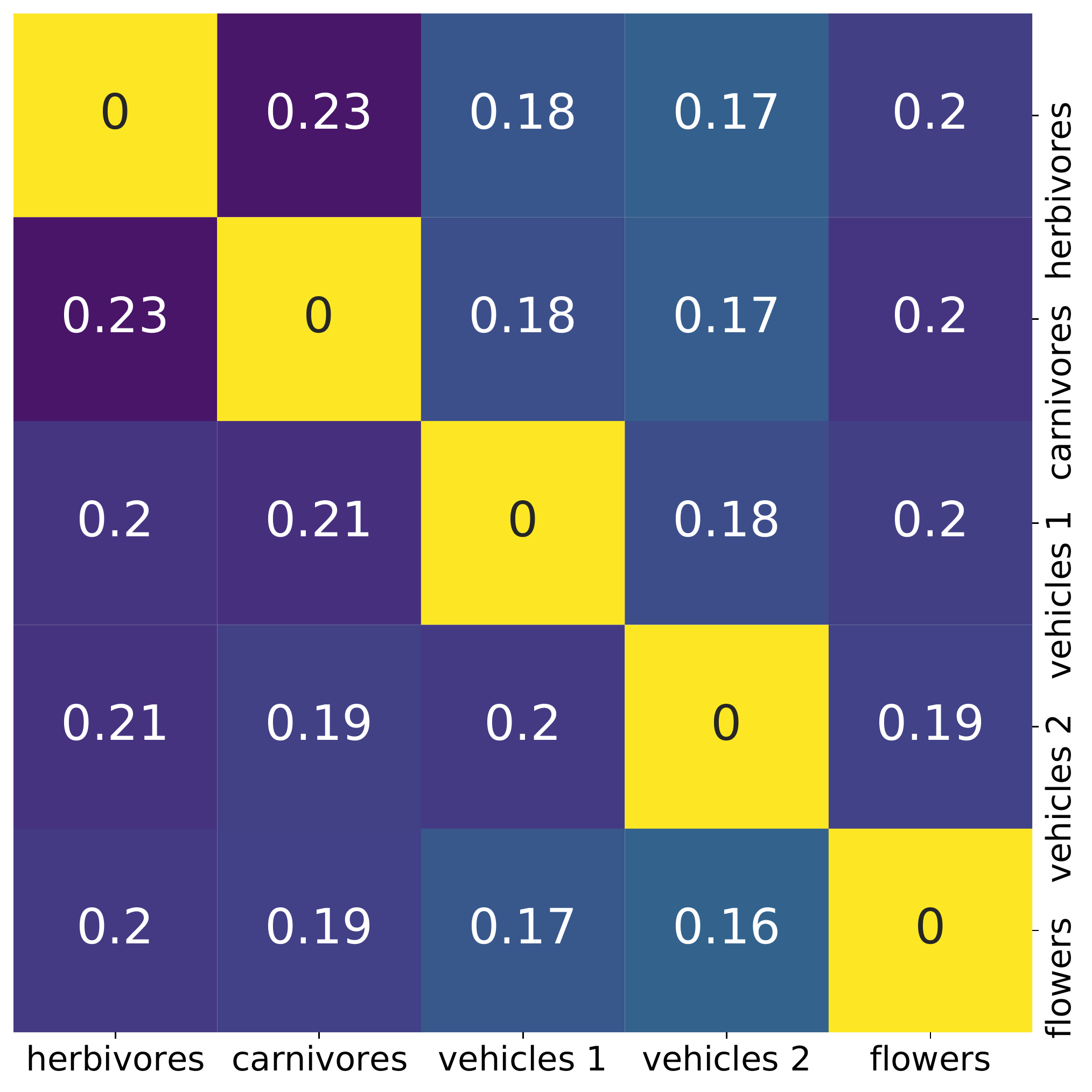}
\caption{}
\label{fig:CNN}
\end{subfigure}
\begin{subfigure}[t]{0.24 \linewidth}
\includegraphics[width=\linewidth]{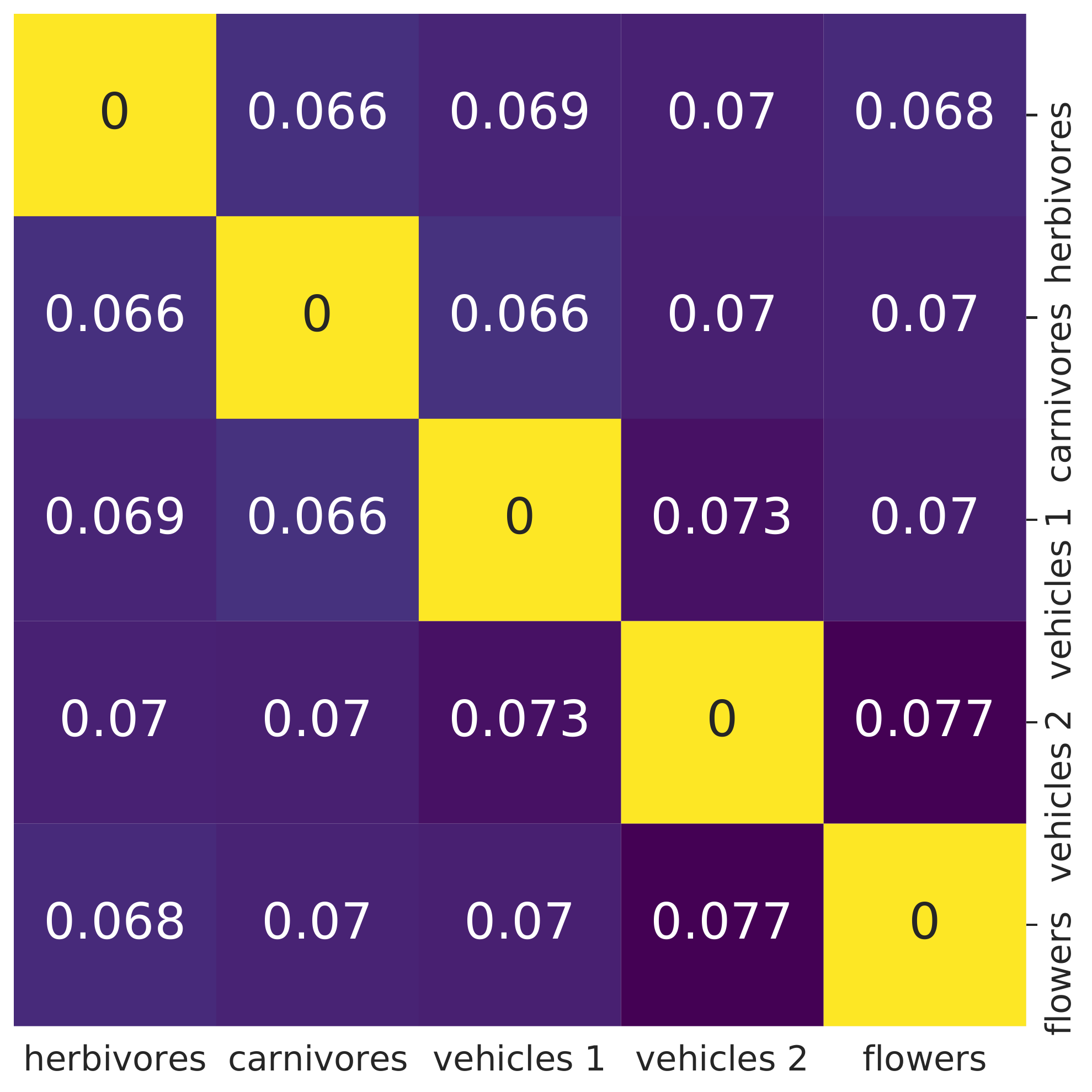}
\caption{}
\label{fig:Task2Vec}
\end{subfigure}
\begin{subfigure}[t]{0.24 \linewidth}
\includegraphics[width=\linewidth]{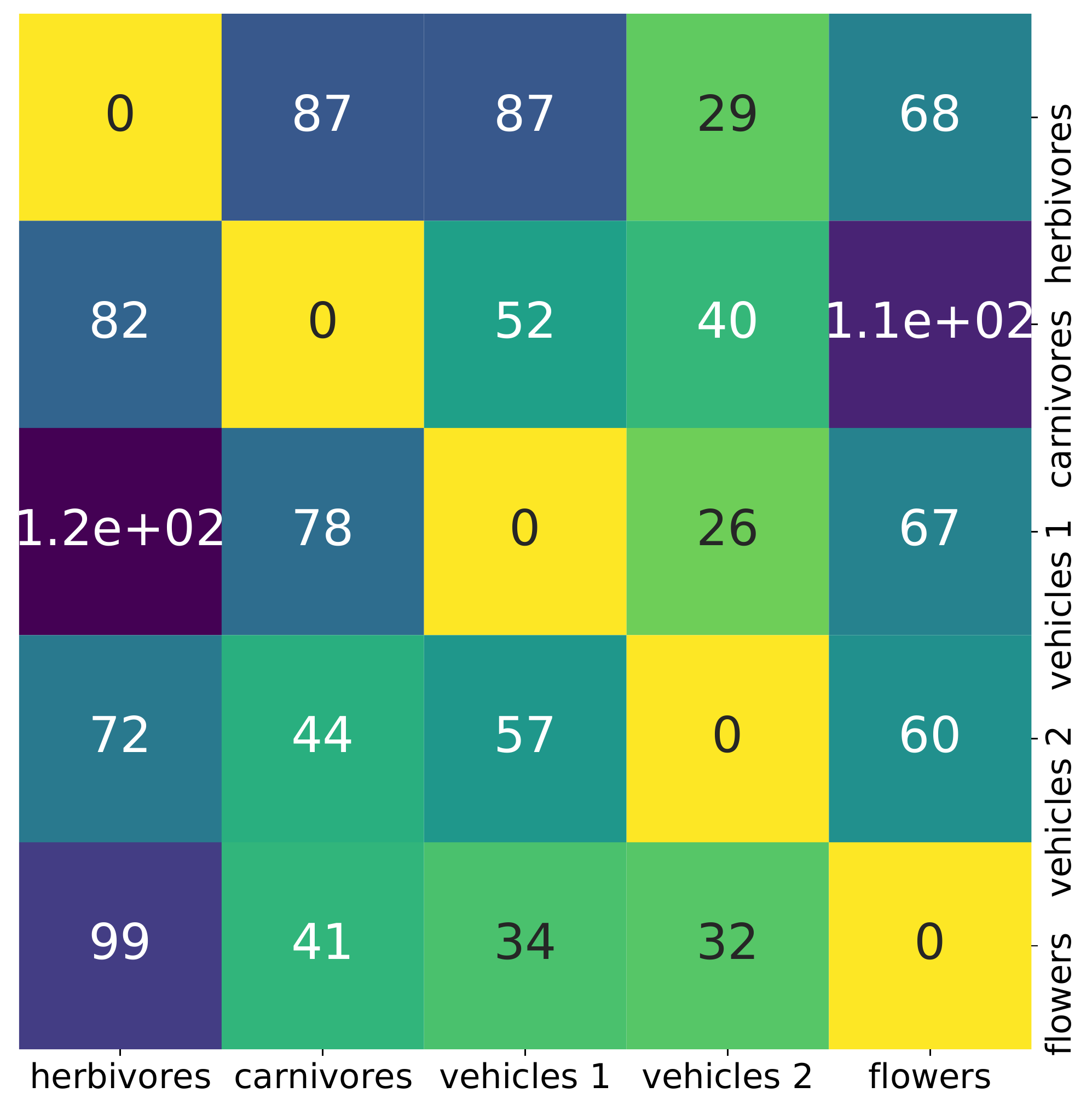}
\caption{}
\label{fig:CNN_FineTune}
\end{subfigure}
\begin{subfigure}[t]{0.24 \linewidth}
\includegraphics[width=\linewidth]{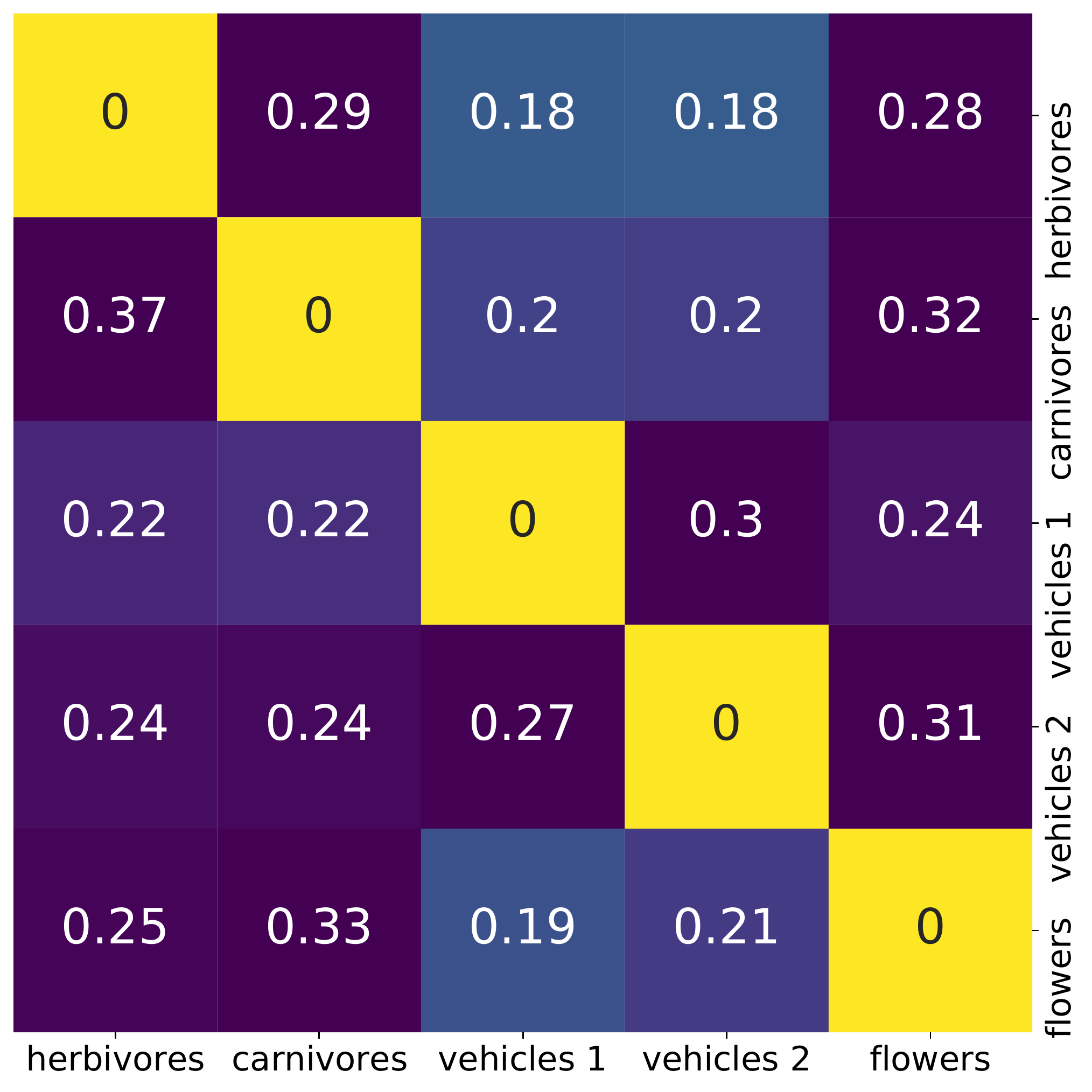}
\caption{}
\label{fig:CNN_uncouplings}
\end{subfigure}
\caption{\cref{fig:CNN} shows coupled transfer distance ($r$ = 0.14, $p$ = 0.05),
\cref{fig:Task2Vec} shows Task2Vec distance ($r$ = 0.07, $p$ = 0.17),
\cref{fig:CNN_FineTune} shows fine-tuning distance ($r$ = 0.36, $p$ = 0.03),
 and~\cref{fig:CNN_uncouplings} shows uncoupled transfer distance  ($r$ = 0.12, $p$ = 0.47).
Numerical values in the first and the last sub-plot can be compared
directly. Coupled transfer broadly agrees with fine-tuning except for
carnivores-flowers and herbivores-vehicles-1. For all tasks, uncoupled transfer
overestimates the distances compared to~\cref{fig:CNN}.
}
\label{fig:c100}
\end{figure}

\heading{Transferring between subsets of CIFAR-100}
We construct five tasks (herbivores, carnivores, vehicles-1, vehicles-2 and
flowers) that are subsets of the CIFAR-100 dataset. Each of these tasks consists
of 5 sub-classes. The distance matrices for coupled transfer, Task2Vec
and fine-tuning are shown in~\cref{fig:CNN},~\cref{fig:Task2Vec}
and~\cref{fig:CNN_FineTune} respectively. We also
show results using uncoupled transfer in~\cref{fig:CNN_uncouplings}.

Coupled transfer estimates that all these subsets of CIFAR-100 are roughly
equally far away from each other with herbivores-carnivores being the farthest
apart while vehicles-1-vehicles-2 being closest. This ordering is consistent
with the fine-tuning distance although fine-tuning results in an extremely large
value for carnivores-flowers and vehicles-1-herbivores. This ordering is mildly
inconsistent with the distances reported by Task2Vec in~\cref{fig:Task2Vec}
the distance for vehicles-1-vehicles-2 is the highest here. Broadly, Task2Vec
also results in a distance matrix that suggests that all tasks are equally far
away from each other.
As has been reported before~\citep{li2019rethinking},
this experiment also demonstrates the fragility of fine-tuning.

Recall that distances for uncoupled transfer
in~\cref{fig:CNN_uncouplings} can be compared directly
to those in~\cref{fig:CNN} for coupled transfer.
Task distances for the former are always larger.
Further, distance estimates of uncoupled transfer do not bear much
resemblance with those of fine-tuning; see for example
the distances for vehicles-2-carnivores, flowers-carnivores, and
vehicles-1-vehicles-2.
This demonstrates the utility of solving a coupled optimization problem in~\cref{eq:algorithm}
which finds a shorter trajectory on the statistical manifold.

Experiments on \tbf{transferring between subsets of Deep Fashion} are given
in~\cref{s:a:additional_expt}. We also computed task distances for tasks with different input domains. For transferring from \tbf{MNIST to CIFAR-10}, the coupled transfer distance is 0.18 (0.06 in the other direction), fine-tuning distance is 554.2 (20.6 in the other direction) and Task2Vec distance is 0.149 (same in the other direction). This experiment shows that can robustly handle diverse input domains and yet again, the coupled transfer distance correlates with the fine-tuning distance .

\subsection{Further analysis of the coupled transfer distance}

\begin{figure}[htpb]
\centering
\begin{subfigure}[t]{0.4 \linewidth}
\includegraphics[width=\linewidth]{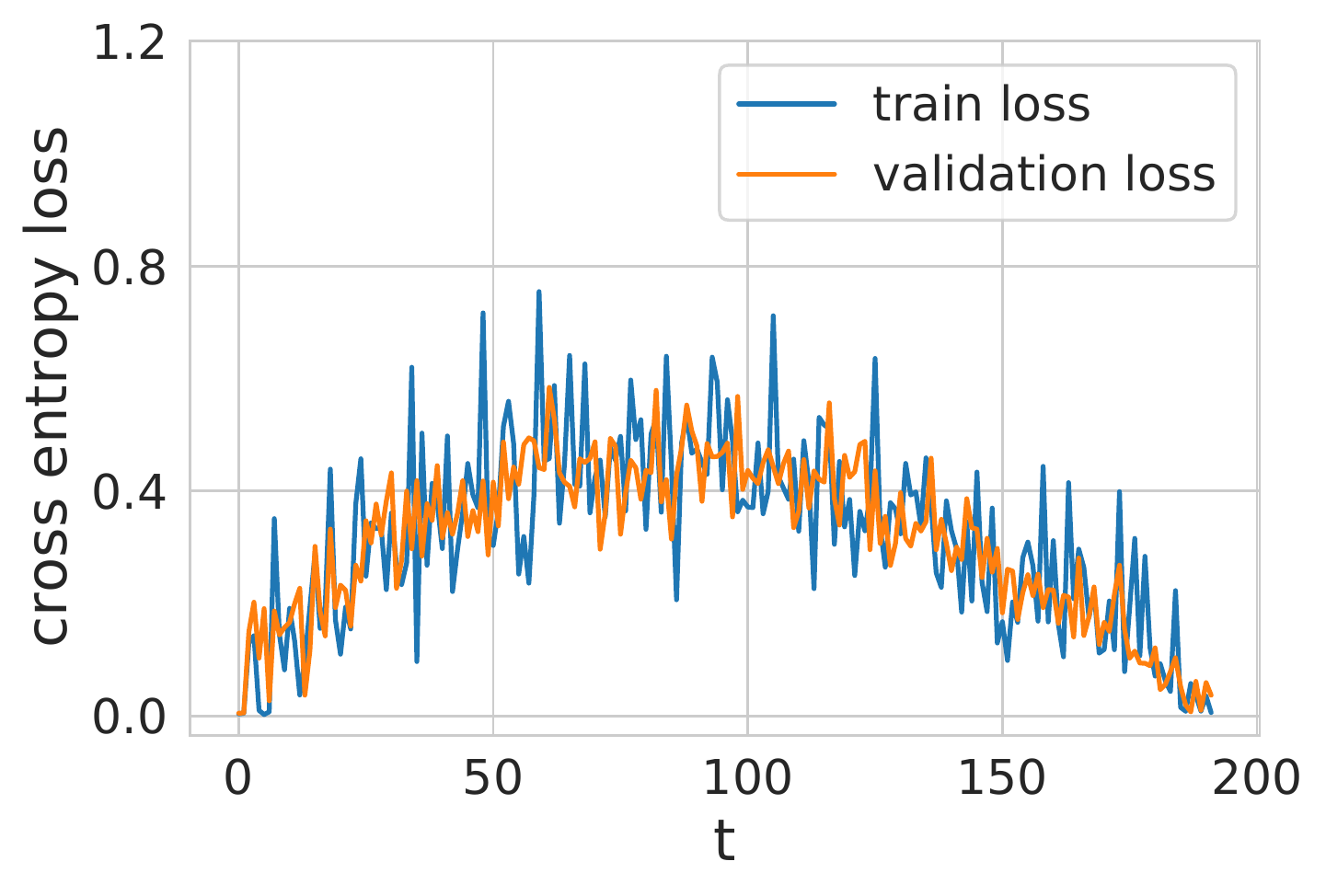}
\caption{}
\label{fig:Fig-4a}
\end{subfigure}
\begin{subfigure}[t]{0.4 \linewidth}
\includegraphics[width=\linewidth]{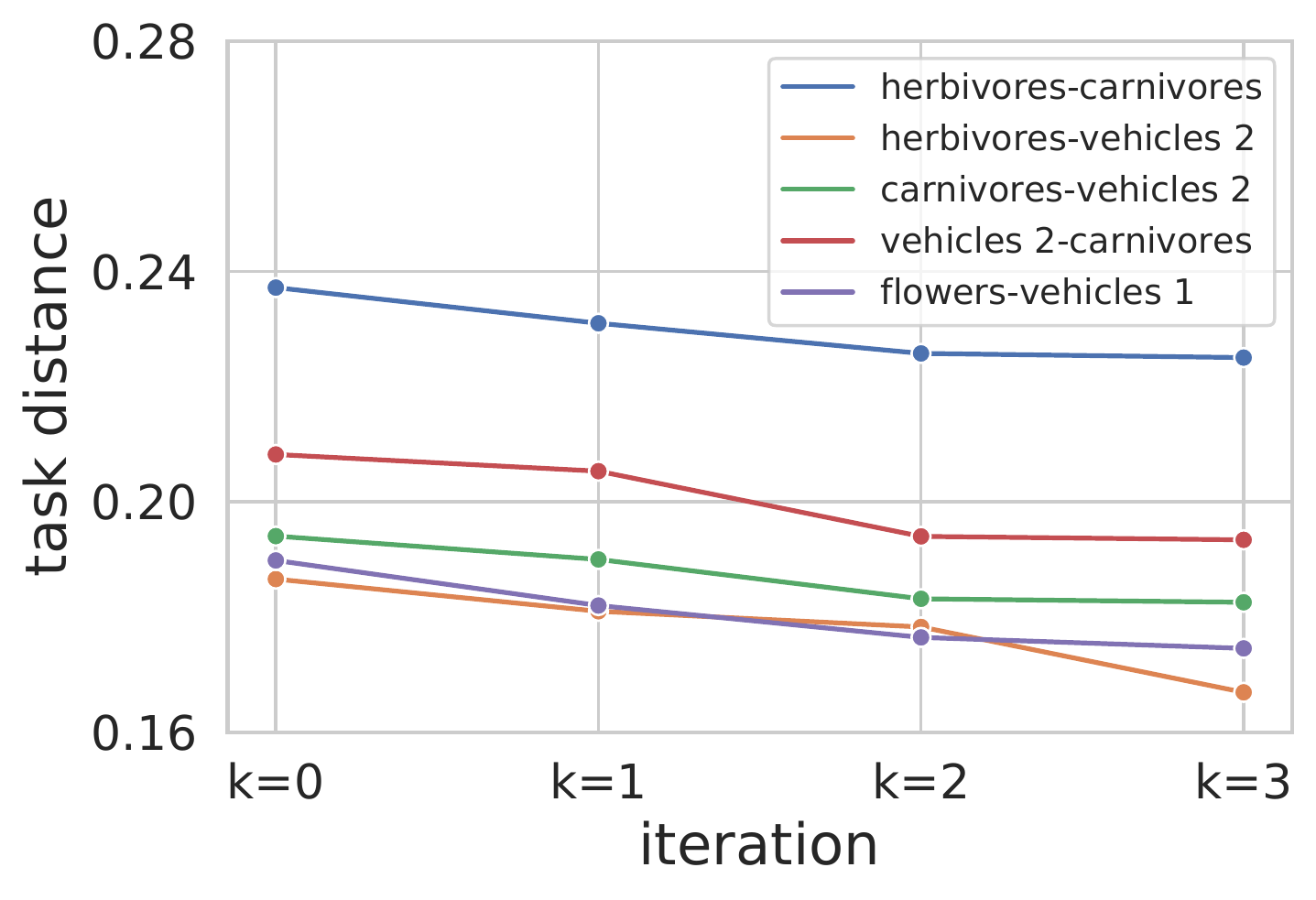}
\caption{}
\label{fig:Fig-4b}
\end{subfigure}
\caption{
\cref{fig:Fig-4a} shows the evolution of the training and test
cross-entropy loss on the interpolated distribution
as a function of the transfer steps in the
final iteration of coupled transfer of vehicles-1-vehicles-2.
As predicted by~\cref{thm:integral_of_fisher_dist_main},
generalization gap along the trajectory is small. \cref{fig:Fig-4b}
shows the convergence of the task distance with the number of iterations $k$
in~\cref{eq:algorithm}; the distance typically converges in 4--5 iterations for
these tasks.
}
\label{fig:loss}
\end{figure}

\heading{Convergence of coupled transfer}
\cref{fig:Fig-4a} shows the evolution of training and test loss as
computed on samples of the interpolated distribution after $k=4$ iterations
of~\cref{eq:algorithm}. As predicted by~\cref{thm:integral_of_fisher_dist_main}
the generalization gap is small throughout the trajectory. Training loss increases
towards the middle; this is expected because the interpolated
task is far away from both source and target tasks there. The interpolation~\cref{eq:interp_finite_algorithm}
could also be a cause for this increase.

We typically require 4--5
iterations of~\cref{eq:algorithm} for the task
distance to converge; this is shown in~\cref{fig:Fig-4b} for a few instances.
This figure also indicates that computing the transport coupling
in~\cref{eq:ot_finite}
independently of the weights and using this coupling
to modify the weights, as done in say~\citep{cui2018large},
results in a larger distance than if one were to optimize the couplings along
with the weights. The coupled transfer finds shorter trajectories for weights
and will potentially lead to better accuracies on target tasks in
studies like~\citep{cui2018large} because it samples more source data.

\begin{figure}[htpb]
\centering
\begin{subfigure}[t]{0.32 \linewidth}
\includegraphics[width=\linewidth]{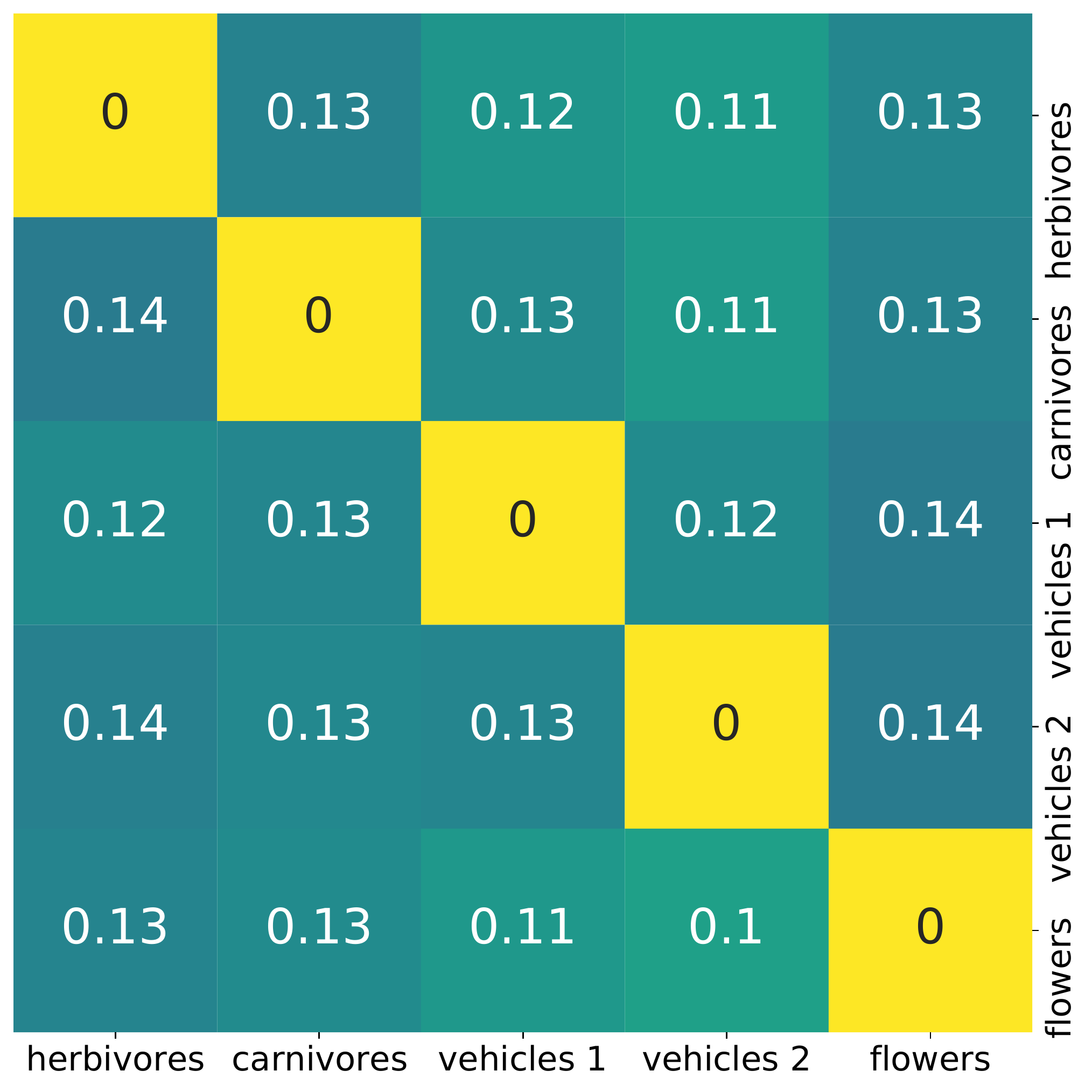}
\caption{}
\label{fig:wrn164_ours}
\end{subfigure}
\hspace{2em}
\begin{subfigure}[t]{0.32 \linewidth}
\includegraphics[width=\linewidth]{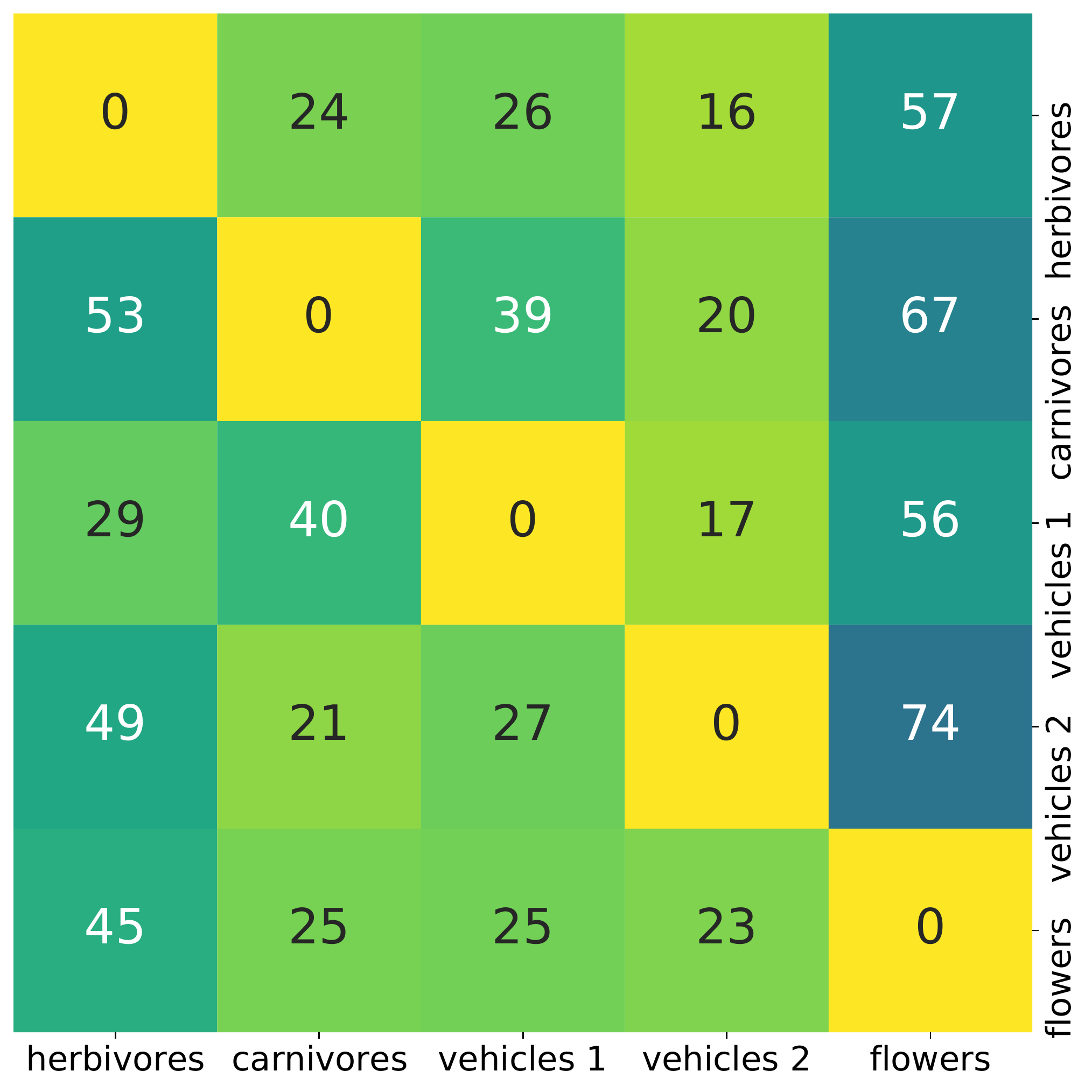}
\caption{}
\label{fig:wrn164_finetune}
\end{subfigure}
\caption{\cref{fig:wrn164_ours} shows coupled transfer distance ($r$ = 0.15, $p$ = 0.01)
and~\cref{fig:wrn164_finetune} shows fine-tuning distance  ($r$ = 0.39, $p$ = 0.01 with itself
and $r$ = 0.21, $p$ = 0.20 with fine-tuning distance in~\cref{fig:CNN_FineTune}).
Numbers in~\cref{fig:wrn164_ours} can be directly compared to those in~\cref{fig:CNN}.
WRN-16-4 model has a shorter trajectory for all task pairs
compared to the CNN in~\cref{fig:CNN} with fewer parameters.
}
\label{fig:wrn164}
\end{figure}

\heading{Models with a larger capacity are easier to transfer}
We next show that using a model with higher capacity results in smaller
distances between tasks. We consider a wide residual network
(WRN-16-4) of~\cite{zagoruyko2016wide} and compute distances on subsets of
CIFAR-100 in~\cref{fig:wrn164}. First note that task distances for
coupled transfer in~\cref{fig:wrn164_ours} are consistent
with those for fine-tuning in~\cref{fig:wrn164_finetune}.
Coupled transfer distances in~
\cref{fig:wrn164_ours}
are much smaller than those in~\cref{fig:CNN}.

Roughly speaking, a high-capacity model can learn a rich set of features,
some discriminative and others redundant not
relevant to the source task. These redundant features are useful
if target task is dissimilar to the source.
This experiment also demonstrates that the information-geometric distance
computed by coupled transfer, which is independent of the dimension of the statistical manifold,
leads to a constructive strategy for selecting architectures for transfer learning. Most methods to compute task distances instead only inform which source target is best suited to pre-train with for the target task.

\begin{figure}[htpb]
\centering
\includegraphics[width=\linewidth]{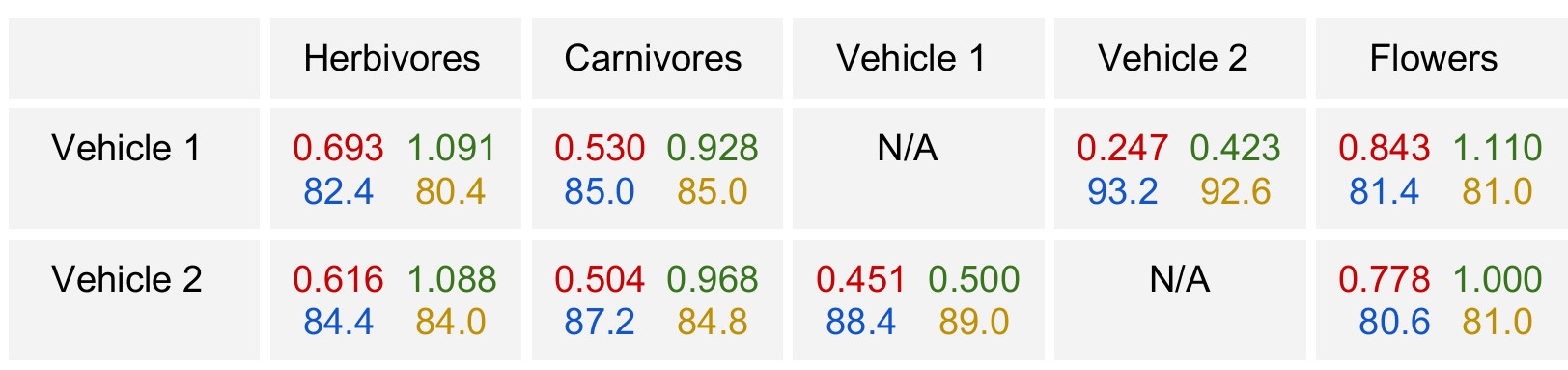}
\vskip 5pt
\caption{Comparison of validation loss (red for coupled transfer, green for fine-tuning) and accuracy (\%) (blue and yellow respectively) between different subsets of CIFAR-100. Optimal transport for the task distribution results in large improvements in the validation loss in all cases; The validation accuracy also improve by 0.4\%--2.5\% in all cases except the last two.}
\label{fig:final_accuracy}
\end{figure}

\tbf{Does coupled transfer lead to better generalization on the target?}
It is natural to ask whether the generalization performance of the model after coupled transfer is better than the one after standard fine-tuning (which does not transport the task). \cref{fig:final_accuracy} compares the validation loss and the validation accuracy after coupled transfer and after standard fine-tuning for pairs of CIFAR-100 tasks. It shows that broadly, the former improves generalization. This is consistent with existing literature~\cite{gao2020free} which employs task interpolation for better transfer. Let us note that improving fine-tuning is not our goal while developing the task distance. In fact, we want the task distance to correlate with the difficulty of fine-tuning.

\begin{figure}[htpb]
\centering
\begin{subfigure}[t]{0.24\linewidth}
\includegraphics[width=\linewidth]{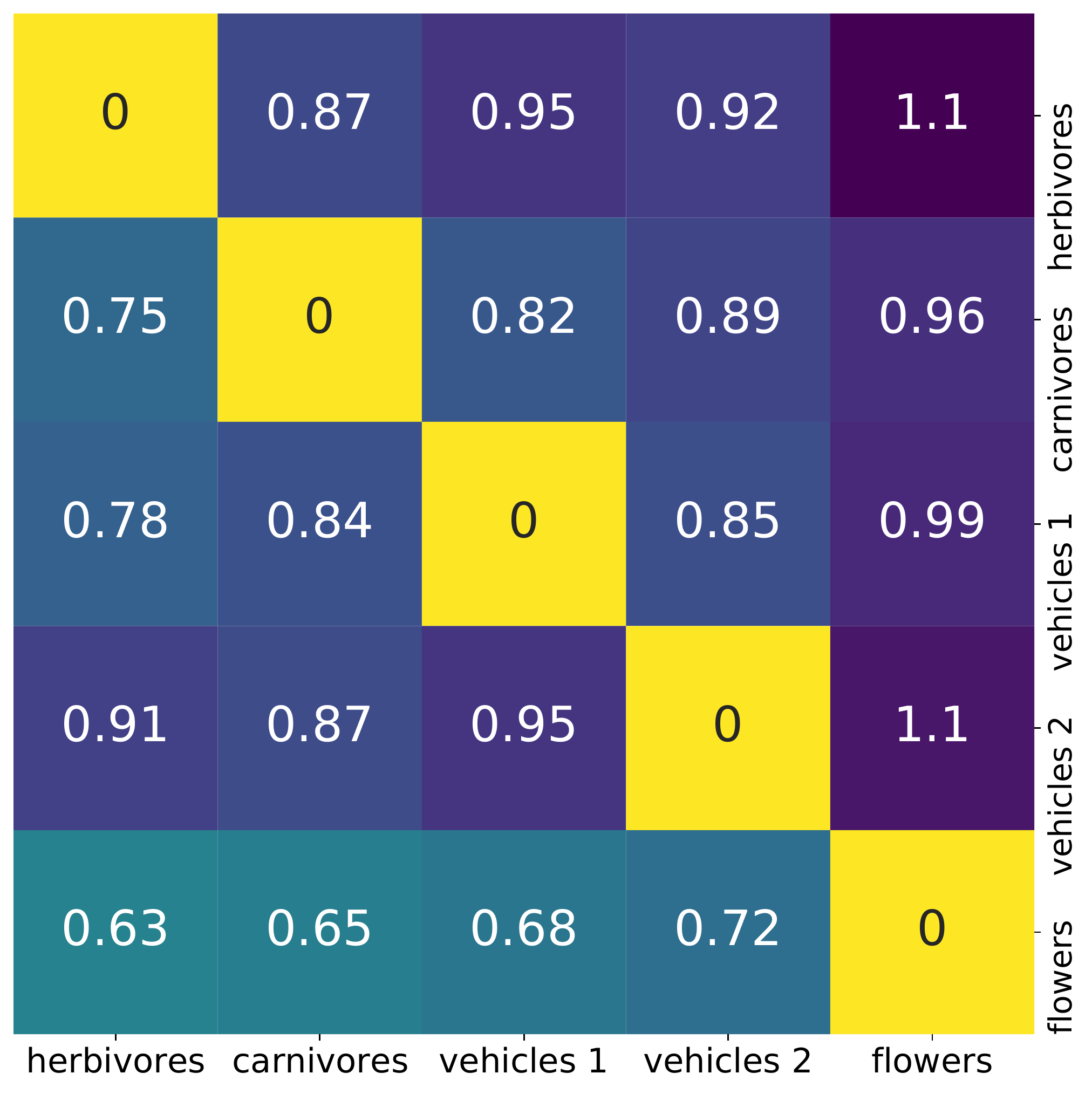}
\caption{}
\label{fig:vae_uncoupled}
\end{subfigure}
\begin{subfigure}[t]{0.24\linewidth}
\includegraphics[width=\linewidth]{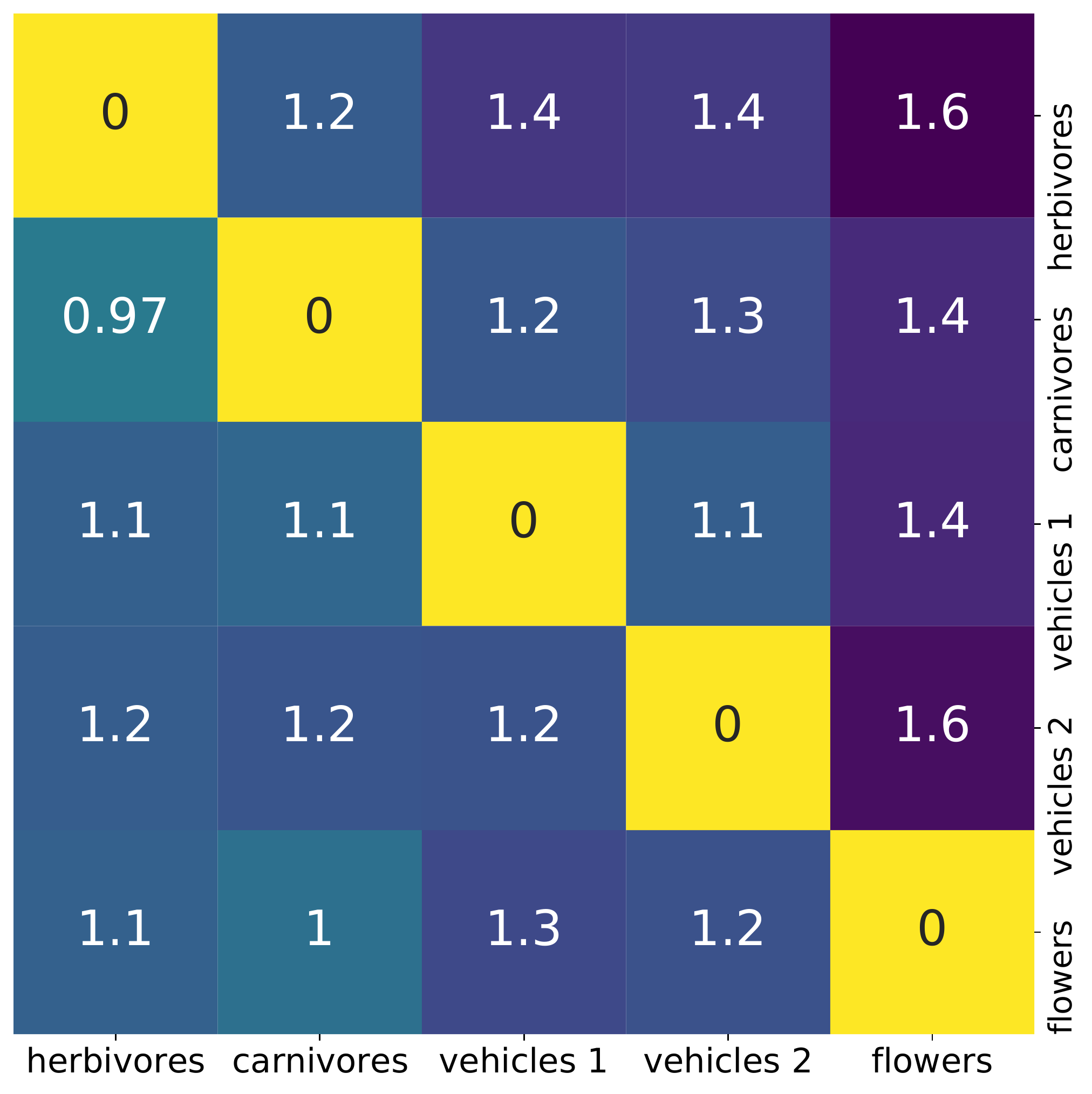}
\caption{}
\label{fig:vae_distance_c100}
\end{subfigure}
\begin{subfigure}[t]{0.24\linewidth}
\includegraphics[width=\linewidth]{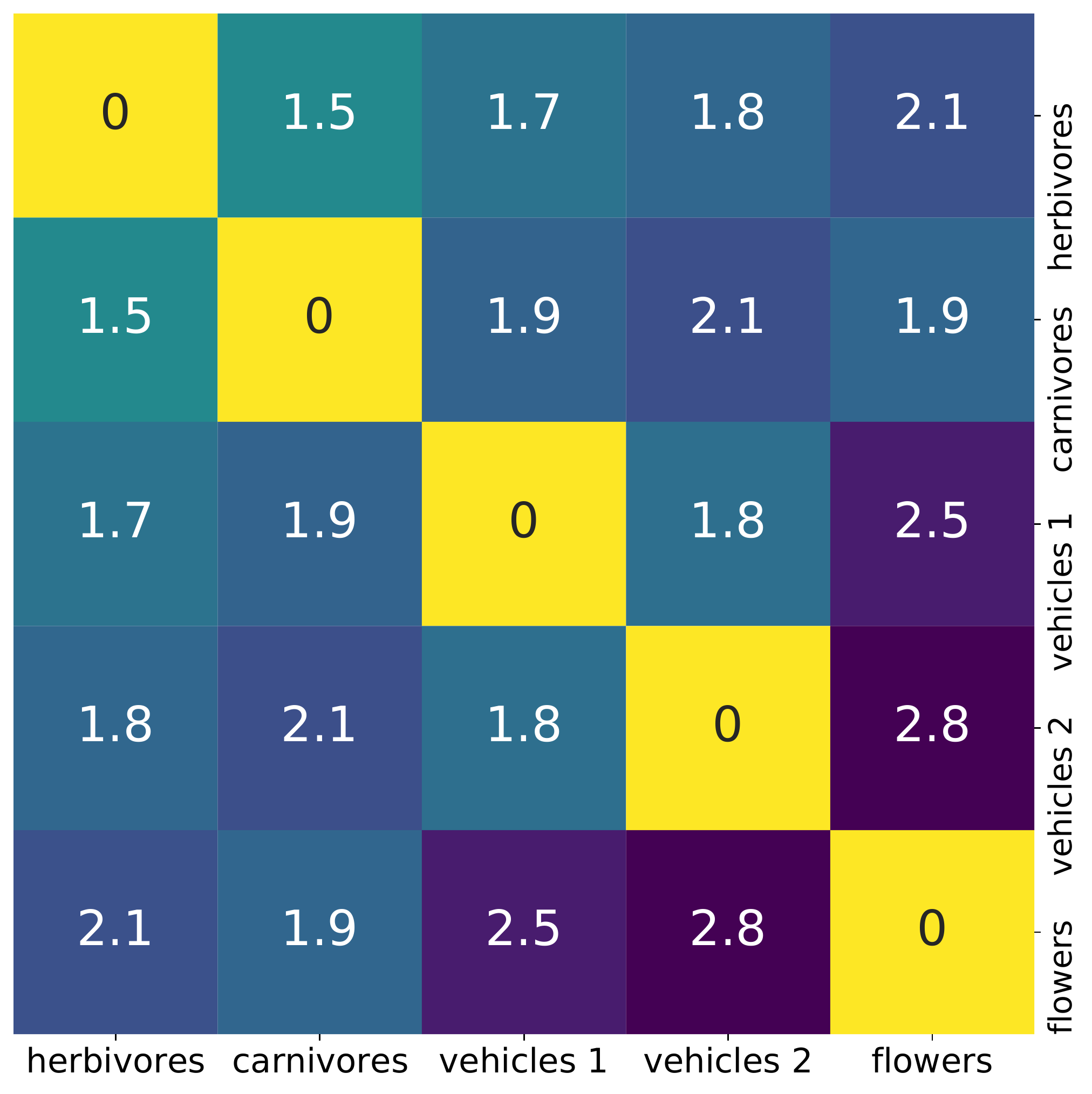}
\caption{}
\label{fig:w2_distance_c100}
\end{subfigure}
\begin{subfigure}[t]{0.24\linewidth}
\includegraphics[width=\linewidth]{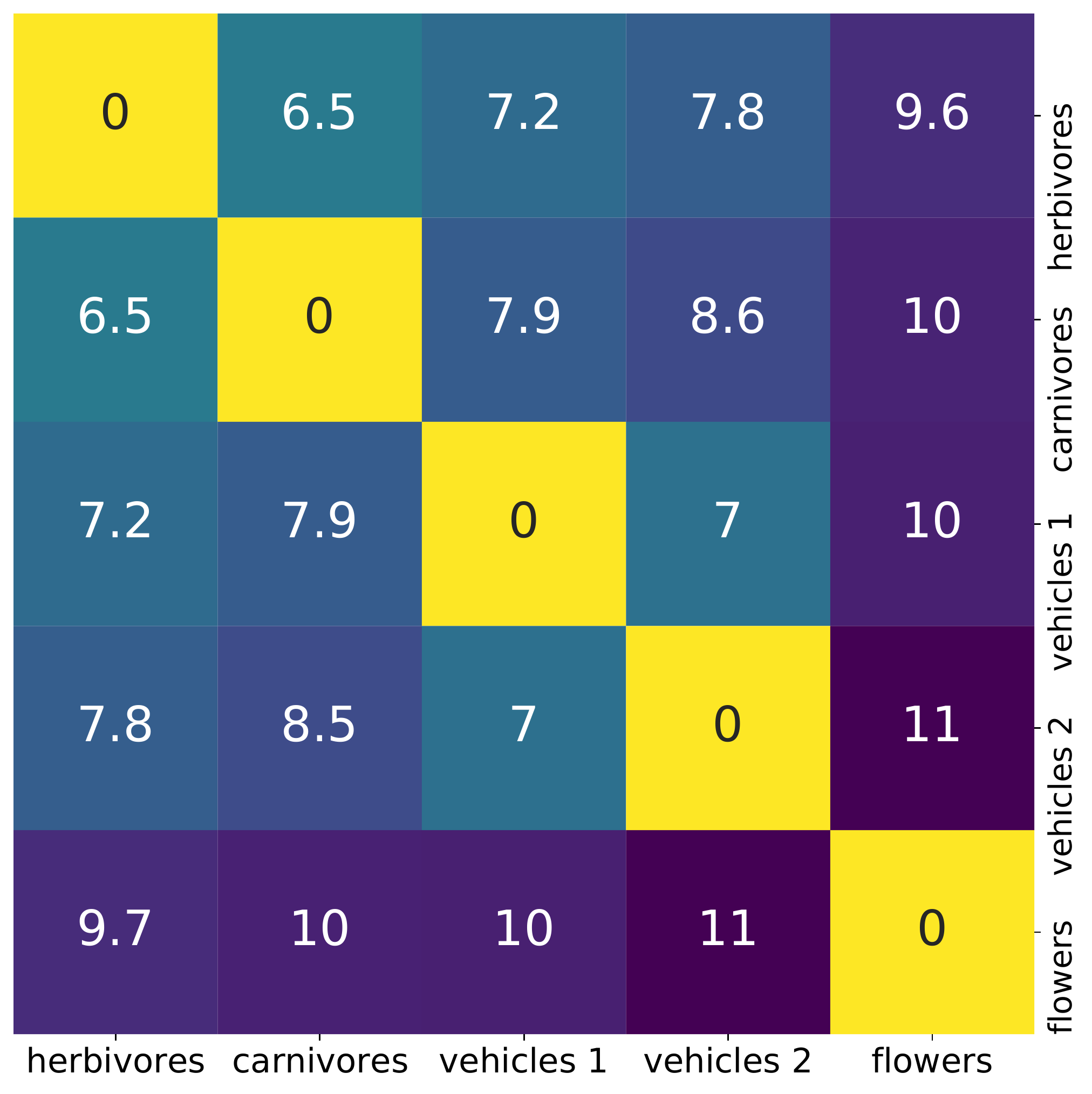}
\caption{}
\label{fig:w2_embedding_distance_c100}
\end{subfigure}
\caption{
\cref{fig:vae_uncoupled} shows task distance computed using the Riemannian length of the weight trajectory for the VAE using a mixture distribution to interpolate the tasks (see~\cref{ss:expt:setup}, $r$ = 0.1, $p$ = 0.76),
\cref{fig:vae_distance_c100} shows the same quantity for directly fine-tuning the VAE ($r$ = 0.09, $p$ = 0.88), \cref{fig:w2_distance_c100} shows task distance using the Wasserstein metric on the pixel-space ($r$ = 0.02, $p$ = 0.22), \cref{fig:w2_embedding_distance_c100} shows distances using Wasserstein metric on the embedding space ($r$ = 0.06, $p$ = 0.40). The last three methods agree with each other very well (see the narrative for $p$-values) but small Mantel test statistic and high $p$-values as compared to~\cref{fig:CNN_FineTune} indicates that these distances are not correlated with the difficulty of fine-tuning.
}
\label{fig:embedding_distance}
\end{figure}

\heading{Comparison with other task discrepancy measures}
\cref{{fig:vae_uncoupled}} shows task distances computed using the Riemannian length of the weight trajectory for the VAE (see~\cref{ss:baselines}) when task is interpolated using a mixture distribution, \cref{fig:vae_distance_c100} shows the same quantity when the VAE is directly fitted to the target task after initialization on the source. \cref{fig:w2_distance_c100} and~\cref{fig:w2_embedding_distance_c100} show the Wasserstein distance on the pixel-space and feature-space respectively. We find that although the four distance matrices in~\cref{fig:embedding_distance} agree with each other very well ($r \approx$ 0.15, $p$ < 0.08 for all pairs, except the VAE with uncoupled transfer), they are very different from the fine-tuning distance in~\cref{fig:CNN_FineTune}. This shows that task distances computed using discrepancy measures on the input space are not reflective of the difficulty of fine-tuning, after all images in these tasks are visually quite similar to each each. Coupled transfer distance explicitly takes the hypothesis space into account and correctly reflects the difficulty of transfer, even if the input spaces are similar.

\section{Related Work}
\label{s:related_work}

\heading{Domain-specific methods}
A rich understanding of task distances has been developed
in computer vision, e.g.,~\citet{zamir2018taskonomy}
compute pairwise distances when different tasks such as classification,
segmentation etc. are performed on the same input data. The goal
of this work, and others such as~\citep{cui2018large},
is to be able to decide
which source data to pre-train to generalize well on a target task.
Task distances have also been widely discussed in the
multi-task learning~\citep{caruana1997multitask} and
meta/continual-learning~\citep{liu2019localized,pentina2014pac,hsu2018unsupervised}.
The natural language processing literature also prevents several
methods to compute similarity between input
data~\citep{mikolov2013efficient,pennington2014glove}.

Most of the above methods are
based on evaluating the difficulty of fine-tuning,
or computing the similarity in some embedding space.
It is difficult to ascertain whether the distances obtained thereby
are truly indicative of the difficulty of transfer; fine-tuning
hyper-parameters often need to be carefully chosen~\citep{li2019rethinking}
and neither is the embedding space unique.
For instance, the uncoupled transfer process that modifies the input data
distribution will lead to a different estimate of task distance.

\heading{Information-theoretic approaches}
We build upon a line of work that combines
generative models
and discriminatory classifiers (see~
\citep{jaakkola1999exploiting,perronnin2010improving}
to name a few) to construct a notion of similarity
between input data.
Modern variants of this
idea include Task2Vec~\citep{achille2019task2vec} which embeds the task
using the diagonal of the FIM and computes distance
between tasks using the cosine distance for this embedding.
The main hurdle in Task2Vec and similar approaches is to design
the architecture for computing FIM: a
small model will indicate that tasks are far away.
\citet{achilleDynamicsReachabilityLearning2019,achilleInformationComplexityLearning2019}
use the KL
divergence between the posterior weight distribution and a prior to
quantify the complexity of a task; distance
between tasks is defined to be the increase in complexity when
the target task is added to the source task. This is an elegant
formalism but it is challenging to compute it accurately and it has not yet been demonstrated for a broad range of datasets.

\heading{Learning-theoretic approaches}
Learning theory typically studies out-of-sample performance on a single task
using complexity measures such as VC-dimension~\citep{vapnik1998statistical}.
These have been adapted to address the difficulty of domain adaptation~\citep{ben2010theory,zhang2012generalization,redko2019advances} which gives a measure of task distance that incorporates the complexity of the hypothesis space. In particular, \citet{ben2010theory} train on a fixed
mixture of the source and target data to minimize which is similar to our interpolated
distribution~\cref{eq:interp_finite_algorithm}.
Theoretical results here corroborate (actually motivate) our experimental result that transferring between the same tasks with a higher-capacity model is easer. A key gap in this literature is that this theory does not consider \emph{how} the model is adapted to target task. For complex models such as deep networks, hyper-parameters during fine-tuning play a crucial role~\citep{li2019rethinking}. Our work fundamentally exploits the idea that the task need not be fixed during transfer, it can also be adapted. Further, our coupled transfer distance is invariant to the particular parametrization of the deep network, which is difficult to achieve using classical learning theory techniques.

\heading{Coupled transfer of data and the model}
Transporting the task using optimal transport is fundamental to how our
coupled transfer distance is defined. This is motivated from two recent studies.
\citet{gao2020free} develop an algorithm that keeps
the classification loss unchanged across transfer. Their
method interpolates between the source and target data
using the mixture distribution from~\cref{ss:uncoupled_transfer_distance}. We
take this idea further and employ optimal transport~\cite{cui2018large}
to modulate the interpolation of the task using the Fisher-Rao distance.
Coupled transport problems on the input data
are also solved for unsupervised
translation~\citep{alvarez-melisGromovWassersteinAlignmentWord2018}. The idea of modifying the task
during transfer using optimal transport is also exploited by~\citet{alvarez2020geometric}
to prescribe task distances and for data augmentation/interpolation and transfer~\citep{alvarez2020gradient}.

\section{Discussion}
\label{s:discussion}

Our work is an attempt to theoretically understand when transfer is easy and when it is not. An often over-looked idea in large-scale transfer learning is that the task need not remain fixed to the target task during transfer. We heavily exploit this idea in the present paper. We develop a ``coupled transfer distance'' between tasks that computes the shortest weight trajectory in information space, i.e., on the statistical manifold, while the task is optimally transported from the source to the target. The most important aspect of our work is that both task and weights are modified synchronously. It is remarkable that this coupled transfer distance is not just strongly correlated with the difficulty of fine-tuning but also theoretically captures the intuitive idea that a good transfer algorithm is the one that keeps generalization gap small during transfer, in particular at the end on the target task.

\begin{footnotesize}
\bibliography{main}
\bibliographystyle{icml2021}
\end{footnotesize}

\onecolumn

\begin{appendices}

\section{Details of the experimental setup}
\label{s:a:expt}

\subsection{Architecture and training.}
\label{ss: arch}
We show results using an 8-layer
convolutional neural network with ReLU nonlinearities, dropout,
batch-normalization with a final fully-connected layer. The larger model
used for experiments in~\cref{fig:wrn164} is a wide-residual-network
(WRN-16-4 architecture of~\citep{zagoruyko2016wide}).

\subsection{Transferring between CIFAR-10 and CIFAR-100}
\label{ss:cifar10_app}
We consider four tasks: (i) all vehicles (airplane, automobile, ship, truck) in CIFAR-10, consisting of 20,000 32$\times$32-sized RGB images; (ii) the remainder, namely six animals in CIFAR-10, consisting of 30,000 32$\times$32-sized RGB images; (iii) the entire CIFAR-10 dataset and
(iv) the entire CIFAR-100 dataset, consisting of 50,000 images and spread across 100 classes.

We pre-train model on source tasks using stochastic gradient descent (SGD) for 60 epochs, with mini-batch size of 20, learning rate schedule is set to $10^{-3}$ for epochs 0 -- 40 and $8 \times 10^{-4}$ for epochs 40 -- 60. When CIFAR-100 is the source dataset, we train for 180 epochs with the learning rate set to $10^{-3}$ for epochs 0 -- 120, and $8 \times 10^{-4}$ for epochs 120 -- 180.

We chose a slightly smaller version of the source and target datasets to compute the distance,
each of them have 19,200 images. The class distribution on all source and target classes is
balanced. We did this to reduce the size of the coupling matrix $\G$ in~\cref{eq:gkp}. The coupling
matrix connecting inputs in the source and target datasets is
$\G \in \reals^{19200 \times 19200}$ which is still quite
large to be tractable during optimization. We therefore use a block diagonal approximation of the coupling matrix;
640 blocks are constructed each of size 30$\times$30
and all other entries in the coupling matrix are set to zero at the beginning of each iteration in~\cref{eq:gkp} after computing the dense coupling matrix using the linear program. This effectively entails that the set of couplings over which we compute the transport is not the full convex polytope in~\cref{ss:transport} but rather
a subset of it.
We sample a mini-batch of 20 images from the interpolated distribution corresponding to this block-diagonal
coupling matrix for each weight update of~\cref{eq:sgd_prox}. We run 40 epochs, i.e., with 19200/20 = 960 weight updates per epoch for computing the weight trajectory at \emph{each iteration} $k$ in~\cref{eq:algorithm}.
The learning rate is fixed to $8 \times 10^{-4}$ in the transfer learning phase.

\subsection{Transferring among subsets of CIFAR-100}
\label{ss:cifar100_app}

The same 8-layer convolutional network is used to show results for transfer between subsets of CIFAR-10 and CIFAR-100. CIFAR-10 is split into the two tasks animals and vehicle again. We construct five tasks (herbivores, carnivores, vehicles-1,vehicles-2 and flowers) that are subsets of the CIFAR-100 dataset. Each of these tasks consists of 5 sub-classes.

We train the model on the source task using SGD for 400 epochs with a mini-batch size of 20. Learning rate is set to $10^{-3}$ for epochs 0 -- 240, and to $8 \times 10^{-4}$ for epochs 240 -- 400.

Tasks that are subsets of CIFAR-100 in the experiments in this section have few samples (2500 each) so we select
2400 images from source and target datasets respectively; we could have chosen a larger source dataset when
transferring from CIFAR-10 animals or
vehicles but we did not so for sake of simplicity. The number 2400 was chosen to make the block diagonal approximation
of the coupling matrix have 120$\times$120 entries in each block; this was constrained by the GPU memory. The coupling
matrix $\G$ therefore has 2400$\times$2400 entries with 20 blocks on the diagonal.

Again, we use a mini-batch size of 20 for 240 epochs (2400/20 = 120 weight updates per epoch) during the transfer
from the source dataset to the target dataset. The learning rate is fixed to $8 \times 10^{-4}$ in the transfer learning phase.

\subsection{Training setup for wide residual network}
\label{ss:wide_app}

We pre-train WRN-16-4 on source tasks using SGD for 400 epochs with a mini-batch size of 20.
Learning rate is $10^{-1}$ for epochs 0 -- 120, $2 \times 10^{-2}$ for epochs 120 -- 240,
$4 \times 10^{-3}$ for epochs 240--320, and $8 \times 10^{-4}$ for epochs 320 -- 400.
Other experimental details are the same as those in~\cref{ss:cifar100_app}.

\section{Experiments on the Deep Fashion dataset}
\label{s:a:additional_expt}

\begin{figure}[!htpb]
\centering
\begin{subfigure}[t]{0.2 \linewidth}
\includegraphics[width=\linewidth]{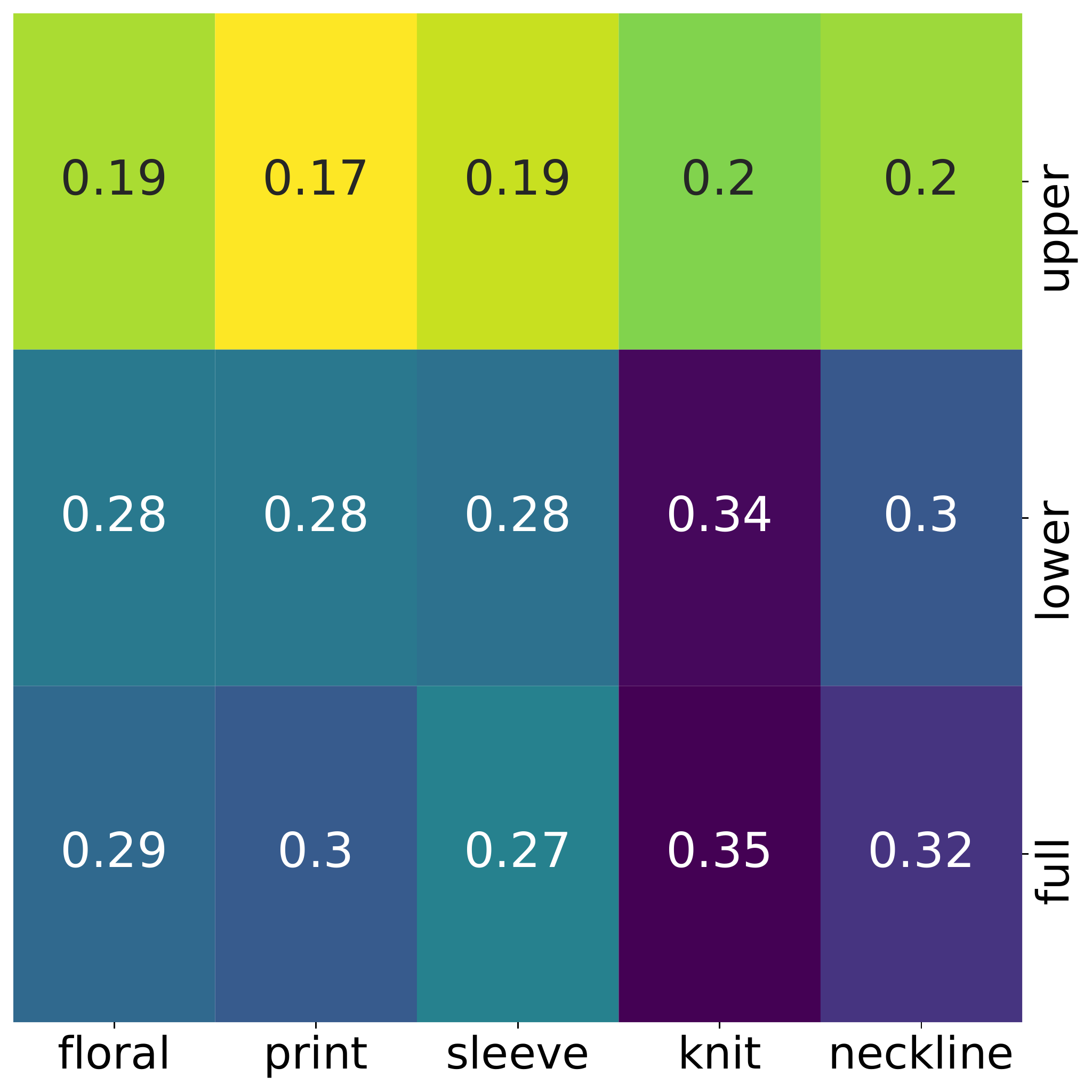}
\caption{}
\label{fig:fashion}
\end{subfigure}
\hspace{2em}
\begin{subfigure}[t]{0.2 \linewidth}
\includegraphics[width=\linewidth]{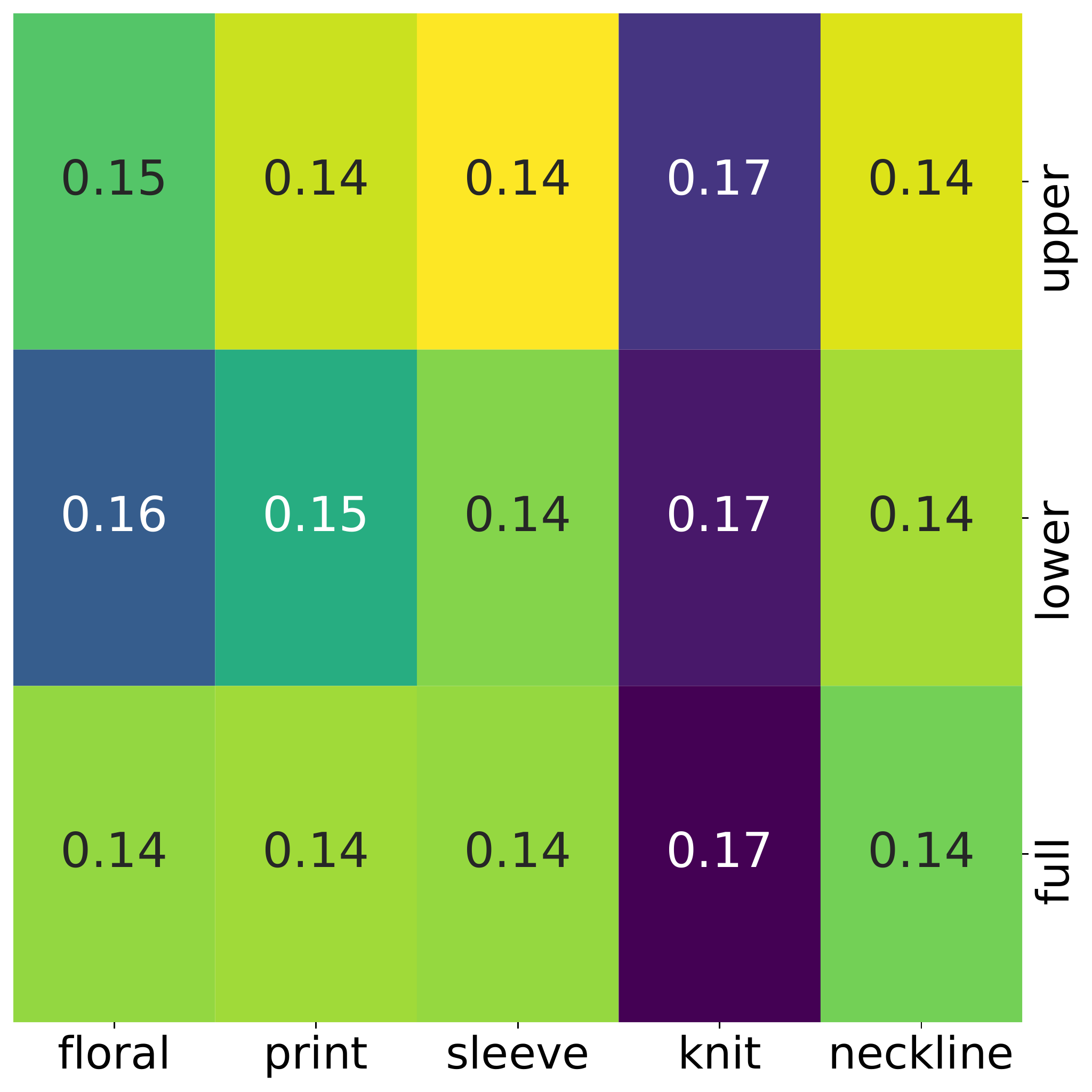}
\caption{}
\label{fig:fashion-Task2Vec}
\end{subfigure}
\hspace{2em}
\begin{subfigure}[t]{0.2 \linewidth}
\includegraphics[width=\linewidth]{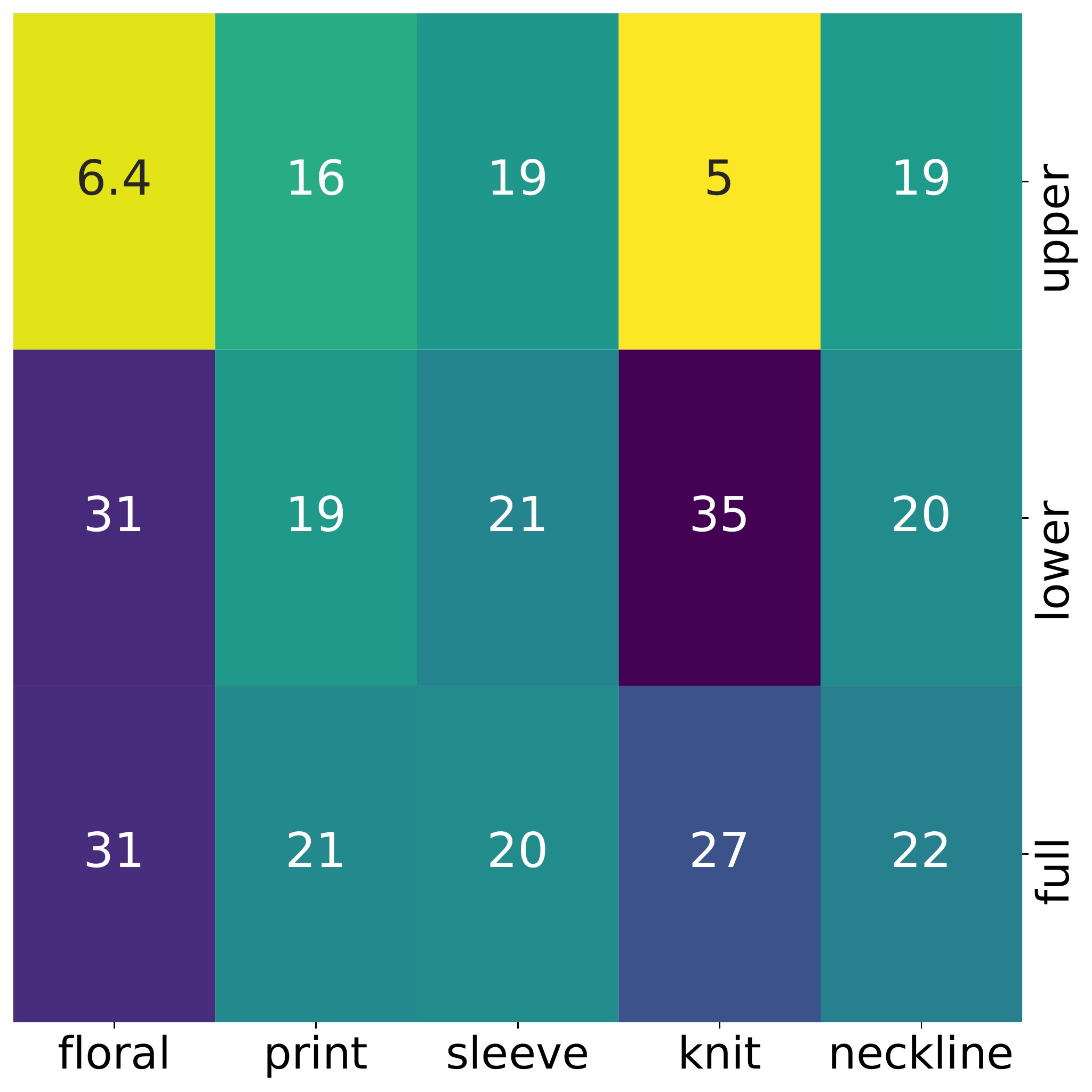}
\caption{}
\label{fig:fashion_FineTune}
\end{subfigure}
\caption{
\cref{fig:fashion} shows distances (numbers in the cell)
among sub-tasks in DeepFashion computed using our coupled transfer
process (r $=$ 0.37, p $=$ 0.33),
~\cref{fig:fashion_FineTune} shows distances estimated using Task2Vec
(r $=$ 0.04, p $=$ 0.75)
while~\cref{fig:fashion_FineTune} shows distances estimated using
fine-tuning (r $=$ 0.54, p $=$ 0.36 with itself).
Numerical values of the distances in this figure are not
comparable with each other.
Coupled transfer, Task2Vec and fine-tuning all agree with that transferring to knit is relatively hard. Transferring from upper-cloth to knit is easy via fine-tuning and coupled transfer correctly estimates this distance to be small; the distance estimated by Task2Vec is much larger in comparison. Since these matrices are non-square, we ran the Mantel test for three 3$\times$3 submatrices (sweep across columns) of these 3$\times$5 matrices and report the mean test statistic and the average $p$-value across these tests above.
}
\label{fig:deepfashion}
\end{figure}

For the Deep Fashion dataset~\citep{liuLQWTcvpr16DeepFashion}, we consider three binary category classification tasks (upper clothes, lower clothes, and full clothes) and five binary attribute classification tasks (floral, print, sleeve, knit, and neckline). We show results in~\cref{fig:deepfashion} using 3$\times$ 5 distance matrices where numbers in each cell indicate the distance between the source task (row) and the target task (column). We show results using a wide-residual-network (WRN-16-4,~\citep{zagoruyko2016wide}).

The model is trained using SGD for 400 epochs with a mini-batch size 20.
Learning rate is $10^{-1}$ for epochs 0 -- 120, $2 \times 10^{-2}$ for epochs 120 -- 240,
$4 \times 10^{-3}$ for epochs 240--320, and $8 \times 10^{-4}$ for epochs 320 -- 400.
We sample 14,000 images from the source and target datasets to compute distances.
A mini-batch size of 20 is used during transfer and we run~\cref{eq:sgd_prox} for 60 epochs (14000/20 = 700 weight updates per epoch).

\section{Proof of~\cref{thm:integral_of_fisher_dist_main}}
\label{ss:a:proof_integral_of_fisher_dist}

We first prove a simpler theorem.

\begin{theorem}
\label{app:thm:simpler_theorem}
Given a trajectory of the weights $\cbr{\wt}_{\t \in [0,1]}$ and a sequence $0 \leq \t_1 <\t_2 <...<\t_{K}\leq 1 $, then for all
$\e > \f{2}{K} \sum_{k= 1}^{K}\RR_{N}(\norm{w(\t_k )}_\fr)$, the probability that
\[
    \f{1}{K}\sum_{k= 1}^{K} \rbr{\E_{(x,y) \sim p_{\t_k}} \sbr{\ell(\w(\t_k), x,y)}
    - \f{1}{N} \sum_{(x,y) \sim \hat{p}_{\t_k}} \ell(\w(\t_k), x,y)}
\]
is greater than $\e$ is upper bounded by
\beq{
    \exp \cbr{-\f{2 K}{M^2} \rbr{\e - \f{2}{K} \sum_{k= 1}^{K}\RR_{N}(\norm{w(\t_k)}_\fr)}^2}.
    \label{thm:integral_of_rademacher_complexity}
}
\end{theorem}

\begin{proof}
For each moment $\t_k$, by taking supremum
\beq{
 \E_{(x,y)\sim p_{\t_k}} \ell(w(\t_k), x, y) - \f{1}{N}\sum_{ (x,y) \sim \hat{p}_{\t_k} }\ell(w(\t_k), x,y) \leq \sup_{ \|w \|_\fr \leq \|w(\t_k) \|_\fr} \rbr{\E_{(x,y)\sim p_{\t_k}} \ell(w, x, y) - \f{1}{N}\sum_{(x,y) \sim \hat{p}_{\t_k}} \ell(w, x, y) },
 \label{eq: supre}
}
where $\|\cdot \|_\fr$ denotes Fisher-Rao norm~\citep{liang2019fisher}. The right hand side of inequality\cref{eq: supre} is a random variable that depends on the drawn sampling set $\hat{p}_{\t_k}$ with size $N$. Denoting
\beq{
    \p(\hat{p}_{\t_k}):\ = \sup_{ \|w \|_\fr \leq \|w(\t_k) \|_\fr} \rbr{\E_{(x,y)\sim p_{\t_k}} \ell(w, x, y) - \f{1}{N}\sum_{(x,y) \sim \hat{p}_{\t_k}} \ell(w, x, y) },
}
We would like to bound the expectation of $\phi(\hat{p}_{\t_k})$ in terms of the Rademacher complexity. In order to do this, we introduce a “ghost sample” with size $N$,
$\hat{p}_{\t_k}'
$, independently drawn identically from $p_{\t_k}(x,y)$, we rewrite the expectations

\aeqs{
\E_{\hat{p}_{\t_k}}\phi(\hat{p}_{\t_k}) &=  \E_{\hat{p}_{\t_k}} \left[ \sup_{ \|w \|_\fr \leq \|w(\t_k) \|_\fr} \left(\E_{(x,y)\sim p_{\t_k}} \ell(w, x, y) - \f{1}{N}\sum_{(x,y) \sim \hat{p}_{\t_k}} \ell(w, x, y) \right) \right]\\
& = \E_{\hat{p}_{\t_k}} \left[ \sup_{ \|w \|_\fr \leq \|w(\t_k) \|_\fr}  \E_{ \hat{p}_{\t_k}'}\left(\f{1}{N}\sum_{(x,y) \sim \hat{p}_{\t_k}'} \ell(w, x, y) - \f{1}{N}\sum_{(x,y) \sim \hat{p}_{\t_k}} \ell(w, x, y) \right) \right] \\
& \leq \E_{\hat{p}_{\t_k}, \hat{p}_{\t_k}', \s} \left[ \sup_{ \|w \|_\fr \leq \|w(\t_k) \|_\fr}  \f{1}{N}\left(\sum_{(x,y) \sim \hat{p}_{\t_k}}\s^i(\ell(w, x, y) - \ell(w, x, y) )\right) \right] \\
& \leq \E_{\hat{p}_{\t_k}, \s}\left[ \sup_{ \|w \|_\fr \leq \|w(\t_k) \|_\fr}  \f{1}{N}\sum_{(x,y) \sim \hat{p}_{\t_k}}\s^i \ell(w, x, y) \right] + \E_{\hat{p}_{\t_k}, \s}\left[ \sup_{ \|w \|_\fr \leq \|w(\t_k) \|_\fr}  \f{1}{N}\sum_{(x,y) \sim \hat{p}_{\t_k}}\s^i \ell(w, x, y) \right] \\
& = 2 \mathcal{R}_{N}( \|w(\t_k)\|_\fr ),
}
where $\s = (\s^1, \s^2,\ldots,\s^N)$ are independent random variables drawn from the Rademacher distribution, the last equality is followed by the definition of Rademacher Complexity within $\|w(\t_k)\|_\fr$-ball in the Fisher-Rao norm. By Hoeffding's lemma, for $\l > 0$
\beq{
\aed{
    \E_{\hat{p}_{\t_k}}\exp\left\{ \l \left(\E_{(x,y)\sim p_{\t_k}} \ell(w(\t_k), x, y) - \f{1}{N}\sum_{(x,y) \sim \hat{p}_{\t_k}} \ell(w(\t_k), x, y)\right)  \right\} &= \E_{\hat{p}_{\t_k}} e^{\l \phi(\hat{p}_{\t_k})} \\
& \leq e^{ \l \E_{\hat{p}_{\t_k}}\phi(\hat{p}_{\t_k}) + \f{\l^2 M^2}{8} }\\ &\leq e^{ 2 \l \mathcal{R}_{N}( \|w(\t_k)\|_\fr ) + \f{\l^2 M^2}{8} }.
}
\label{ineq: rad}
}
For each moment $\t_k$, we have inequality\cref{ineq: rad}, which implies
\aeqs{
  &\E_{\hat{p}_{\t_k}:\ 1 \leq k\leq K} \exp\left\{  \l \sum_{k= 1}^{K}\rbr{ \E_{(x,y)\sim p_{\t_k}} \ell(w(\t_k), x, y) - \f{1}{N}\sum_{(x,y) \sim \hat{p}_{\t_k}} \ell(w(\t_k), x, y) } \right\}\\
  &=\prod_{k= 1}^{K} \E_{\hat{p}_{\t_k}}\exp\left\{ \l \left(\E_{(x,y)\sim p_{\t_k}} \ell(w(\t_k), x, y) - \f{1}{N}\sum_{(x,y) \sim \hat{p}_{\t_k}} \ell(w(\t_k), x, y)\right)  \right\}\\
  &\leq \exp\left\{\sum_{k= 1}^{K} \left[2\l \mathcal{R}_{N}( \|w(\t_k)\|_\fr ) + \f{\l^2 M^2}{8}\right] \right\}.
}
Finally for all $K \e > 2 \sum_{k= 1}^{K}\RR_{N}(\norm{w(\t_k)}_\fr)$, by Markov's inequality
\beq{
    \aed{
    &Pr\left\{ \sum_{k= 1}^{K}\rbr{ \E_{(x,y)\sim p_{\t_k}} \ell(w(\t_k), x, y) -  \f{1}{N}\sum_{(x,y) \sim \hat{p}_{\t_k}} \ell(w(\t_k), x, y)} > K \epsilon \right\}\\
    &\leq \exp\left\{ - \l K \epsilon + \sum_{k= 1}^{K} \left[2\l \mathcal{R}_{N}( \|w(\t_k)\|_\fr ) + \f{\l^2 M^2}{8}\right] \right\}
  }
  \label{ineq: mark}
}
Put $\l = \f{ 4K\rbr{ \epsilon - \f{2}{K} \sum_{k= 1}^{K}\RR_{N}(\norm{w(\t_k)}_\fr)} }{M^2}$ in right hand side of inequality\cref{ineq: mark}, then we finish the proof.
\end{proof}

\section*{Proof of~\cref{thm:integral_of_fisher_dist_main}}

The upper bound in~\cref{ineq: mark} above states that we should minimize the
Rademacher complexity of the hypothesis space in order to ensure that the
weight trajectory has a small generalization gap at all time instants. For linear
models, as discussed in the main paper~\citep{liang2019fisher}, the Rademacher complexity can be related to
the Fisher-Rao norm $\inner{w}{g w}$. The Fisher-Rao distance on the manifold, namely
\beq{
    \int_0^1 \E_{x \sim p_\tau(x)} \sbr{\sqrt{2 \kl\rbr{p_{\wt}(\cdot | x),\ p_{\wtp}(\cdot | x)}}} \d \t
    =\int_0^1 \E_{x \sim p_\tau(x)}  \sqrt{\inner{\dot{\wt}}{g(\wt) \dot{\wt}}}\  \d \t
    \label{eq:intro_problem_1}
}
is only a lower bound on the integral of the Fisher-Rao norm along the weight trajectory.
We therefore make some additional assumptions in this section to draw out a crisp
link between the Fisher-Rao \emph{distance} and generalization gap along the trajectory.

Let $\ell(w; x, y) = - \log p_w(y|x)$ be the cross-entropy loss on sample $(x,y)$.
We assume that at each moment $\t \in [0, 1]$, our model $p_{w(\t)}(y|x)$ predicts on the interpolating distribution $p_\t(y|x)$ well, that is
\[
    p_{w(\t)}(y|x)  \approx  p_\t(y|x)
\]
for all input $x$; this is a reasonable assumption and corresponds to taking a large number of mini-batch updates in~\cref{eq:sgd_prox}.
We approximate the FIM using the empirical FIM, i.e., we approximate the distribution $p_\t(y | x)$ as a Dirac-delta distribution on the interpolated labels $y_\t(x)$.
Observe that
\beq{
    \aed{
        \inner{\dot{\wt}}{g(\wt) \dot{\wt}}  &= \inner{\dot{ \wt}}{\E_{y|x \sim p_{\t}} \partial_{w} \ell_{\wt}(y|x) \partial_{w} \ell_{\wt}(y|x)^\top \dot{ \wt }}  \\
        &\approx  \inner{\dot{ \wt }}{\partial_{w}\ell(\wt; x, y_\t(x))\ \partial_{w}\ell(\wt; x, y_\t(x))^\top \dot{ \wt } }\\
        & = \left| \f{\ell(\wtp; x, y_\t(x)) -  \ell(\wt; x, y_\t(x))}{\d \t}\right|^2\\
        & = \left| \f{\Delta \ell(\wt)}{\d \t}\right|^2,
    }
\label{eq: approxi}
}
where we use the shorthand
\[
    \Delta \ell(\wt) := \ell(\wtp; x, y_\t(x)) -  \ell(\wt; x, y_\t(x)),
\]
and plug~\cref{eq: approxi} in the integration in ~\cref{eq:intro_problem_1}
\beq{
    \aed{
    \int_0^1 \E_{x \sim p_\tau(x)} \sbr{\sqrt{2 \kl\rbr{p_{\wt}(\cdot | x),\ p_{\wtp}(\cdot | x)}}} \d \t &=\int_0^1 \E_{x \sim p_\tau(x)}  \sqrt{\inner{\dot{\wt}}{g(\wt) \dot{\wt}}}\  \d \t \\
    &\approx \int_0^1 \E_{x \sim p_\tau(x)}  \sbr{\abr{\Delta \ell(\wt}}.
    }
\label{eq: int_appro}
}

On the other hand, for moment $\t$ let $\omt \ni \wt$ be a compact neighborhood of $w(\t)$ in weights space, Rademacher complexity of the class of loss function is upper bounded as following
\beq{
    \aed{
        \RR_N(\omt) &= \E_{\ph \sim p_{\t}^{N}}\E_\s \sbr{\sup_{w \in \omt} \f{1}{N}\sum_{i=1}^N\ \s^i \ell(w; x^i, y^i)}\\
        &  = \E_{\ph \sim p_{\t}^{N}}\E_\s \sbr{\sup_{w \in \omt} \f{1}{N}\sum_{i=1}^N\ \s^i \ell(\wt; x^i, y^i) + \s^i\rbr{ \ell(w; x^i, y^i) - \ell(\wt; x^i, y^i)} } \\
        & \leq \E_{\ph \sim p_{\t}^{N}}\E_\s\sbr{ \f{1}{N}\sum_{i=1}^N\ \s^i \ell(\wt; x^i, y^i) + \sup_{w \in \omt}\f{1}{N}\sum_{i=1}^N\ | \ell(w; x^i, y^i) - \ell(\wt; x^i, y^i)| }  \\
        & = 0 + \E_{\ph \sim p_{\t}^{N}} \sbr{ \sup_{w \in \omt}\f{1}{N}\sum_{i=1}^N\ | \ell(w; x^i, y^i) - \ell(\wt; x^i, y^i)| }\\
        & \longrightarrow \sup_{w \in \omt}\E_{x \sim p_{\t}}| \ell(w; x, y_\t(x)) - \ell(\wt; x, y_\t(x))|
    },
\label{eq: rad_bound}
}
as $N$ goes to infinity. The last step in~\cref{eq: rad_bound} is followed by the compactness of $\omt$ and the Lipschitz continuity of the loss function. Let
\beq{
    \omt:= \{w | \E_{x \sim p_{\t}}| \ell(w; x, y_\t(x)) - \ell(\wt; x, y_\t(x))| \leq  \E_{x \sim p_{\t}}| \ell(\wtp; x, y_\t(x)) - \ell(\wt; x, y_\t(x))|\},
}
be the neighborhood of $\wt$ within which the loss function changes less than $\abs{\Delta \ell(\wt)}$.
Compare this with~\cref{eq: int_appro}, the Rademacher complexity of $\omt$
is exactly upper bounded by integration increments appearing in the
expression for the Fisher-Rao distance.
If we substitute $\|w(\t)\|_\fr$-ball in~\cref{thm:integral_of_rademacher_complexity}
with this modified $\omt$, we have the following theorem.

\begin{theorem}
\label{thm:integral_of_fisher_dist}
Given a trajectory of the weights $\cbr{\wt}_{\t \in [0,1]}$ and a sequence $0 = \t_0 \leq \t_1 <\t_2 <...<\t_{K}\leq 1 $, for all
$\e > 2 \sum_{k= 1}^{K}(\t_k - \t_{k-1})\E_{x \sim p_{\t}} |\Delta \ell(w(\t_{k-1}))|$, the probability that
\[
    \f{1}{K}\sum_{k= 1}^{K} \rbr{\E_{(x,y) \sim p_{\t_k}} \sbr{\ell(\w(\t_k), x,y)}
    - \f{1}{N} \sum_{(x,y) \sim \hat{p}_{\t_k}} \ell(\w(\t_k), x,y)}
\]
is greater than $\e$ is upper bounded by
\beq{
    \exp \cbr{-\f{2 K}{M^2} \rbr{\e - 2 \sum_{k= 1}^{K}(\t_k - \t_{k-1}) \E_{x \sim p_{\t_k}} \sbr{\abr{\Delta \ell(w(\t_{k-1}))}} }}.
}
\end{theorem}
\begin{proof}
The proof is same as in~\cref{thm:integral_of_rademacher_complexity} except for substituting $\RR_N(\norm{w(\t_k)}_{\fr})$ with $\RR_N(\Omega_{\t_k})$ and using upper bounds~\cref{eq: rad_bound}, and
\beq{
\aed{
    &\Omega_{\t_k} = \{w | \E_{x \sim p_{\t_k}}| \ell(w; x, y_{\t_k}(x)) - \ell(w(\t_k); x, y_{\t_k}(x))|\\
    &\leq K (\t_k - \t_{k-1}) \E_{x \sim p_{\t_k}}| \ell(w(\t_k); x, y_{\t_k}(x)) - \ell(w(\t_{k-1}); x, y_{\t_k}(x))|\}.
    }
}
\end{proof}

We can now relate the Fisher-Rao distance~\cref{eq:intro_problem_1} and the generalization bound in~\cref{thm:integral_of_fisher_dist}. For instance, if $\left| \f{\d}{\d \t}\ell(\wt; x, y_\t(x)) \right|$ is Riemann integrable over $\t$, then as $K$ goes to infinity, there exists a sequence $0 = \t_0 \leq \t_1 <\t_2 <...<\t_{K}\leq 1 $ such that
\beq{
    \aed{
        \sum_{k= 1}^{K}(\t_k - \t_{k-1})\E_{x \sim p_{\t_k}}| \ell(w(\t_k); x, y_{\t_k}(x)) - \ell(w(\t_{k-1}); x, y_{\t_k}(x))|
        &\longrightarrow \int_0^1 \E_{x \sim p_\tau(x)}  \left| \ell(\wtp; x, y_\t(x)) -  \ell(\wt; x, y_\t(x))\right|  \\
        &\approx\int_0^1 \E_{x \sim p_\tau(x)} \sbr{\sqrt{2 \kl\rbr{p_{\wt}(\cdot | x),\ p_{\wtp}(\cdot | x)}}} \d \t.
    }
}
This shows that computing the Fisher-Rao distance between two points on the statistical
manifold results in a weight trajectory that minimizes the the generalization gap
of weights trained on the interpolated distribution along the trajectory. In other
words, one may either think of our coupled transfer process as computing the
Fisher-Rao distance or as finding a weight trajectory that connects weights with
a small generalization gap.

\section{Frequently Asked Questions (FAQs)}

\begin{enumerate}
\item \tbf{How is this distance better than methods such as Wasserstein distance, Maximum Mean Discrepancy (MMD), Hellinger distance or other $f$-divergences to measure distances between probability distributions?}

Measuring distance between learning tasks is different than measuring distances between the respective data distributions. The above concepts can only measure distances between data distributions, they do not consider the hypothesis class used to transfer across the two distributions and therefore do not reflect the true difficulty of transfer. The experiment in~\cref{fig:embedding_distance} demonstrates this. This point in fact is the central motivation of our paper. Also see the discussion of related work in~\cref{s:related_work}.

\item \tbf{Why do your distances range from small to large values?}

We discuss this in~\cref{rem:coupled_comparable}.
The scale of distances can be quite different for different hypothesis spaces
but this is not a problem if they can be compared across
architectures for the same task pair. Since the coupled transfer distance
measures the length of the trajectory on the statistical manifold which is invariant to the specific parameterization of the model, the numerical value of the distance
has a sound grounding in theory and not on some arbitrary scale.
Further, just like the cosine distance scales with the inner product and can be normalized using the $\ell_2$ norm of the respective vectors, we envision that our distance can be normalized using the coupled transfer distance to some ``canonical'' task (say, actual vs. fake source/target images) to get a better dynamic range. We are currently studying which tasks are good canonical tasks for this purpose.

\item \tbf{The coupled transfer distance trains the model multiple times between source and target tasks to estimate the distance. How is this useful in practice to select, say, a good source dataset to pre-train from? Interesting formulation, but too complex to use in practice.}

We think of our work as a first step towards the challenging problem of understanding distances between learning tasks. Our final goal is indeed to use the tools developed here for practical applications, e.g., to design methods that can select the best source task to transfer from while fitting a given task or the best architecture to transfer between a given set of tasks, but we are not there yet. The practical utility of this work is to identify that typical methods in the literature for measuring task distances (see related work discussed in~\cref{s:related_work}) leave a lot on the table. Theoretically they do not explicitly characterize the hypothesis class being transferred. Empirically, distances estimated by typical methods do not correlate strongly with the difficulty of fine-tuning (see~\cref{fig:c10_c100,fig:c100}). Our development provides concrete theoretical tools to understand other task distances that correlate well with the coupled transfer distance, and thereby the difficulty of fine-tuning.

For the same reason, we do not think the technical complexity of formalizing and computing the coupled transfer distance should take anything away from its intellectual metric. Our goal is to develop theoretical tools to understand when transfer between tasks is easy and when it is not, it is not to develop a good fine-tuning algorithm.

\item \tbf{Does coupled transfer obtain better generalization error on the target task than standard fine-tuning?}

Coupled transfer explicitly modifies the task while standard fine-tuning does not, so this is a natural question. We have explored it in~\cref{fig:final_accuracy}.
Our experiment shows that, broadly, the coupled transfer improves generalization. This is consistent with existing literature, e.g.,~\citet{gao2020free}, which employs task interpolation for better transfer learning. We however note that improving fine-tuning is not our goal in this paper; in fact, we want our task distance to correlate with the difficulty of fine-tuning.

\item \tbf{Feature extractor $\varphi$ for initializing $\G^0$ is trained on a generic task, how is this task related to source/target?}

We discuss this on Lines 198--215 (right column) in the main paper. The feature extractor is only used to initialize the coupling $\G^0$, couplings in successive iterations $\G^k$ are computed using the ground metric in~\cref{eq:ckp} and do not use the feature extractor.

Using a feature extractor to compute OT distances is quite common in the literature, e.g.,~\citep{cui2018large}. We use a ResNet-50 pre-trained on ImageNet as the feature generator to compute the initialization $\G^0$ for all experiments in this paper. ImageNet is a different task than the ones considered in this paper (subsets of MNIST, CIFAR-10, CIFAR-100 and Deep Fashion). If the feature generator's task is closely related to only one of the source/target tasks but not the other, the task distance will require more iterations to converge. For our experimental setup, ImageNet is, roughly speaking, a superset of the tasks we analyze, this enables the coupled transfer distance in our experiments to converge within 4--5 iterations. Note that each iteration of~\cref{eq:algorithm} is quite non-trivial and takes a few GPU-hours; it performs multiple epochs of weight updates and estimates $C_{ij}$ along the trajectory to update all the blocks of the coupling matrix $\G^k$.

\item \tbf{The expression for the interpolated distribution in~\cref{eq:interp_finite} is for the quadratic ground metric $C_{ij} = \norm{\xs^i-\xt^j}_2^2$ but the ground metric in~\cref{eq:ckp} is different.}

The interpolation in~\cref{eq:interp_finite} McCann's displacement convexity~\citep{mccann1997convexity} for the space of probability measures under the Wasserstein metric. This result identifies when functionals on the space of probability measures are convex along geodesics. More formally, if $F: \PP(\Om) \to \reals$ is $\l$-geodesically-convex functional, then
\[
    (1-\t) F(\r_0) + t F(\r_1) \geq F(\r_\t) + \f{\l}{2} \t (1-\t) W_2(\r_0, \r_1)^2;
\]
here $\r_0, \r_1 \in \PP(\Om)$ are two probability measures supported on the set $\Om$ and $\r_\t$ is the interpolant at time $\t$ along the geodesic in $W_2$ metric joining them. Computing displacement interpolation for general ground metrics, even analytically, is difficult; see~\citet[Chapters 16--17]{villani2008optimal}. It is therefore very popular in the optimal transport literature to study interpolation under the quadratic ground metric.
In order to keep the implementation simple and focus on the main idea of coupled transfer, we use the expression for displacement interpolation $p_\tau$ in~\cref{eq:interp_finite} for the quadratic ground metric $C_{ij} = \norm{\xs^i-\xt^j}_2^2$ but compute the optimal coupling $\G$ using the Fisher-Rao distance $C_{ij} = \dfr(p_{w(0)}(\cdot\ |\ \xs^i), p_{w(1)}(\cdot\ |\ \xt^j))$ as the tasks are interpolated using the coupling of the previous iteration $\G^{k-1}$; see~\cref{eq:ckp}. Note that this does not change the fact that $p_\tau$ is \emph{an interpolation}, it is however not a displacement interpolation anymore for our particular chosen ground metric $C_{ij}$. This is a pragmatic choice which keeps our theoretical development tractable.

\item \tbf{Why use $\text{Beta}(\tau, 1-\tau)$ to interpolate?}

We discuss this on Lines 217--228 in the main paper.
Mathematically, employing this technique really means that we use some other ground metric than the quadratic cost in the OT problem;
this is a minor modification with a big benefit
of keeping the interpolated task within the manifold of natural images.

\item \tbf{How do you compute the integral in~\cref{eq:ckp}?}

Integral on $\t$ in~\cref{eq:ckp} is computed using its Riemann approximation along
the weight trajectory $\cbr{w(\t): t \in [0,1]}$ given by~\cref{eq:sgd_prox}.

\item \tbf{Why do~\cref{thm:integral_of_fisher_dist_main} and~\cref{app:thm:simpler_theorem} do not use the standard PAC-learning analysis?}

PAC analysis without ground-truth labels for the data from the interpolated distribution is difficult. We therefore bound the generalization gap in terms of the loss $\ell(w, x, y)$ where the label generating mechanism is a simple linear interpolation between one-hot labels of the source and target tasks.
Let us note that a PAC-Bayes bound between the source and target posterior weight distributions is given in Achille et al., 2019c.

\item \tbf{Why should a larger model have a smaller coupled transfer distance in~\cref{fig:wrn164} compared to~\cref{fig:c100}?}

We discuss this on Lines 374--383 in the main paper.

\end{enumerate}

\end{appendices}

\end{document}


\ignore{

\twocolumn[
\aistatstitle{\mytitle}

\aistatsauthor{Yansong Gao \And Pratik Chaudhari}

\aistatsaddress{
    Applied Mathematics and Computational Science,\\
    University of Pennsylvania.\\
    Email: \href{mailto:gaoyans@sas.upenn.edu}{gaoyans@sas.upenn.edu}
    \And
    Electrical and Systems Engineering,\\
    University of Pennsylvania.\\
    Email: \href{mailto:pratikac@seas.upenn.edu}{pratikac@seas.upenn.edu}
    }
]

\graphicspath{{./fig/}}

\begin{abstract}
This paper computes a distance between tasks modeled as joint distributions on data and labels. We develop a stochastic process that transports the marginal on the data of the source task to that of the target task, and simultaneously updates the weights of a classifier initialized on the source task to track this evolving data distribution. The distance between two tasks is defined to be the shortest path on the Riemannian manifold of the conditional distribution of labels given data as the weights evolve. We derive connections of this distance with Rademacher complexity-based generalization bounds; distance between tasks computed using our method can be interpreted as the trajectory in weight space that keeps the generalization gap constant as the task distribution changes from the source to the target. Experiments on image classification datasets show that this task distance helps predict the performance of transfer learning: fine-tuning techniques have an easier time transferring to tasks that are close to each other under our distance.
\end{abstract}

\section{Introduction}
\label{s:intro}

A part of the success of Deep Learning stems from the fact that deep networks
learn features that are discriminative yet flexible.
Models pre-trained on a task can be
easily adapted to perform well on other tasks. The transfer learning literature
forms an umbrella for such adaptation techniques. Transfer learning indeed works
very well, see for instance~\citet{mahajan2018exploring,dhillon2019a,kolesnikov2019large,joulinLearningVisualFeatures2016}
for image classification or~\citet{47751} for
language modeling, to name a few of the
many large-scale demonstrations.
There are however also situations when
transfer learning does not work well.
For instance, a pre-trained ImageNet model is a
poor representation to transfer to images in radiology~\citep{merkow2017deepradiologynet}.

It stands to reason
that if source and target tasks are ``close'' to each other then we
should expect transfer learning to work well;
it may be difficult to transfer across tasks that are ``far away''.
We lack theoretical tools that define when two learning tasks are
close to each other; while there are numerous candidates in the current
literature a unified understanding of these domain-specific methods remains
elusive. We also lack algorithmic tools to robustly transfer models
on new tasks, for instance, fine-tuning methods require careful
design~\citep{li2019rethinking} and it is unclear
what one should do if they do not work well.
Our approach on this problem is based on the following two ideas.

\begin{figure}[!t]
\centering
\includegraphics[width=\linewidth]{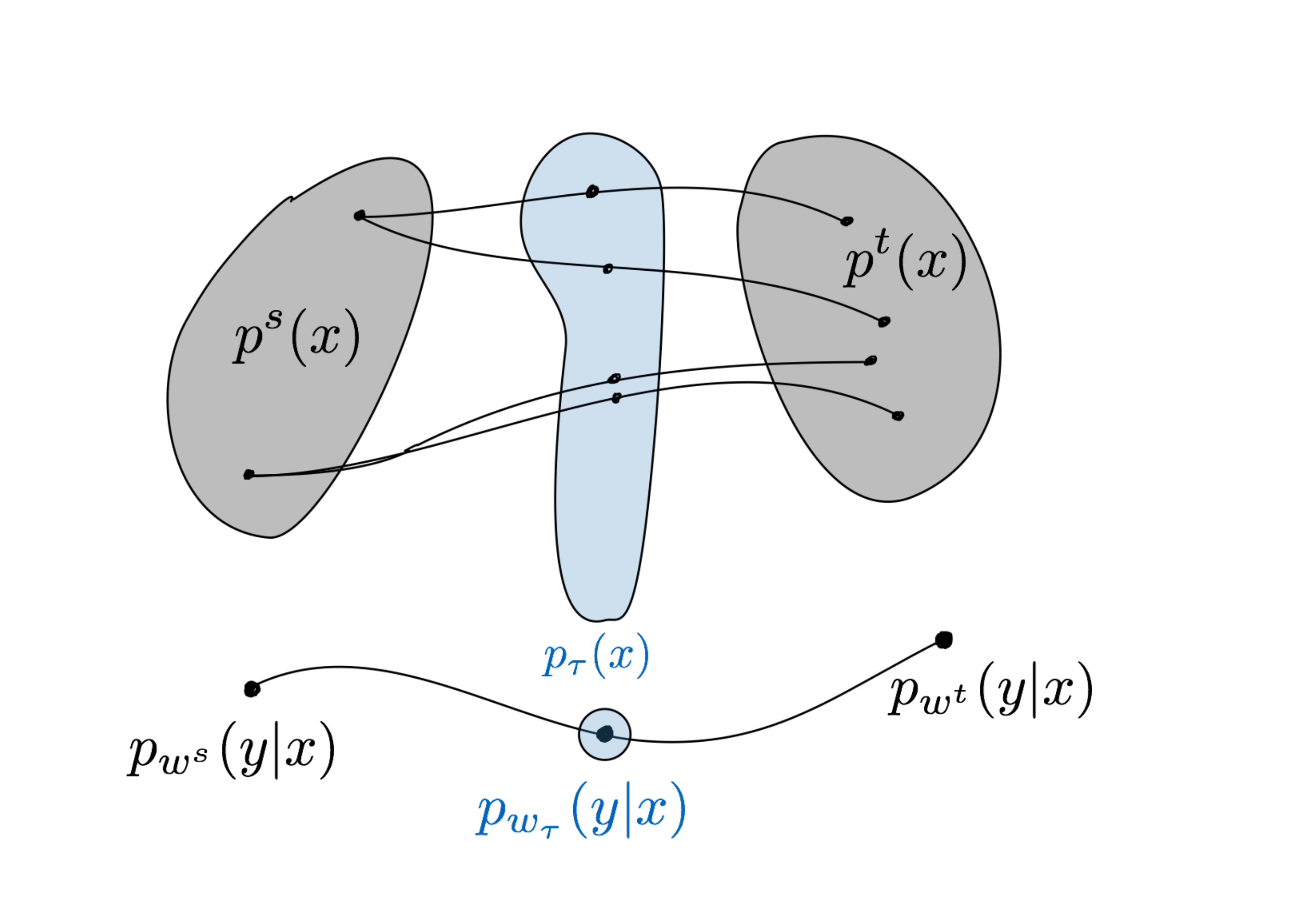}
\caption
{\tbf{Coupled transfer of the data and the conditional distribution}.
We solve an optimization problem that transports
the source data distribution $\ps(x)$ to the
target distribution $\pt(x)$ as $\t \to 1$
while simultaneously updating the model
using samples from the interpolated distribution $\ptau(x)$. This modifies the
conditional distribution $p_\ws(y|x)$ on the source task to the
corresponding distribution on the target task $p_\wb(y|x)$.
Distance between source and target tasks is defined to be the length
of the optimal trajectory of the weights under the Fisher Information Metric.
}
\label{fig:schematic}
\end{figure}

\paragraph{Main ideas}
First, most algorithmic methods that devise a distance between tasks in the
current literature, do not take into account the hypothesis space of the model.
We argue that
transferring the representation of a small model with limited capacity from a
source task to a target task is difficult because there are fewer redundant
features in the model. The distance between the same two tasks may
be small if a high-capacity model is being transferred; this is
especially pertinent when
transferring deep networks. A definition of task distance or an algorithm
for instantiating this definition,
therefore needs to take the capacity of the hypothesis class into account.

Second, distance between tasks as measured using techniques in the current
literature does not take into account the dynamics of learning; distances
often depend on how the transfer was performed. For instance, if one considers
the number of epochs of fine-tuning required to reach a certain accuracy,
a different strategy may result in a different distance.
A sound notion of distance between tasks should not depend
on the specific dynamical process used of transfer.

\paragraph{Summary of contributions}
Given data $x$ and labels $y$, we model the source and target tasks as
joint distributions $\ps(x,y)$ and $\pt(x,y)$ respectively where
\[
    \ps(x,y) = p^s_{\ws}(y | x)\ \ps(x);
\]
here $\ws$ are the
parameters (weights) of a classifier on the source task. The target
task is decomposed analogously. We define the
distance between finite-sample \emph{datasets} $\psh = \cbr{(x_i, y_i)\sim
\ps}_{i=1}^{\ns}$ drawn from tasks (with $\pth$ defined analogously)
as the solution of the optimization problem
\beq{
    \aed{
        &\scalemath{\mscale}{\min_{\G \in \Pi}
                \int_0^1 \E_{(x,y) \sim p_\tau(x,y)}
                \sbr{\sqrt{2 \kl\rbr{p_{\wt}(\cdot | x),\ p_{\wtp}(\cdot | x)}}}}\\
    &\scalemath{\mscale}{\trm{subject to}\
            \f{\d \wt}{\dt} = \widehat{\grad}_w \cbr{\E_{(x,y) \sim \ptau}
            \sbr{\log p_{\wt}(y | x)}}};
        }
    \label{eq:intro_problem_1}
}
where
\beq{
    \aed{
        &\scalemath{\mscale}{p_\tau(x,y) =
        \sum_{i=1}^\ns \sum_{j=1}^\nt \G_{ij}\
                \delta_{\cbr{(1-\t) \xs_i + \t \xt_j}}\
                \delta_{\cbr{(1-\t) \ys_i + \t \yt_j}}},\\
        &\scalemath{\mscale}{
        \Pi = \cbr{\G \in \reals^{\ns \times \nt}_+: \G
        \ones = \psh,\ \G^\top \ones = \pth}}.
    }
    \label{eq:intro_problem_2}
}
The square root of the $\kl$-divergence measures the infinitesimal
distance traveled along a geodesic on the manifold of probability distributions
$p_\wt(y|x)$. Distance as defined
in~\cref{eq:intro_problem_1,eq:intro_problem_2}
is the path length on this manifold.
Weights $w(\t)$ evolve from their initial value $w^s$
using stochastic gradient descent-based updates in~\cref{eq:intro_problem_1}
while the interpolated data distributed $p_\tau$ to which weights are being fitted to
evolves simultaneously from its
initial value $\ps$ to its final value $\pt$
using the linear coupling matrix $\G$.

The distance in~\cref{eq:intro_problem_1} is the
Fisher-Rao distance on the manifold of distributions $p_w(y  | x)$.
We use this to draw a link to Rademacher
complexity-based generalization bounds which gives an intuitive
understanding of the trajectory
of weights computed in~\cref{eq:intro_problem_1}. We show
that our approach modifies
the task's data distribution and weights so as to to minimize the
\emph{average generalization gap} along the trajectory that joins the
weights on the source task $\ws$ to their terminal value on the target task $w(1)$.

We devise an algorithmic procedure to solve the optimization problem
in~\cref{eq:intro_problem_1,eq:intro_problem_2} for image classification
tasks that are subsets of CIFAR-10 and
CIFAR-100 datasets~\citep{krizhevskyLearningMultipleLayers2009}.
We show in~\cref{s:expt} that the coupled transfer process estimates distances
that are more consistent with the difficulty of fine-tuning, as compared to
baselines.

\section{Theoretical setup}
\label{s:setup}

Consider a dataset $\psh = \cbr{(x_i, y_i) \sim \ps}_{i=1,\ldots,\ns}$
where $x_i \in X, y_i \in Y$ denote input data and their ground-truth annotations respectively.
Data are drawn from a distribution $\ps$ supported on $X \times Y$.
Training a parameterized model, say a deep network, involves minimizing
the cross-entropy loss $\ells(w) := -\f{1}{\ns} \sum_{i=1}^\ns \log p_w(y_i | x_i)$
using a sequence of weight updates
\beq{
    \dw(\t)/\dt = -\widehat{\grad} \ells(w);\ w(0) = \ws;
    \label{eq:sgd}
}
these are typically implemented using Stochastic Gradient Descent (SGD). The
notation $\widehat{\grad} \ells(w)$ indicates a stochastic estimate of the
gradient using a subset of the data.

We will denote the marginal of the joint
distributions on the input data as $\ps(x)$ and $\pt(x)$ respectively.

\subsection{Low-capacity models are difficult to transfer}
\label{ss:low_capacity}

The Information Bottleneck (IB)
Principle~\citep{tishby2000information} abstracts a parametric
classifier as a Markov chain $x \to z \to y$ where $z$ is called the
representation. The main idea behind IB is to learn sufficient and minimal
representations that discard information in the data that are not relevant to
predicting
labels~\citep{shwartz2017opening}.
\citet{gao2020free} build upon this idea to design a Lagrangian to study
transfer learning
\beq{
    F_{\ps}(\l, \g) = \min_{\ee, \dd, \cc} \cbr{R + \l D + \g C}.
    \label{eq:ff}
}
Here $\ee$ is an encoder that constructs the representation $z$,  $\cc$ predicts
the labels and $\dd$ is the decoder measures the trade-off between redundant
features that give a lossless representation of the data and discriminative
features for the classifier. The quantities $R, D$ and $C$
are the rate of the encoder, distortion of the decoder and the classification
loss respectively (see~\cref{s:a:low_capacity} for more details).

The situation
with a low-capacity model (encoder and classifier) can be modeled as follows.
If source and target task are similar
then transfer is easy because features of the encoder trained on source
are also good for the target. If the tasks are dissimilar, observe
that KKT conditions give
$\d F_{\ps} = - \d R - \l \d D - \g \d C$.
A low-capacity model needs a large $\g$ to achieve the same classification loss
$C$ which according to the KKT condition leads to larger value of the $D$
(for fixed $\l$). This indicates that the model does not learn
redundant features
that may be potentially useful on a dissimilar target task. This is not
the case for a model with large capacity where~\cref{eq:ff} learns both redundant
and discriminative features, without being forced to make a
trade-off between them.

\ignore{
Consider a binary hypothesis space $\HH$ with a
finite
VC-dimension~\citep{vapnik1998statistical}. Let the population
risk of a hypothesis $h \in \HH$ on the source task be denoted by $\es(h)$
and the risk on the target task by $\et(h)$. For
the sake
of clarity, let us
consider the realizable case when there exists a hypothesis $h^* \in \HH$ which
can perfectly classify both the source and target tasks. Let the support
of the hypothesis space be denoted by $\AA_\HH$, a set $A \in \AA_\HH$ if
there exists a hypothesis $h \in \HH$ that predicts $h(x) = 1$ for all $x \in A$.
The work of~\citet{ben2010theory} shows that
\beq{
    \et(h) \leq \es(h) + \f{1}{2}\ d_{\HH \D \HH}(\ps(x), \pt(x))
    \label{eq:ben_david_bound}
}
where $d_{\HH \D \HH}(\ps(x), \pt(x))$ is a distance
between the source and target input distributions measured using the so-called
symmetric difference hypothesis space $\HH \D \HH$
$\HH \D \HH = \cbr{h(x) \oplus h'(x): h, h' \in
\HH}$. Each hypothesis $g \in \HH \D \HH$ labels as positive all inputs where
a pair of hypotheses $h, h' \in \HH$ disagree. The distance $d_{\HH \D \HH}$
is defined as twice the
maximal gap in probability that the source and target task assign to
subsets $A\in \AA_{\HH \D \HH}$
\[
    d_{\HH \D \HH}(\ps(x), \pt(x)) =
    2 \sup_{A \in \AA_{\HH \D \HH}} \abr{\P_\ps[A] - \P_\pt [A]}.
\]
This is a notion of distance between between the tasks that also takes the
complexity of the hypothesis space into account.
Given the same source and target tasks,
the quantity $d_{\HH \D \HH}(\ps(x), \pt(x))$ is a decreasing function of the
VC-dimension of $\HH$. Indeed, if the number of available hypotheses
is small, the inputs on which they disagree is large,
while for a large hypothesis class
almost all sets $A \subset X$ are contained in $\AA_{\HH \D \HH}$.
Theorem 2 in~\citep{ben2010theory} provides finite-sample bounds for the
case when the model is trained on a mixture of data from source and
target tasks to minimize the convex combination of the empirical losses and
leads to a similar conclusion.

The above argument sketch suggests that higher the capacity of the model,
easier the transfer to a new task. A notion of task distance therefore
also needs to consider the complexity of the hypothesis space being
transferred.
}

\subsection{Fisher-Rao metric on the manifold of probability distributions}
\label{ss:fisher_rao}

Consider a manifold
$M = \cbr{p_w(z): w \in \reals^p}$
of probability distributions. Information Geometry~\citep{amariInformationGeometryIts2016}
studies invariant geometrical structures on such manifolds.
For two points $w,w' \in M$, we can use the
Kullback-Leibler (KL) divergence
$
    \kl \sbr{p_w, p_{w'}} = \int \d p_w(z) \log \rbr{p_w(z)/p_{w'}(z)},
$
to obtain a Riemannian structure on $M$. Such a structure allows
the infinitesimal distance $\ds$ on the manifold to be written as
\beq{
    \ds^2 = 2 \kl \sbr{p_w, p_{w + \dw}} = \sum_{i,j=1}^p g_{ij}\ \dw_i \dw_j
    \label{eq:riemannian_distance}
}
where the Fisher Information Matrix (FIM) $(g_{ij})$ with each
\beq{
    g_{ij}(w) = \int \d p_w(z) \rbr{\partial_{w_i} \log p_w(z)} \rbr{\partial_{w_j} \log p_w(z)}
    \label{eq:fim_def}
}
is positive-definite. The weights $w$
play the role of a coordinate system for computing the distance.
The FIM is the Hessian of the $\kl$-divergence; we may think of
the FIM as quantifying the
amount of information present in the model about the data it
was trained on.
The FIM is the unique metric on $M$ (up to scaling) that is preserved
under diffeomorphisms~\citep{bauerUniquenessFisherRao2016}. This
property motivated~\citet{liang2019fisher} to define
a geometric notion of model complexity
called the Fisher-Rao norm as
\beq{
    \norm{w}_\fr^2 = \inner{w}{g\ w}.
    \label{eq:fisher_rao_norm}
}
This will be discussed further in~\cref{ss:rademacher}.

Given a continuously differentiable curve
$\cbr{w(\t)}_{\t \in [0,1]}$ on the manifold $M$
we can compute its length by integrating the
infinitesimal distance $\abr{\ds}$ along it.
The shortest length curve between two points
$w, w' \in M$
induces a metric on $M$ known as the Fisher-Rao distance~
\citep{rao1945information}
\beq{
    \scalemath{\mscale}{
    \dfr(w, w') = \min_{\substack{w:\ w(0) = w\\ w(1) = w'}}\ \int_0^1
        \sqrt{ \inner{\dot{w}(\t)}{g(w(\t)) \dot{w}(\t)}} \dt}
    \label{eq:fisher_rao};
}
Let us note that shortest paths on a
Riemannian manifold are geodesics, i.e., they are locally ``straight
lines''.

\begin{assumption}[Fisher-Rao distance is computed
only for the conditional distribution]
\label{rem:conditional_assumption}
Although we are interested in the manifold of joint
distributions we will only parametrize the conditional, i.e.,
we write
\beq{
    p^s_w(x, y) := \ps(x)\ p_w(y | x).
    \label{eq:conditional_assumption}
}
The marginal on input is not parameterized. This is a
simplifying assumption and allows us to decouple the data
distribution from the conditional; we can tackle the former using
techniques in optimal transport and the latter using techniques
in information geometry. Parameterizing the joint distribution
directly and using a unified approach to compute the task distance
is possible but will require generative modeling of
$\ps(x)$ which is computationally challenging.
\end{assumption}

Under~\cref{rem:conditional_assumption}, the FIM in~\cref{eq:fim_def}
can be written as
\[
    \scalemath{\mscale}{
    g_{ij}(w) = \E_{\substack{x \sim \ps(x),\ y \sim p_w(y | x)}}
            \sbr{\partial_{w_i} \log p_w(y | x)\ \partial_{w_j} \log p_w(y | x)
            }
    }.
\]
The FIM is difficult to compute for large
models and approximations often work poorly~\citep{kunstner2019limitations}.
Notice that we do not need to compute the FIM but only need to compute the
distance $\abs{\d s}$. Given a trajectory of weights
$\cbr{w(\t)}_{\t \in [0,1]}$
we can compute its length
directly by averaging the
square root of the $\kl$-divergence between the
conditional distributions of labels given data.

\subsection{Transporting the data distribution}
\label{ss:transport}

We next focus on the marginals on the input data. We would like
to modify the input distribution from $\ps(x)$ to $\pt(x)$
during transfer. We will
use tools from optimal transportation (OT) for this purpose; see~
\citet{santambrogio2015optimal,peyreComputationalOptimalTransport2019}
for an elaborate introduction to OT.

Let $\Pi(\ps, \pt)$ be the set of
joint distributions with first marginal equal to $\ps(x)$ and
second marginal $\pt(x')$. The Kantorovich relaxation of the OT
problem~\citet{} solves for
\[
    W_2^2(\ps, \pt) = \inf \cbr{\int \norm{x-x'}^2\ \d \g: \g \in \Pi(\ps, \pt)}
\]
to compute the best joint coupling $\g^* \in \Pi$. The solution of this
problem is the Wasserstein metric $W_2(\ps, \pt)$ between the two distributions.
In this paper, we are interested, not in the Wasserstein metric, but the transport
trajectory that the optimal coupling $\g^*$ entails. This is the subject of displacement
interpolation~\citep{mccannConvexityPrincipleInteracting1997}: it turns out
that the geodesic that joins $\ps$ and $\pt$
is a locally distance minimizing curve in the $W_2$ metric. If $\ptau$ is the
distribution at an intermediate step $\t \in [0, 1]$, we have
\[
    W_2(\ps, \ptau) = \t  W_2(\ps, \pt).
\]
The path that the optimal coupling $\g^*$ takes is therefore a
\emph{constant-speed} geodesic.

We are interested in instantiating this idea
for source and target input \emph{datasets} (we denote
these by $\psh(x)$ and $\pth(x)$)
that consist of finite samples $\ns$ and $\nt$ respectively.
The development is more convenient in this case and the set of
transport plans is a (convex) polytope
\beq{
    \scalemath{\mscale}{\Pi(\psh, \pth) = \cbr{\G \in \reals^{\ns \times
    \nt}_+:\ \G \ones_\ns = \psh, \G^\top \ones_\nt = \pth}}
    \label{eq:polytope}
}
and the optimal coupling is given by
\beq{
    \G^* = \argmin_{\G \in \Pi(\psh, \pth)} \cbr{\inner{\G}{C} - \e H(\G)}
    \label{eq:ot_finite}
}
where $C_{ij} = \norm{x_i - x'_j}_2^2$
is the matrix of pairwise distances between the source and target
data. The inner product in the first term measures the
total cost $\sum_{ij} \G_{ij} C_{ij}$
incurred for the transport and minimizing
it directly is typically done using interior point methods.
This can be accelerated using an entropic penalty
$H(\G) = -\sum_{ij} \G_{ij} \log \G_{ij}$ popularized
by~\citet{cuturi2013sinkhorn}.
McCann's interpolation for the finite-dimensional case with the
quadratic loss $C_{ij}$ can be
written explicitly as the distribution
\beq{
    \ptau(x) = \sum_{i=1}^\ns \sum_{j=1}^\nt \G^*_{ij}\ \delta_{(1 - \t) x_i + \t x'_j}(x).
    \label{eq:interp_finite_only_x}
}
Notice that this is a sum of Dirac-delta distributions
supported at interpolated \emph{input data} $x = (1 - \t) x_i + \t x'_j$.
We can also create pseudo labels for samples from $\ptau$ by a linear
interpolation of the one-hot encoding of their respective labels to get
\beq{
    \ptau(x, y) = \sum_{i=1}^\ns \sum_{j=1}^\nt \G^*_{ij}\ \delta_{(1 - \t) x_i + \t x'_j}(x)\
    \delta_{(1 - \t) y_i + \t y'_j}(y).
    \label{eq:interp_finite}
}

\paragraph{Modifications to the interpolated distribution}

We next make two practically motivated modifications to the interpolated
distribution $\ptau(x,y)$.

First, the quadratic distance $C_{ij} = \norm{x_i - x'_j}^2$
is not a reasonable notion of visual/text data that have strong
local correlations.
It is therefore beneficial to compute $C_{ij}$ using a feature extractor,
say a large neural network $\varphi$, that is trained on some
generic task
\beq{
    C_{ij} := \norm{\varphi(x_i) - \varphi(x'_j)}_2^2.
    \label{eq:cij_phi}
}
This gives us a good coupling matrix $\G$
in practice because the feature space $\varphi(x)$ is
much more Euclidean-like than the original input space;
similar ideas are often employed in the metric
learning literature~\citep{snell2017prototypical,hu2015deep,qi2018low}.

Second, the peculiar form of the
interpolating distribution in~\cref{eq:interp_finite} that consists
of convex combinations of inputs and labels is
the result of the quadratic cost.
Samples from such an interpolated distribution will have
visual artifacts for image-based data.
In practice, we treat the time $\t$ as parameter of
a Beta-distribution $\trm{Beta}(\t, 1-\t)$. Thereby samples
from $\ptau$ are similar to those created in Mixup regularization~
\cite{zhang2017mixup}; a fraction $\t$ of the samples are similar
to those from $\ps$ and the remainder are similar to those from $\pt$.
Note that which data is used to form the Mixup combinations is still
governed by the coupling matrix $\G^*$.
We use this trick for both inputs and labels in our experiments.

\section{Methods}
\label{s:methods}

We now combine the development
of~\crefrange{ss:fisher_rao}{ss:transport} to transport the marginal on the
data and modify the weights on the statistical manifold. This section
also discusses techniques to efficiently implement the approach and make
it scalable to large deep networks. \cref{ss:rademacher} discusses an
alternate perspective on this coupled transfer process using the connection
between Rademacher complexity and the Fisher-Rao norm.

\subsection{Interpolating tasks using a mixture distribution}
\label{ss:mixture}

Interpolating the source and target tasks using a mixture distribution
is a simple way to demonstrate the main idea of our approach. For $\t \in [0,1]$,
consider
\beq{
    \ptau(x, y) = (1 -\t) \ps(x, y) + \t \pt(x, y).
    \label{eq:interp_mixture}
}
This amounts to, on average, $1-\t$ fraction of samples from $\psh$ and the rest from $\pth$.
At time instant $\tau$, weights of the classifier are updated using SGD to
fit samples from $\ptau$. We write this as
\beq{
    \scalemath{\mscale}{
    \d \wt/\dt = \widehat{\grad}_w \E_{(x,y) \sim \ptau}
            \sbr{\log p_{\wt}(y | x)};\ w(0) = \ws}
    \label{eq:sgd_interp}
}
Weights $\wt$
can be thought of as fitted to the task $\ptau$ for every $\t$,
in particular for $\t = 1$, the weights
$w(1)$ is fitted to $\pt$. We can now integrate the length of the
trajectory using~\cref{eq:fisher_rao} to compute the distance
between tasks.

Changes in the data distribution and
updates to the weights are not synchronized in this approach.
For instance, changes in the data may be unfavorable to the current
weights and this forces
a different trajectory in the weight space as the weights struggle to track
$\ptau$. If changes in data were synchronized with those in weights, the weight
trajectory would be different
\emph{and necessarily shorter} because the
KL-divergence in~\cref{eq:riemannian_distance} is large if the conditional distribution
changes quickly; our experimental results also corroborate this.

\subsection{Modifying the task and weights simultaneously}
\label{ss:projection}

We now reintroduce the transport process for the data distribution. For
a coupling matrix $\G$, the interpolated distribution corresponding to the
squared Euclidean cost in OT is given by~\cref{eq:interp_finite}.
Observe that since $\G \in \reals^{\ns \times \nt}$, the $(ij)^{\trm{th}}$
entry of this matrix indicates the
interpolation of source input $x^s_i \in \psh$ with that
of target input $x^t_j \in \pth$.
The distance between two tasks as defined
in~\cref{eq:intro_problem_1,eq:intro_problem_2}
can now be computed for the two \emph{datasets} $\psh$ and $\pth$
iteratively as follows. Given an initialization $\G^0$
computed using a feature extractor in~\cref{eq:cij_phi},
we perform the following updates at each iteration.
\begin{subequations}
\aeq{
    &\scalemath{\mscale}{
    \G^{k+1} = \argmin_{\G \in \Pi} \cbr{\inner{\G}{C^k} - \e H(\G)
            - \l^{-1} \inner{\G}{\G^k}}},
    \label{eq:gkp}\\
    &\scalemath{\mscale}{C^k_{ij} = \int_0^1 \sqrt{2 \kl\sbr{p_{\wtk}(\cdot |
    x^\tau_{ij}), p_{\wtpk}(\cdot | x^\tau_{ij})}}},
    \label{eq:ckp}\\
    &\scalemath{\mscale}{
    \f{\d \wtkp}{\dt} = \widehat{\grad}_w \cbr{\E_{(x,y) \sim \ptau}
            \sbr{\log p_{\wtkp}(y | x)}}},
    \label{eq:sgd_prox}\\
    &\scalemath{0.8}{
    \ptau(x, y) = \sum_{i=1}^\ns \sum_{j=1}^\nt \G^k_{ij}\ \delta_{(1 - \t) x^s_i + \t x^t_j}(x)\
    \delta_{(1 - \t) y^s_i + \t y^t_j}(y)}.
    \label{eq:interp_finite_algorithm}
}
\label{eq:algorithm}
\end{subequations}
At each iteration,
the matrix of costs $C^k_{ij}$ is used to store the cost of
transporting the input $x^s_i$ to $x^t_j$ along the weight trajectory
$\cbr{\wtk}_{\t \in [0,1]}$
obtained in~\cref{eq:sgd_prox}; all trajectories are initialized
at $w^k(0) = \ws$. Observe
that the transport cost has the length of the trajectory in weight space
(the integral in~\cref{eq:fisher_rao}) incorporated into it.
This is our current candidate for the OT
cost matrix, similar to one in~\cref{eq:ot_finite}. Given these costs,
we can compute the new coupling matrix $\G^{k+1}$ using~\cref{eq:gkp}
which is in turn used in the next iteration to compute the interpolated
distribution
$\ptau$ via~\cref{eq:interp_finite_algorithm}. Computing the task distance
is a non-convex optimization problem and we therefore include a proximal
term in~\cref{eq:gkp}
to keep the coupling matrix close to the one in the previous step $\G^k$.
This has the added effect of keeping the entire trajectory of weights
$\cbr{\wtkp}_{\t \in [0,1]}$ close to the trajectory in the previous iteration.
Proximal point iteration~\citep{bauschkeConvexAnalysisMonotone2017}
is insensitive to the step-size $\l$
and it is therefore beneficial to employ it in these updates.

Let us note that the task distance computed in~\cref{eq:algorithm} is
asymmetric, the length of the trajectory for transferring from $\psh$ to $\pth$
is different from the one that transfers from $\pth$ to $\psh$.

\begin{remark}[Fisher-Rao distance can be compared across
different architectures]
The length of the shortest path between two points on the manifold
of distributions $p_w(y | x)$, namely the Fisher-Rao distance, does
not depend on the embedding dimension of the manifold $M$. More specifically,
the distance between tasks as computed by this length does not depend on the
number of parameters of the neural architecture, it only depends upon the
capacity to fit the conditional distribution $p_w(y | x)$. This enables a
desirable property: for the same two tasks, task distance using our approach
is numerically comparable across different architectures.
\end{remark}

\begin{remark}[Scaling up the computation]
\label{rem:scaling_up}
The formulation in~\cref{eq:algorithm} updates $\G \in \reals^{\ns \times \nt}$
and $\wt \in \reals^p$. Executing the updates, even for large deep networks
and standard datasets is easy for the weights. The coupling matrix $\G$
has a large number of entries and
it is therefore challenging. Some common approaches
to handling large-scale OT problems are hierarchical
methods~\citep{lee2019hierarchical},
and greedy computation~\citep{carlier2010knothe}.
In practice, we initialize~\cref{eq:gkp} with a block-diagonal
approximation of the coupling
matrix using~\cref{eq:cij_phi} as the costs and perform mini-batch
updates on the non-zero entries of $\G$. At each iteration, we sample from
the interpolated distribution~\cref{eq:interp_finite_algorithm} using only
entries of $\G^k$ that are a part of the mini-batch.
Experiments in~\cref{s:expt} show that the weight trajectory converges under
such mini-batch updates of $\G^k$.
\end{remark}

\subsection{An alternative perspective via Rademacher complexity}
\label{ss:rademacher}

We next study how the trajectory given by the formulation of~
\cref{eq:algorithm}
looks under the lens of learning theory. We show that we can interpret
the solution of
coupled transfer as a trajectory that minimizes the integral of the
generalization gap as the
task and the weights are modified.
This gives us an intuitive understanding of what qualifies as good transfer;
indeed weight trajectories that do not lead to degradation of
the generalization gap result in weights on the target task $\wb$ that also
generalize well.

We introduce a few quantities before giving the main result. We
consider binary classification in this section for clarity. Given a
$w \in A$, we define the
empirical Rademacher complexity~\citep{bartlettRademacherGaussianComplexities2001}
as
\beq{
    \widehat{\RR}_N(A) = \E_\s \sbr{\sup_{w \in A} \f{1}{N}
    \sum_{i=1}^N\ \e^i \ell(w; x^i, y^i)},
    \label{eq:emp_rad}
}
where $\s^i$ are independent and uniformly
distributed on $\cbr{-1,1}$
and $\ell(w; x^i, y^i)$ is the loss on the $i^{\trm{th}}$ datum
of a dataset $\ph$ with $N$ samples.
We will choose the set $A$ to be the
$r$-ball in the Fisher-Rao norm
\[
    A := \cbr{w: \norm{w}_\fr \leq r},
\]
and write the corresponding complexity as $\hat{\RR}_N(r)$.
The Rademacher complexity is the expectation of the empirical complexity over
draws of different datasets
\[
    \RR_N(r) = \E_{\ph \sim p}\sbr{\widehat{\RR}_N(r)}.
\]
The classical Rademacher complexity-based generalization
bound characterizes the ability of binary classifiers $h$
in a hypothesis class $h \in \HH$ to fit random noise.
We have that for all $h \in \HH$ the absolute
value of the generalization error and the training error is upper bounded by
\beq{
    \RR_{2N}(\HH) + 2 \sqrt{\f{\log(1/\delta)}{N}}
    \label{eq:classical_rademacher}
}
with probability at least $1-\delta$. We build upon this to obtain
the following theorem.

\begin{theorem}
\label{thm:integral_of_rademacher_complexity}
Given a trajectory of the weights $\cbr{\wt}_{\t \in [0,1]}$ and a sequence $0 \leq \t_1 <\t_2 <...<\t_{K}\leq 1 $, then for all
$\e > \f{2}{K} \sum_{k= 1}^{K}\RR_{N}(\norm{w_{t_k}}_\fr)$, the probability that
\[
    \f{1}{K}\sum_{k= 1}^{K} \rbr{\E_{(x,y) \sim p_{\t_k}} \sbr{\ell(\w_{\t_k}, x,y)}
    - \f{1}{N} \sum_{(x,y) \sim \hat{p}_{\t_k}} \ell(\w_{\t_k}, x,y)}
\]
is greater than $\e$ is upper bounded by
\beq{
    \exp \cbr{-\f{2 K}{M^2} \rbr{\e - \f{2}{K} \sum_{k= 1}^{K}\RR_{N}(\norm{w_{t_k}}_\fr)}^2}.
    \label{eq:upper_bound}
}
\end{theorem}
The proof
is provided in~\cref{s:a:proof:integral_of_rademacher_complexity}.
In other words, ensuring that the generalization gap of the model
is small during transfer can be achieved by ensuring that the Rademacher
complexity $\RR_{N}(\norm{\wt}_\fr)$ is small at all times during transfer.

\paragraph{Specializing the result for multi-layer linear models.}
Characterizing the Rademacher complexity of the $r$-ball in Fisher-Rao
norm is difficult. However, for multi-layer linear
classifiers with input domain
$X \subset \reals^p$,~\citet{liang2019fisher}
showed that
\[
    \RR_N(r) \leq r \sqrt{\f{p}{N}}
\]
assuming the data covariance matrix $\E_{x \sim p}[x^\top x]$
is full-rank.
For such a model, minimizing the integral
of the Rademacher complexity is achieved by minimizing
\beq{
    \int_0^1 \sqrt{\inner{\wt}{g(\wt) \wt}}\ \dt;
    \label{eq:rademacher_fisher_rao}
}
which is an upper-bound on the Fisher-Rao distance of the trajectory
between $p_\ws$ and $p_\wb$. The
optimization problem in~\cref{eq:algorithm} finds the latter
and we have thus obtained a close connection between
the coupled transfer process and the intuitive idea that as the model
is transferred, keeping the
generalization gap small at all instants of the trajectory would lead
to a good generalization gap on the target dataset.

As the authors in~\citet{liang2019fisher} discuss further,
the Fisher-Rao norm ball is an envelope of popular norms such as spectral
norm~\citep{bartlett2017spectrally}, group norm~\citep{neyshabur2015norm}
and path norm~\citep{neyshabur2015path} introduced to characterize the complexity
of deep neural networks. The Rademacher complexity using these other norms
can therefore be upper-bounded in terms of the Fisher-Rao norm, which leads
to a similar conclusion for non-linear models.

\section{Experimental evidence}
\label{s:expt}

This section discusses experiments on image classification datasets
which demonstrate that the distance computed using our methods is
consistent with the difficulty of fine-tuning from the source dataset to
the target dataset. We compare and contrast our distance estimtes with existing
methods, discuss how the optimization
problem in~\cref{eq:algorithm} converges to its solution across iterations
and show that larger models are easier to transfer between tasks.

\subsection{Setup}
\label{ss:expt:setup}

We use the CIFAR-10, CIFAR-100 datasets for our experiments. Source and target
tasks consist of subsets of these datasets, each task with one or more
of the original classes inside it.
We show results using an 8-layer
convolutional neural network with ReLU nonlinearities, dropout,
batch-normalization with a final fully-connected layer along with
a larger wide-residual-network (WRN-16-4,~\citep{zagoruyko2016wide}).
More details of pre-processing, architecture
and training procedure are provided in~\cref{s:a:expt}.

\paragraph{Baselines}
The first baseline is Task2Vec~\citep{achille2019task2vec} which
embeds tasks using the diagonal of the FIM of a model
trained on them individually; cosine distance between these vectors
is defined as the distance. We compute the robust approximation of the FIM
via Monte Carlo updates
as done by the original authors.

The second baseline (fine-tuning) directly
computes the length of the trajectory in the
weight space, i.e., $\int \abr{\d w}$.
The trajectory is truncated when validation
accuracy on the target task is 95\% of its final validation accuracy.
Note that
no adaptation of input data is
performed and the model directly takes SGD updates on the target task
after being pre-trained on the source task. The learning rate for each model
was tuned across all datasets
to ensure that the validation accuracy on the target dataset is good and fixed thenceforth
for all experiments.
The number of epochs required
for fine-tuning is a popular way to measure distance
between tasks~\citep{kornblith2019better}.

The third baseline
(uncoupled transfer) uses a mixture of the source and target data,
where the interpolating parameter is sampled from
$\trm{Beta}(\t, 1-\t)$ (see~\cref{ss:mixture}) in order to be consistent with the
way we implement coupled transfer and avoid visual artifacts in the input data.
Length of the trajectory is computed using the FIM metric in this case, which
enables direct comparison of the task distances for coupled and uncoupled transfer.

\subsection{Transferring between CIFAR-10 and CIFAR-100}
\label{ss:c10_c100}

We consider four tasks (i) all vehicles (airplane, automobile, ship, truck) in CIFAR-10,
(ii) the remainder, namely six animals in CIFAR-10, (iii) the entire CIFAR-10 dataset and
(iv) the entire CIFAR-100 dataset. We show results in~\cref{fig:c10_c100}
using 4$\times$4 distance matrices where numbers in each cell indicate the distance between
the source task (row) and the target task (column).

Coupled transfer shows similar trends as fine-tuning, e.g., the tasks animals-CIFAR-10 or
vehicles-CIFAR-10
are close to each other while CIFAR-100 is far away from all tasks (it is closer to CIFAR-10
than others). Task distance is asymmetric in~\cref{fig:CIFAR10},~\cref{fig:CIFAR10_FineTune}.
Distance from CIFAR-10-animals is smaller than animals-CIFAR-10; this is
expected because animals is a subset of CIFAR-10.
Task2Vec distance estimates in~\cref{fig:CIFAR10-Task2Vec}
are qualitatively quite different from these two; the distance matrix is symmetric.
Also, while fine-tuning from animals-vehicles is relatively easy, Task2Vec
estimates the distance between them to be the largest.

This experiment also shows that our approach can scale to medium-scale datasets
and can handle situations when the source and target task have different number
of classes.

\begin{figure}[!htpb]
\centering
\begin{subfigure}[t]{0.325 \linewidth}
\includegraphics[width=\linewidth]{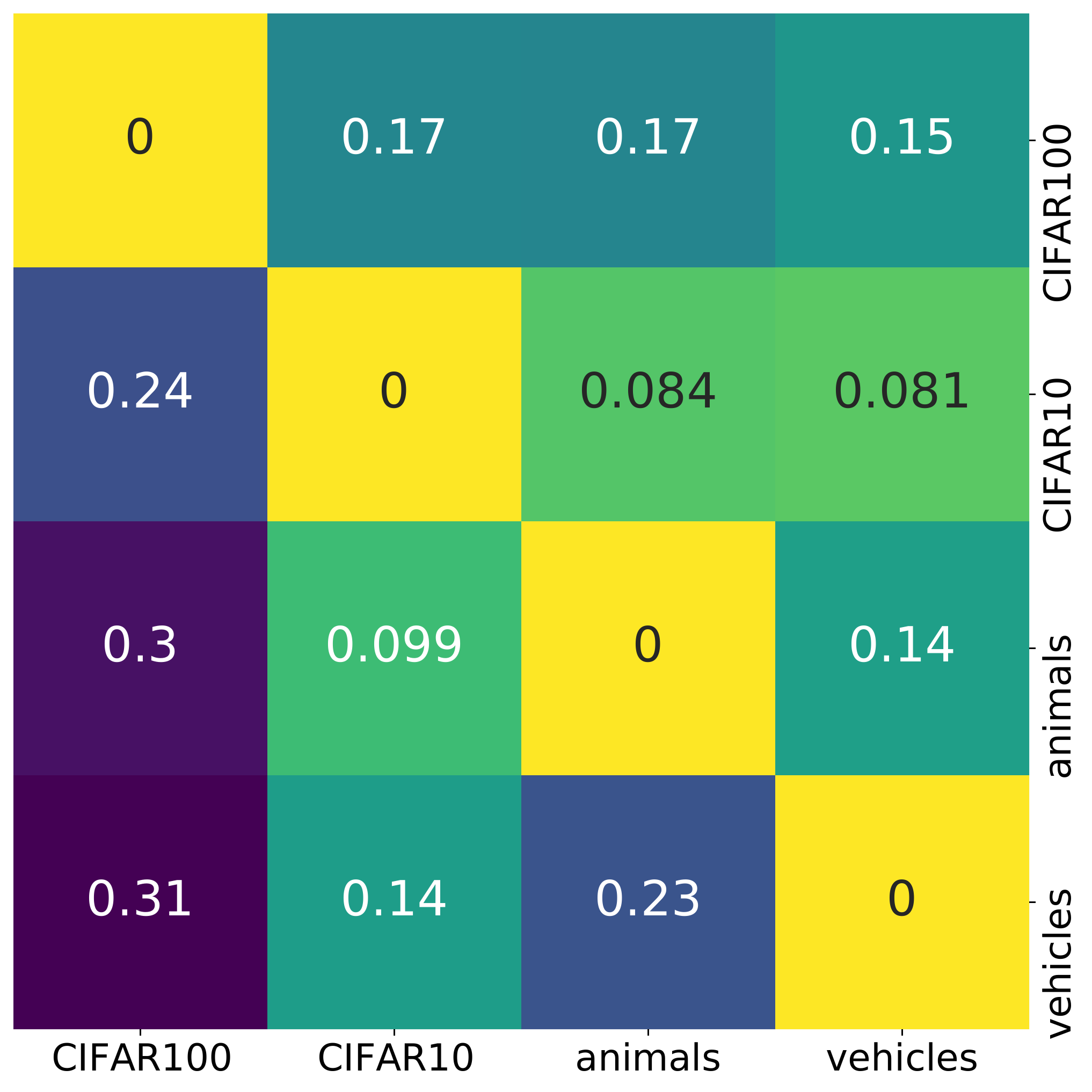}
\caption{}
\label{fig:CIFAR10}
\end{subfigure}
\begin{subfigure}[t]{0.325 \linewidth}
\includegraphics[width=\linewidth]{CIFAR10-Task2Vec}
\caption{}
\label{fig:CIFAR10-Task2Vec}
\end{subfigure}
\begin{subfigure}[t]{0.325 \linewidth}
\includegraphics[width=\linewidth]{CIFAR10_FineTune}
\caption{}
\label{fig:CIFAR10_FineTune}
\end{subfigure}
\caption{
\cref{fig:CIFAR10} shows distances (numbers in the cell)
computed using our coupled transfer
process, ~\cref{fig:CIFAR10-Task2Vec} shows distances estimated using Task2Vec
while~\cref{fig:CIFAR10_FineTune} shows the distance estimating using
fine-tuning. The numerical values of the distances in this figure are not
comparable with each other. Coupled transfer distances satisfy certain sanity
checks, e.g., transferring to a subset task is easier
than transferring from a subset task (CIFAR-10-vehicles/animals).
}
\label{fig:c10_c100}
\end{figure}

\subsection{Transferring among subsets of CIFAR-100}
\label{ss:c100}

We construct five tasks (herbivores, carnivores, vehicles-1, vehicles-2 and
flowers) that are subsets of the CIFAR-100 dataset. Each of these tasks consists
of 5 sub-classes. The distance matrices for coupled transfer, Task2Vec
and fine-tuning are shown in~\cref{fig:CNN},~\cref{fig:Task2Vec}
and~\cref{fig:CNN_FineTune} respectively. We also
show results using uncoupled transfer in~\cref{fig:CNN_uncouplings}.

\begin{figure}[!htpb]
\centering
\begin{subfigure}[t]{\fscale \linewidth}
\includegraphics[width=\linewidth]{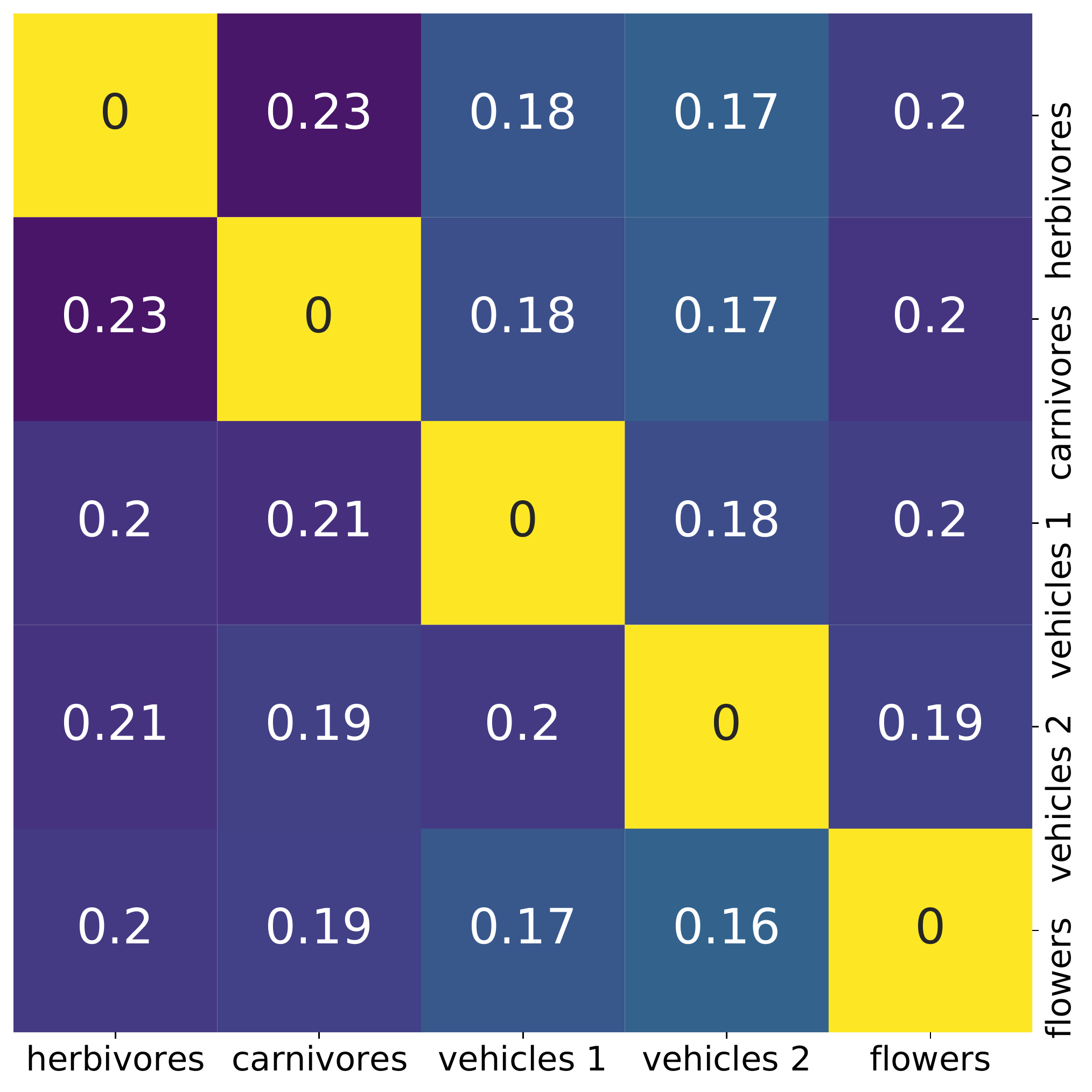}
\caption{}
\label{fig:CNN}
\end{subfigure}
\hspace{2em}
\begin{subfigure}[t]{\fscale \linewidth}
\includegraphics[width=\linewidth]{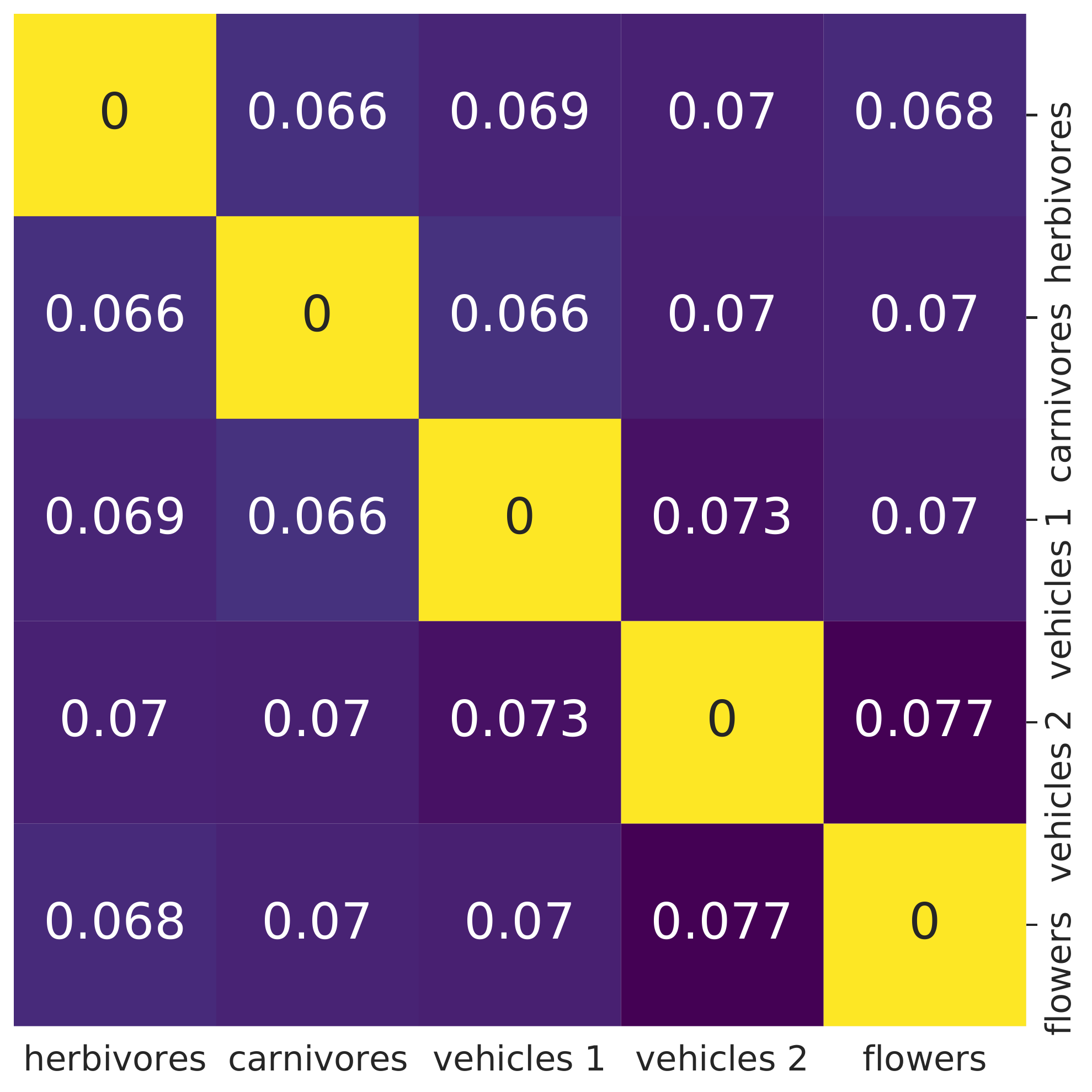}
\caption{}
\label{fig:Task2Vec}
\end{subfigure}

\begin{subfigure}[t]{\fscale \linewidth}
\includegraphics[width=\linewidth]{CNN_FineTune}
\caption{}
\label{fig:CNN_FineTune}
\end{subfigure}
\hspace{2em}
\begin{subfigure}[t]{\fscale \linewidth}
\includegraphics[width=\linewidth]{CNN_uncouplings}
\caption{}
\label{fig:CNN_uncouplings}
\end{subfigure}
\caption{\cref{fig:CNN} shows the distance for coupled transfer,
\cref{fig:Task2Vec}
shows the distance for Task2Vec, \cref{fig:CNN_FineTune} shows the distance
for fine-tuning and~\cref{fig:CNN_uncouplings} shows the distance for uncoupled
transfer. Numerical values the first and the last sub-plot can be compared
directly. Coupled transfer broadly agrees with fine-tuning except for
carnivores-flowers and herbivores-vehicles-1. For all tasks, uncoupled transfer
overestimates the distances compared to~\cref{fig:CNN}.
}
\label{fig:c100}
\end{figure}

Coupled transfer estimates that all these subsets of CIFAR-100 are roughly
equally far away from each other with herbivores-carnivores being the farthest
apart while vehicles-1-vehicles-2 being closest. This ordering is consistent
with the fine-tuning distance although fine-tuning results in an extremely large
value for carnivores-flowers and vehicles-1-herbivores. This ordering is mildly
inconsistent with the distances reported by Task2Vec in~\cref{fig:Task2Vec}
the distance for vehicles-1-vehicles-2 is the highest here. Broadly, Task2Vec
also results in a distance matrix that suggests that all tasks are equally far
away from each other.
As has been reported before~\citep{li2019rethinking},
this experiment also demonstrates the fragility of fine-tuning.

Recall that distances for uncoupled transfer
in~\cref{fig:CNN_uncouplings} can be comparable directly
to those in~\cref{fig:CNN} for coupled transfer.
Task distances for the former are always larger.
Further, distance estimates of uncoupled transfer do not bear much
resemblance with those of fine-tuning; see for example
the distances for vehicles-2-carnivores, flowers-carnivores, and
vehicles-1-vehicles-2.
This demonstrates
the utility of solving a coupled optimization problem in~\cref{eq:algorithm}
which finds a shorter trajectory on the statistical manifold.

\paragraph{Verifying the convergence of coupled transfer}
We use an iterative
algorithm to approximate the optimal couplings between source and target data.
\cref{fig:Fig-4a} shows the evolution of training and test loss as
computed on samples of the interpolated distribution after 4 iterations of~
\cref{eq:algorithm}. As predicted by~\cref{thm:integral_of_rademacher_complexity}
the generalization gap. The training loss increases
towards the middle; this is expected because the interpolated
distribution is maximally far away from both the source and target data
distributions at this point. The convex combination in~\cref{eq:interp_finite_algorithm}
keeps computations tractable but could also be a cause for this increase.

We typically require 4--5
iterations of~\cref{eq:algorithm} for the task
distance to converge; this is shown in~\cref{fig:Fig-4b} for a few instances.
This figure also indicates that computing the transport coupling
in~\cref{eq:ot_finite}
independently of the weights and using this coupling
to modify the weights, as done in say~\citep{cui2018large},
results in a larger distance than if one were to optimize the couplings along
with the weights. The coupled transfer finds shorter trajectories for weights
and will potentially lead to better accuracies on target tasks in
studies like~\citep{cui2018large} because it samples more source data.

\begin{figure}[!htpb]
\centering
\begin{subfigure}[t]{0.45 \linewidth}
\includegraphics[width=\linewidth]{Fig-4a}
\caption{}
\label{fig:Fig-4a}
\end{subfigure}
\begin{subfigure}[t]{0.45 \linewidth}
\includegraphics[width=\linewidth]{Fig-4b}
\caption{}
\label{fig:Fig-4b}
\end{subfigure}
\caption{\cref{fig:Fig-4a} shows the evolution of the training and test
cross-entropy loss on the interpolated distribution
as a function of the transfer steps in the
final iteration of coupled transfer of vehicles-1-vehicles-2.
As predicted by~\cref{thm:integral_of_rademacher_complexity},
generalization gap along the trajectory is small. \cref{fig:Fig-4b}
shows the convergence of the task distance with the number of iterations $k$
in~\cref{eq:algorithm}; the distance typically converges in 4--5 iterations for
these tasks.
}
\label{fig:loss}
\end{figure}

\begin{figure}[!htpb]
\centering
\begin{subfigure}[t]{\fscale \linewidth}
\includegraphics[width=\linewidth]{WideRes-16-4}
\caption{}
\label{fig:wrn164_ours}
\end{subfigure}
\hspace{2em}
\begin{subfigure}[t]{\fscale \linewidth}
\includegraphics[width=\linewidth]{WideRes-16-4-FineTune}
\caption{}
\label{fig:wrn164_finetune}
\end{subfigure}
\caption{\cref{fig:wrn164_ours} shows the task distance using coupled transfer
and~\cref{fig:wrn164_finetune} show the fine-tuning task distance. The numbers
in~\cref{fig:wrn164_ours} can be directly compared to those in~\cref{fig:CNN}.
The larger WRN-16-4 model predicts a smaller task distance for all pairs
compared to the smaller convolutional network in~\cref{fig:CNN}.
}
\label{fig:wrn164}
\end{figure}

\paragraph{Larger capacity results in smaller task distance}

We next show that using a model with higher capacity results in smaller
distances between tasks. We consider a wide residual network
(WRN-16-4) of~\cite{zagoruyko2016wide} and compute distances on subsets of
CIFAR-100 in~\cref{fig:wrn164}. First note that task distances for
coupled transfer in~\cref{fig:wrn164_ours} are consistent
with those for fine-tuning in~\cref{fig:wrn164_finetune}.
Coupled transfer distances in~
\cref{fig:wrn164_ours}
are much smaller compared to those in~\cref{fig:CNN}.
This is consistent with our argument in~\cref{ss:low_capacity}.
Roughly speaking, a high-capacity model can learn a rich set of features,
some discriminative and others redundant not
relevant to the source task. These redundant features are useful
if target task is dissimilar to the source.
This experiment also demonstrates that the information-geometric distance
computed by coupled transfer can be compared directly across different
architectures; this is not so for most methods in the literature to compute
distances between tasks. This gives a constructive strategy for selecting
architectures for transfer learning.

\section{Related work}
\label{s:related_work}

\paragraph{Domain-specific methods}
A rich understanding of task distances has been developed
in computer vision, e.g.,~\citet{zamir2018taskonomy}
compute pairwise distances when different tasks such as classification,
segmentation etc. are performed on the same input data. The goal
of this work, and others such as~\citep{cui2018large},
is to be able to decide
which source data to pre-train to generalize well on a target task.
Task distances have also been widely discussed in the
multi-task learning~\citep{caruana1997multitask} and
meta/continual-learning~\citep{liu2019localized,pentina2014pac,hsu2018unsupervised}.
The natural language processing literature also prevents several
methods to compute similarity between input
data~\citep{mikolov2013efficient,pennington2014glove}.

Most of the above methods are
based on evaluating the difficulty of fine-tuning,
or computing the similarity in some embedding space.
It is difficult to ascertain whether the distances obtained thereby
are truly indicative of the difficulty of transfer; fine-tuning
hyper-parameters often need to be carefully chosen~\citep{li2019rethinking}
and the embedding space is not unique.
For instance, the uncoupled transfer process that modifies the input data
distribution will lead to a different estimate of task distance.

\paragraph{Information-theoretic approaches}
We build upon a line of work that combines
generative models
and discriminatory classifiers (see~
\citep{jaakkola1999exploiting,perronnin2010improving}
to name a few) to construct a notion of similarity
between input data.
Modern variants of this
idea include Task2Vec~\citep{achille2019task2vec} which embeds the task
using the diagonal of the FIM and computes distance
between tasks using the cosine distance for this embedding.
The main hurdle in Task2Vec and similar approaches is to design
the architecture for computing FIM: a
small model will indicate that tasks are far away.
\citet{achilleDynamicsReachabilityLearning2019,achilleInformationComplexityLearning2019}
use the KL
divergence between the posterior weight distribution and a prior to
quantify the complexity of a task; distance
between tasks is defined to be the increase in complexity when
the target task is added to the source task. This is an elegant
formalism to define task distances but instantiating these ideas for
deep networks requires drastic approximations, e.g., a Gaussian posterior
on the weight space.


\paragraph{Model complexity}
Learning theory typically studies out-of-sample performance on a single task.
\ignore{
In~\cref{ss:low_capacity}, we used the results~\citet{ben2010theory}
to argue that transfer is easier with high-capacity models. They also train
models on a mixture distribution of the source and target data to minimize the
convex combination of the two empirical risks; this is close to the
interpolation distribution in~\cref{eq:interp_finite_algorithm}.
}
Our goal is to account for the model complexity while defining the task
distance.
Complexity measures such as
VC-dimension~\citep{vapnik1998statistical},
come with a number of caveats when applied to deep networks
because these measures are not reparameterization invariant.
We exploited the geometric characterization of
the statistical manifold~\citet{amariInformationGeometryIts2016} that
leads to invariant quantities such as the Fisher-Rao distance.


\paragraph{Coupled transfer of data and the model}

A key idea of our work is to observe that the marginal
on the input can be transported in addition to the weights of the model.
This is motivated from two recent studies. \citet{gao2020free}
develop an algorithm that keeps
the classification loss unchanged across transfer. Their
method interpolates between the source and target data distribution
using a mixture distribution (we use it as a baseline,
see~\cref{ss:mixture} and~\cref{ss:c100}).
Our work exploits this idea and computes
the optimal way to modify both the input distribution and the weights.
We use ideas from optimal transportation to compute the transport on
input data; see~\cite{cui2018large} who also solve an optimal transport
problem approximately to estimate task distances.
Coupled transport problems on the input data
are also solved for unsupervised
translation~\citep{alvarez-melisGromovWassersteinAlignmentWord2018}.

\section{Discussion}
\label{s:discussion}

Our work is an attempt to theoretically understand when transfer
is easy and when it is not. An often over-looked idea in large-scale
transfer learning is that the dataset need not remain fixed to the target
task during transfer. We heavily exploit this idea in the present
paper and develop
an optimization framework to adapt both the input data distribution
and the weights from the source to the target. Although
a metric is never unique, this gives legitimacy
to our task distance. We compute the \emph{shortest}
distance in information space, i.e., the manifold of the
conditional distributions. It is remarkable that this concept
is closely related to the intuitive idea that a good transfer
algorithm is one that keeps the generalization gap small during transfer,
in particular at the end on the target task.

The most drastic approximation in this paper was to forgo a generative
framework for the input distribution.
This opens an interesting direction
for future work which formulates
the distance between tasks simply as the shortest geodesic on
the manifold of joint distributions $p_w(x, y)$.

\begin{small}
\clearpage
\setlength{\bibsep}{0.25em}
\bibliography{main}
\bibliographystyle{apalike}
\end{small}